%% file: main.tex
\documentclass{article} 
\usepackage{iclr2024_conference,times}
\input{math_commands.tex}  
\usepackage{hyperref}
\usepackage{bbm}
\usepackage{graphicx}
\usepackage{subfigure}
\usepackage{adjustbox} 
\usepackage{empheq}
\usepackage{caption}  
\usepackage{amsmath, amssymb, amsthm,bm}
\usepackage{mathtools}
\usepackage{booktabs}
\usepackage{color}
\usepackage{enumitem}
\usepackage{wrapfig}
\usepackage{url}
\usepackage{multirow}
\usepackage[symbol]{footmisc}

\definecolor{darkpastelgreen}{rgb}{0.01, 0.75, 0.24}
\DeclarePairedDelimiterX{\infdivx}[2]{[}{]}{%
  #1\;\delimsize\|\;#2%
}
\newcommand{\kl}{D_{KL}\infdivx}

\title{Interpretable Diffusion via \\ Information Decomposition}

\author{Xianghao Kong\textsuperscript{1 *} ,
Ollie Liu\textsuperscript{2 *},
Han Li\textsuperscript{1}, 
Dani Yogatama\textsuperscript{2}, 
Greg Ver Steeg\textsuperscript{1} \\
\textsuperscript{1}University of California Riverside, 
\textsuperscript{2}University of Southern California \\
\texttt{\{xkong016,hli358,gregoryv\}@ucr.edu}, \texttt{\{zliu2898, yogatama\}@usc.edu} 
}

\iclrfinalcopy 
\begin{document}
\vspace{-1mm}

\maketitle
\vspace{-1mm}
\begin{abstract}
Denoising diffusion models enable conditional generation and density modeling of complex relationships like images and text. 
However, the nature of the learned relationships is opaque making it difficult to understand precisely what relationships between words and parts of an image are captured, or to predict the effect of an intervention. 
We illuminate the fine-grained relationships learned by diffusion models by noticing a precise relationship between diffusion and information decomposition. 
Exact expressions for mutual information and conditional mutual information can be written in terms of the denoising model. 
Furthermore, \emph{pointwise} estimates can be easily estimated as well, allowing us to ask questions about the relationships between specific images and captions.
Decomposing information even further to understand which variables in a high-dimensional space carry information is a long-standing problem.
For diffusion models, we show that a natural non-negative decomposition of mutual information emerges, allowing us to quantify informative relationships between words and pixels in an image. 
We exploit these new relations to measure the compositional understanding of diffusion models, to do unsupervised localization of objects in images, and to measure effects when selectively editing images through prompt interventions. 
\end{abstract}
\vspace{-1mm}
\section{Introduction}

Denoising diffusion models are the state-of-the-art for modeling relationships between complex data like images and text. 
While diffusion models exhibit impressive generative abilities, we have little insight into precisely what relationships are learned (or neglected).
Often, models have limited value without the ability to dissect their contents. 
For instance, in biology specifying which variables have an effect on health outcomes is critical.
As AI advances, more principled ways to probe learned relationships are needed to reveal and correct gaps between human and AI perspectives. 

\begin{figure}[htbp]
    \centering
    \includegraphics[width=0.93\textwidth,trim={5cm 10mm 4cm 7mm},clip]{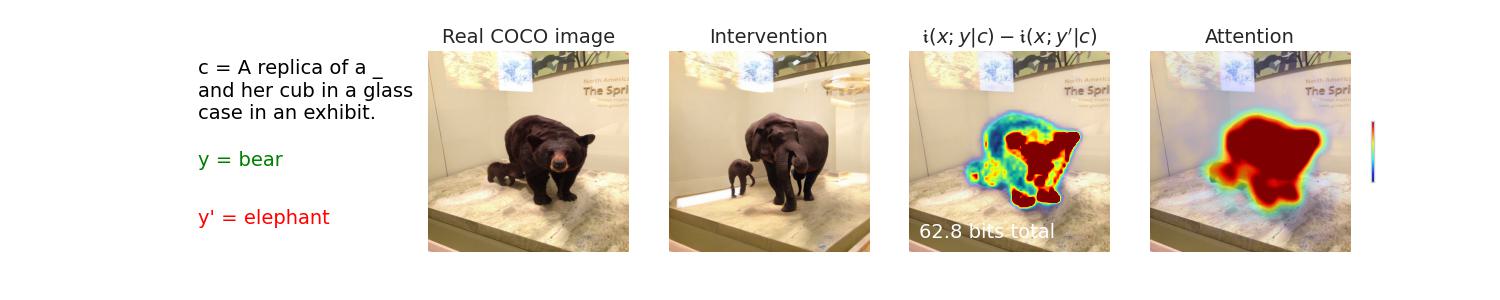}  
        \includegraphics[width=0.93\textwidth,trim={5cm 10mm 4cm 13mm},clip]{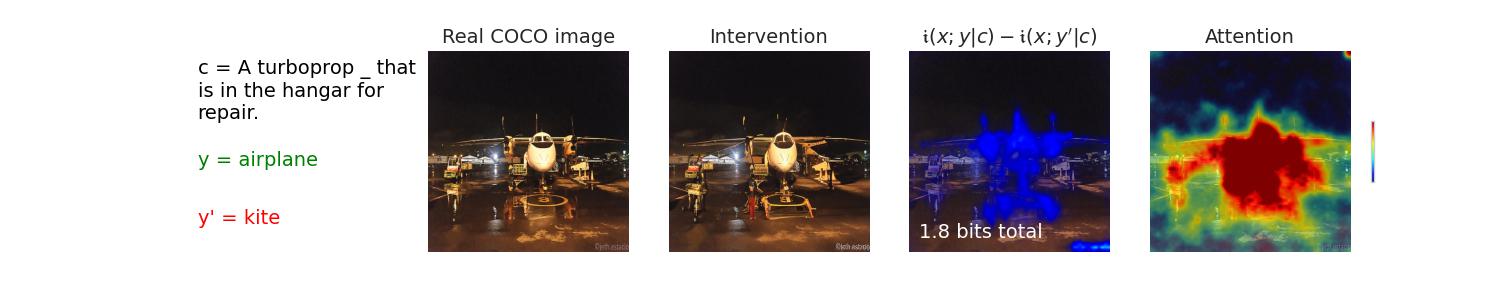}  
    \caption{
    We start (left) with a real image from the COCO dataset. We do a ``prompt intervention'' (\S\ref{sec:intervention}) to generate a new image. Next we show conditional mutual information, illustrated using our pixel-wise decomposition, and attention maps for the modified word. Top row shows an image where prompt intervention has an effect, while in the bottom row it has little effect. Conditional mutual information reflects the effect of intervention while attention does not.}
    \label{fig:abstract} \vspace{-5mm}
\end{figure}

Quantifying the relationships learned in a complex space like text and images is difficult. 
Information theory offers a black-box method to gauge how much information flows from inputs to outputs.  
This work proceeds from the novel observation that diffusion models naturally admit a simple and versatile information decomposition that allows us to pinpoint information flows in fine detail which allows us to understand and exploit these models in new ways.

For denoising diffusion models, recent work has explored how \emph{attention} can highlight how models depends on different words during generation~\citep{tang2022daam,concept_discovery, zhao2023unleashing, tian2023diffuse, wang2023diffusion, zhang2023diffusionengine, ma2023diffusionseg, he2023discriminative}. 
Our information-theoretic approach diverges from attention-based methods in three significant ways. 
\underline{First}, attention requires not just white-box access but also dictates a particular network design. Our approach abstracts away from architecture details, and may be useful in the increasingly common scenario where we interact with large generative models only through black-box API access~\citep{dalle2}. 
\underline{Second}, while attention is engineered toward specific tasks such as segmentation \citep{tang2022daam} and image-text matching \citep{he2023discriminative}, our information estimators can adapt to diverse applications. As an illustrative example, in \S\ref{sec:aro} we automate the evaluation of compositional understanding for Stable Diffusion \citep{latent_diffusion}. 
\underline{Third}, information flow as a dependence measure better captures the effects of interventions. Attention within a neural network does not necessarily imply that the final output depends on the attended input. 
Our Conditional Mutual Information (CMI) estimator correctly reflects that a word with small CMI will not affect the output (Fig.~\ref{fig:abstract}).
We summarize our main contributions below. 
\begin{itemize}[nosep]
    \item We show that denoising diffusion models directly provide a natural and tractable way to decompose information in a fine-grained way, distinguishing relevant information at a per-sample (image) and per-variable (pixel) level. The utility of information decomposition is validated on a variety of tasks below.  
    \item We provide a better quantification of compositional understanding capabilities of diffusion models. 
    We find that on the ARO benchmark \citep{vlmbagofwords}, diffusion models are significantly underestimated due to sub-optimal alignment scores.
    
    
    \item We examine how attention and information in diffusion models localize specific text in images. While neither exactly align with the goal of object segmentation, information measures more effectively localize abstract words like adjectives, adverbs, and verbs.
    \item How does a prompt intervention modify a generated image? It is often possible to surgically modify real images using prompt intervention techniques, but sometimes these interventions are completely ignored. 
    We show that CMI is more effective at capturing the effects of intervention, due to the ability to take contextual information into account. 
\end{itemize}

\section{Methods: Diffusion is Information Decomposition}\label{sec:methods}

\subsection{Information-theoretic Perspective on Diffusion Models}
A diffusion model can be seen as a noisy channel that takes samples from the data distribution, $\vx \sim p(X=\vx)$, 
and progressively adds Gaussian noise, $\vz \equiv \sqrt{\sigma(\logsnr)} \vx + \sqrt{\sigma(-\logsnr)} \eps$, with $\eps \sim \mathcal N(0, \mathbb I)$ (a variance preserving Gaussian channel with log SNR, $\alpha$, using the standard sigmoid function). 
By learning to reverse or \emph{denoise} this noisy channel, we can generate samples from the original distribution~\citep{jaschaneq}, a result with remarkable applications~\citep{dalle2}. 
The Gaussian noise channel has been studied in information theory since its inception~\citep{shannon}. 
A decade before diffusion models appeared in machine learning, 
 \citet{guo} demonstrated that the information in this Gaussian noise channel, $I(X;X_\logsnr)$, is \emph{exactly} related to the mean square error for optimal signal reconstruction.
This result was influential because it demonstrated for the first time that \emph{information-theoretic} quantities could be related to \emph{estimation of optimal denoisers}.  
In this paper, we are interested in extending this result to other mutual information estimators, and to \emph{pointwise} estimates of mutual information. Our focus is not on learning the reverse or denoising process for generating samples, but instead to \emph{measure relationships} using information theory.

For our results, we require the optimal denoiser, or Minimum Mean Square Error (MMSE) denoiser for predicting $\eps$
 at each noise level, $\logsnr$. 
\be\label{eq:denoiser}
\epshat(\vx) \equiv \arg \min_{\epsbar(\cdot)} \E_{p(\vx), p(\eps)} \left[ \norm{\eps - \epsbar(\vz)} \right]
\ee
Note that we predict the noise, $\eps$, but could equivalently predict $\vx$. This optimal denoiser is exactly what diffusion models are trained to estimate. Instead of using denoisers for generation, we will see how to use them for measuring information-theoretic relationships. 

For the denoiser in Eq.~\ref{eq:denoiser}, the following expression holds exactly.
\be\label{eq:nll}
-\log p(\vx) = \half \int \E_{p(\eps)} \left[ \norm{\eps - \epshat(\vz)} \right] d\logsnr + const
\ee
The value of the constant, $const$, will be irrelevant as we proceed to build Mutual Information (MI) estimators and a decomposition from this expression. 
This expression shows that solving a denoising \emph{regression} problem (which is easy for neural networks) is \emph{equivalent} to density modeling. 
No differential equations need to be referenced or solved to make this exact connection, unlike the approaches appearing in \citet{diffusion_sde} and \citet{mcallester2023mathematics}.
The derivation of this result in \citet{kong22} closely mirrors \citet{guo}'s original result, and is shown in App.~\ref{app:proof} for completeness. 
This expression is extremely powerful and versatile for deriving fine-grained information estimators, as we now show.

\subsection{Mutual Information and Pointwise Estimators} \label{sec:mi}
Note that \eqref{eq:nll} also holds with arbitrary conditioning. Let $\vx, \vy \sim p(X=\vx, Y=\vy)$ and $\epshat(\vz | \vy)$ be the optimal denoiser for $p(\vx | \vy)$ as in \eqref{eq:denoiser}. Then we can write the conditional density as follows.
\be\label{eq:cond_nll}
-\log p(\vx|\vy) = \half \int \E_{p(\eps)} \left[ \norm{\eps - \epshat(\vz|\vy)} \right] d\logsnr + const
\ee
This directly leads to an estimate of the following useful Log Likelihood Ratio (LLR). 
\be\label{eq:llr}
\log p(\vx|\vy) - \log p(\vx) = \half \int \E_{p(\eps)} \left[ \norm{\eps - \epshat(\vz)} - \norm{\eps - \epshat(\vz|\vy)} \right] d\logsnr
\ee
The LLR is the integrated reduction in MMSE from conditioning on auxiliary variable, $\vy$. 
The mutual information, $I(X;Y)$ can be defined via this LLR, 
$I(X;Y) \equiv \E_{p(\vx, \vy)} \left[  \log p(\vx|\vy) - \log p(\vx) \right].$
We write MI using information theory notation, where capital $X,Y$ are used to refer to functionals of random variables with the distributions $p(\vx, \vy)$ \citep{cover}. 
MI is an average measure of dependence, but we are often interested in the strength of a relationship for a single point, or \emph{pointwise information}. 
Pointwise information for a specific $\vx, \vy$ is sometimes written with lowercase as $\ii(\vx; \vy)$ and is defined so that the average recovers MI~\citep{finn2018pointwise}. 
\citet{fano} referred to what we call ``mutual information'' as ``average mutual information'', and considered what we call pointwise mutual information to be the more fundamental quantity. Pointwise information has been especially influential in NLP~\citep{levy2014neural}.
$$
I(X;Y) = \E_{p(\vx, \vy)} [ \ii(\vx; \vy) ]  \qquad  \mbox{\emph{Defining property of pointwise information}}
$$
Pointwise information is not unique and both quantities below satisfy this property. 
\begin{equation}\label{eq:pointwise}
\begin{split}
\ii^s(\vx; \vy) &\equiv \half \int \E_{p(\eps)} \left[ \norm{\eps - \epshat(\vz)} - \norm{\eps - \epshat(\vz|\vy)} \right] d\logsnr  \\
\ii^o(\vx; \vy) &\equiv \half \int \E_{p(\eps)} \left[ \norm{\epshat(\vz) - \epshat(\vz|\vy)} \right] d\logsnr  
\end{split}
\end{equation}
The first \textbf{\underline{s}tandard definition} comes from using $\log p(\vx|\vy) - \log p(\vx)$ written via \eqref{eq:llr}. 
The second more compact definition is derived using the \textbf{\underline{o}rthogonality principle} in \S\ref{sec:orthogonality}.
$\ii^s$ has higher variance due to the presence of extra $\eps$ terms, while $\ii^o$ has lower variance and is always non-negative. 
We will explore both estimators, but find the lower variance version that exploits the orthogonality principle is generally more useful (see \S\ref{app:MMSE}). 
Note that while (average) MI is always non-negative, pointwise MI can be negative as $\ii^s(\vx; \vy) = \log p(\vx|\vy) - \log p(\vx) < 0$ occurs when the observation of $\vy$ makes $\vx$ appear \emph{less} likely. We can say negative pointwise information signals a ``misinformative'' observation~\citep{finn2018pointwise}.

All these expressions can be given conditional variants, where we condition on a random variable, $C$, defining the context. 
CMI and its pointwise expression can be related as $I(X;Y|C) = \E_{p(\vx, \vy, \vc)} [ \ii(\vx; \vy | \vc) ]$.
The pointwise versions of \eqref{eq:pointwise} can be obtained by conditioning all the denoisers on $C$, e.g., $\epshat(\vz|\vy) \rightarrow \epshat(\vz|\vy, \vc)$.

\subsection{Pixel-wise Information Decomposition}  \label{sec:cmi}
The pointwise information, $\ii(\vx;\vy)$, does not tell us which variables, $x_j$'s, are informative about which variables, $y_k$. If $\vx$ is an image and $\vy$ represents a text prompt, then this would tell us which parts of the image a particular word is informative about. 
One reason that information decomposition is highly nontrivial is that scenarios can arise where information in variables is synergistic, for example~\citep{williamsbeer}. Our decomposition instead proceeds from the observation that the correspondence between information and MMSE leads to a natural decomposition of information into a sum of terms for each variable. If $\vx \in \sR^n$, we can write $\ii(\vx; \vy) = \sum_{j=1}^n \ii_j(\vx; \vy)$ with: 
%
\begin{equation}\label{eq:pixel}
\begin{aligned}
\ii^s_j(\vx; \vy) &\equiv \half \int \E_{p(\eps)} \left[ (\eps - \epshat(\vx_\alpha))_j^2 - (\eps - \epshat(\vx_\alpha|\vy))_j^2 \right] d\logsnr  \\
\ii^o_j(\vx; \vy) &\equiv \half \int \E_{p(\eps)} \left[ (\epshat(\vx_\alpha) - \epshat(\vx_\alpha|\vy))_j^2 \right] d\logsnr 
\end{aligned}
\end{equation}

In other words, both variations of pointwise information can be written in terms of squared errors, and we can decompose the squared error into the error for each variable. For pixel-wise information for images with multiple channels, we sum over the contribution from each channel. 

We can easily extend this for conditional information. Let $\vx$ represent a particular image, $\vy = \{ y_*, \vc \} = $ \{``object'', ``a person holding an \_''\}. Then we can estimate $\ii_j(\vx;y_*|\vc)$, which represents the information that word $y_*$ has about variable $\vx_j$, conditioned on the context, $\vc$. To get estimates using \eqref{eq:pixel}, we just add conditioning on $\vc$ on both sides. Denoising images conditioned on arbitrary text is a standard task for diffusion models. 
An example of the estimator is shown in Fig.~\ref{fig:abstract}, where the highlighted region represents the pixel-wise value of $\ii^{o}_j(\vx;y_*|\vc)$ for pixel $j$. 

\subsection{Numerical Information Estimates}
All information estimators we have introduced require evaluating a one-dimensional integral over an infinite range of SNRs. To estimate in practice we use importance sampling to evaluate the integral as in \citep{vdm,kong22}. We use a truncated logistic for the importance sampling distribution for $\logsnr$. Empirically, we find that contributions to the integral for both very small and very large values of $\logsnr$ are close to zero, so that truncating the distribution has little effect.  
Unlike MINE~\citep{belghazi2018mine} or variational estimators \citep{poole2019variational} the estimators presented here do not depend on optimizing a direct upper or lower bound on MI. 
Instead, the estimator depends on finding the MMSE of both the conditional and unconditional denoising problems, and then combining them to estimate MI using \eqref{eq:llr}. However, these two MMSE terms appear with opposite signs. In general, any neural network trained to minimize MSE may not achieve the global minimum for either or both terms, so we cannot guarantee that the estimate is either an upper or lower bound. 
In practice, neural networks excel at regression problems, so we expect to achieve reasonable estimates. In all our results we use pretrained diffusion models which have been trained to minimize mean square under Gaussian noise, as required by \eqref{eq:denoiser} (different papers differ in how much weight each $\logsnr$ term receives in the objective, but in principle the MMSE for each $\alpha$ is independent, so the weighting shouldn't strongly affect an expressive enough neural network).
\vspace{-1mm}
\section{Results}

Section~\ref{sec:methods} establishes a precise connection between optimal denoisers and information. 
For experiments we consider diffusion models as approximating optimal denoisers which can be used to estimate information.  
All our experiments are performed with pre-trained latent space diffusion models~\citep{latent_diffusion}, Stable Diffusion v2.1 from Hugging Face unless otherwise noted.  
Latent diffusion models use a pre-trained autoencoder to embed images in a lower resolution space before doing typical diffusion model training. We always consider $\vx$ as the image in this lower dimensional space, but in displayed images we show the images after decoding, and we use bilinear interpolation when visualizing heat maps at the higher resolution. See \S\ref{sec:settings} for experiment details and links to open source code.

\subsection{Relation Testing with Pointwise Information}
\label{sec:aro}

First, we consider information decomposition at the ``pointwise'' or per-image level. We make use of our estimator to compute summary statistics of an image-text pair and quantify qualitative differences across samples. As a novel application scenario, we apply our pointwise estimator to analyze compositional understanding of Stable Diffusion on the ARO benchmark \citep{vlmbagofwords}. Referring readers to \citep{vlmbagofwords} for more detailed descriptions, the ARO benchmark is a suite of discriminative tasks wherein a VLM is commissioned to align an image $\vx$ with its ground-truth caption $\vy$ from a set of perturbations $\ \gP = \{\tilde{\vy}_j\}_{j=1}^M$ induced by randomizing constituent orders from $\vy$. The compositional understanding capability of a VLM is thus measured by the accuracy: 
$$
\E_{\vx, \vy}[\mathbbm{1}\big(\vy = \argmax_{\vy^\prime \in \{\vy\} \bigcup \gP} s(\vx, \vy^\prime) \big)],
$$
where $s: \gX \times \gY \rightarrow \R$ is an alignment score. For contrastive VLMs, $s$ is chosen to be the cosine similarity between encoded image-text representations. We choose $s(\vx, \vy) \equiv \ii^o(\vx; \vy)$ as our score function for diffusion models. In contrast, \citep{he2023discriminative} compute $s$ as an aggregate of latent attention maps, while \citep{krojer2023diffusion} adopt the negative MMSE. In Table \ref{tab:aro}, we report performances of OpenCLIP \citep{ilharco_gabriel_2021_5143773} and Stable Diffusion 2.1 \citep{latent_diffusion} on the ARO benchmark, while controllilng model checkpoints with the same text encoder for fair comparison.


\vspace{-1mm}
\begin{table}[h]
\caption{Accuracy (\%) of Stable Diffusion and its OpenCLIP backbone on the ARO Benchmark. $\star$: conducts additional fine-tuning with compositional-aware hard negatives.}
\label{tab:aro}
\begin{center}
\small
\begin{tabular}{lcccc}
\hline
\multicolumn{1}{c}{\bf Method}  & \bf VG-A & \bf VG-R & \bf COCO & \bf Flickr30k
\\ \hline 
OpenCLIP \citep{ilharco_gabriel_2021_5143773} & 64.6 & 51.4 & 32.8 & 40.5\\
DiffITM \citep{krojer2023diffusion} & 62.9 & 50.0 & 23.5 & 33.2\\
$\text{HardNeg-DiffITM}^\star$ \citep{krojer2023diffusion} & 67.6 & 52.3 & 34.4 & 48.6\\
Info. (Ours) & \textbf{72.0} & \textbf{69.1} & \textbf{40.1} & \textbf{49.3}\\
\hline
\end{tabular}
\end{center}
\end{table}
\vspace{-1mm}

We observe that Stable Diffusion markedly improves in compositional understanding over OpenCLIP. Since the text encoder is frozen, we can attribute this improvement solely to the denoising objective and the visual component. More importantly, our information estimator significantly outperforms MMSE \citep{krojer2023diffusion}, decidedly proving that previous works underestimate compositional understanding capabilities of diffusion models. Our observation provides favorable evidence for adapting diffusion models for discriminative image-text matching \citep{he2023discriminative}. However, these improvements are smaller compared to contrastive pre-training approaches that make use of composition-aware negative samples \citep{vlmbagofwords}. 
\vspace{-1mm}
\begin{wraptable}{r}{6.4cm}
\centering
\caption{Selected results on VG-R.}
\label{tab:vgr-select}
\small
\begin{tabular}{lrrr}
\toprule
{}  &      Info.  ($\uparrow$) &   OpenCLIP ($\uparrow$) \\
\midrule
\textbf{Accuracy (\%)}  & 69.1 & 51.4 & {} \\ \midrule
{\color{magenta} beneath}  & {\color{magenta} 50.0} & {\color{magenta} 90.0}\\
{\color{magenta} covered in} & {\color{magenta} 14.3} & {\color{magenta} 50.0}\\
{\color{magenta} feeding} & {\color{magenta} 50.0} & {\color{magenta} 100.0}\\
{\color{darkpastelgreen} grazing on} & {\color{darkpastelgreen} 60.0} & {\color{darkpastelgreen} 30.0}\\
{\color{darkpastelgreen} sitting on} &  {\color{darkpastelgreen} 78.9} & {\color{darkpastelgreen} 49.7}\\
{\color{darkpastelgreen} wearing} & {\color{darkpastelgreen} 84.1} & {\color{darkpastelgreen} 44.9}\\
\bottomrule
\end{tabular}
\end{wraptable}

\vspace{-1mm}
In Table \ref{tab:vgr-select} we report a subset of fine-grained performances across relation types, highlighting those with over $30\%$ performance improvement in {\color{darkpastelgreen} green} and those with over $30\%$ performance decrease in {\color{magenta} magenta}. Interestingly, our highlighted categories correlate well with performance changes incurred by composition-aware negative training, despite using different CLIP backbones (c.f. Table 2 of \citep{vlmbagofwords}). Most improvements are attributed to verbs that associate subjects with conceptually distinctive objects (e.g. ``\textit{\underline{A boy} sitting on \underline{a chair}}"). Our observation suggests that improvements in compositional understanding may stem predominantly from these ``low-hanging fruits", and partially corroborate the hypothesis in \citep{rassin2023linguistic}, which posits that incorrect associations between entities and their visual attributes are attributed to the text encoder's inability to encode linguistic structure. We refer readers to \S\ref{sec:aro-appendix} for complete fine-grained results, implementation details, and error analysis.

\subsection{Pixel-wise Information and Word Localization} 
\label{sec:word_loc}
Next, we explore ``pixel-wise information'' that words in a caption contain about specific pixels in an image. According to \S\ref{sec:mi} and \S\ref{sec:cmi}, it naturally leads us to consider two potential approaches for validating this nuanced relationship.  The first one entails concentrating solely on the mutual information between the object word and individual pixels. The second approach involves investigating this mutual information when given the remaining context in the caption, i.e., conditional mutual information. As the success of our experiment relies heavily on the alignment between images and text, we carefully filtered two datasets, COCO-IT and COCO-WL from the MSCOCO \citep{lin2015microsoft} validation dataset. For specific details about dataset construction, please refer to \S\ref{app:data}. Meanwhile, we also provide an image-level information analysis on these two datasets in \S\ref{app:rela_mi_cmi}, and visualize the diffusion process and its relation to information for 10 cases in \S\ref{app:MMSE}.

\subsubsection{Visualize mutual information and conditional mutual information}
\begin{figure}[h]
    \centering
    \includegraphics[width=0.45\textwidth,trim={2cm 9mm 1.7cm 13mm},clip]{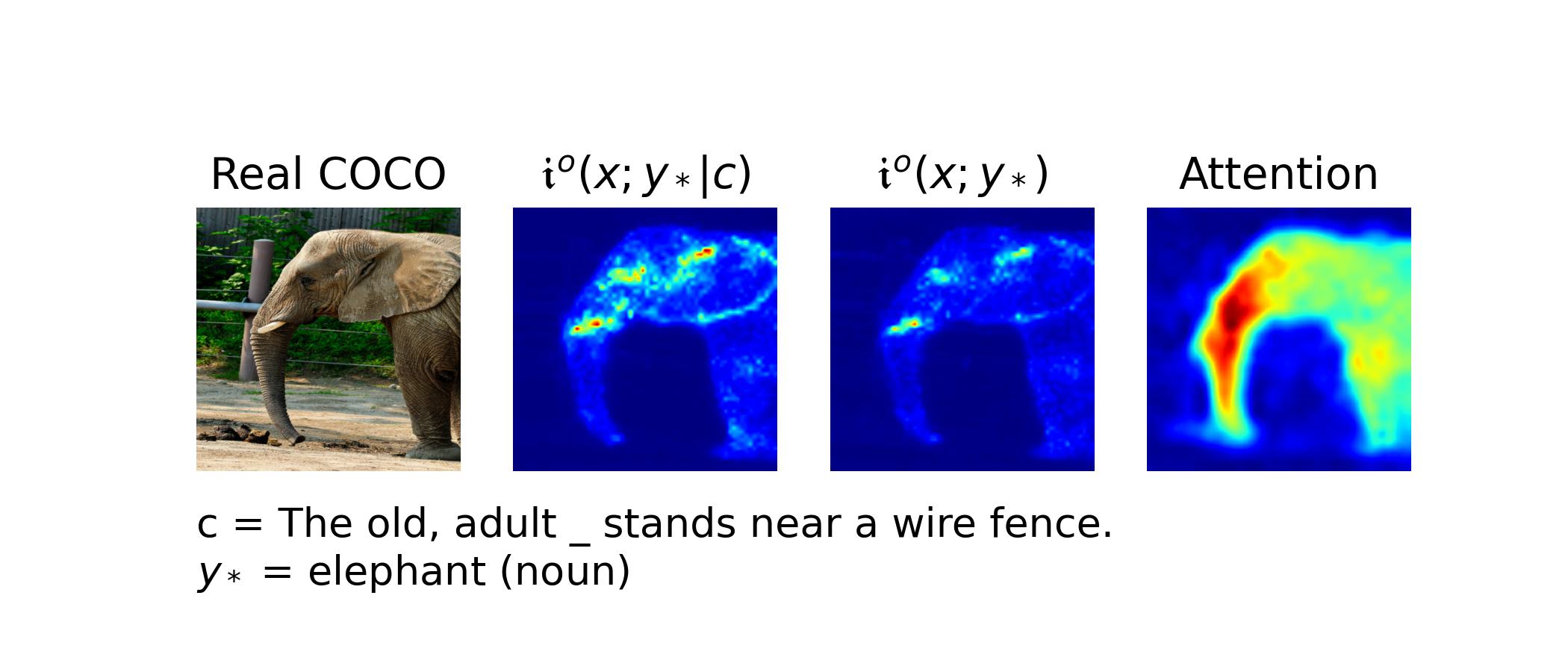} 
    \includegraphics[width=0.45\textwidth,trim={2cm 9mm 1.7cm 13mm},clip]{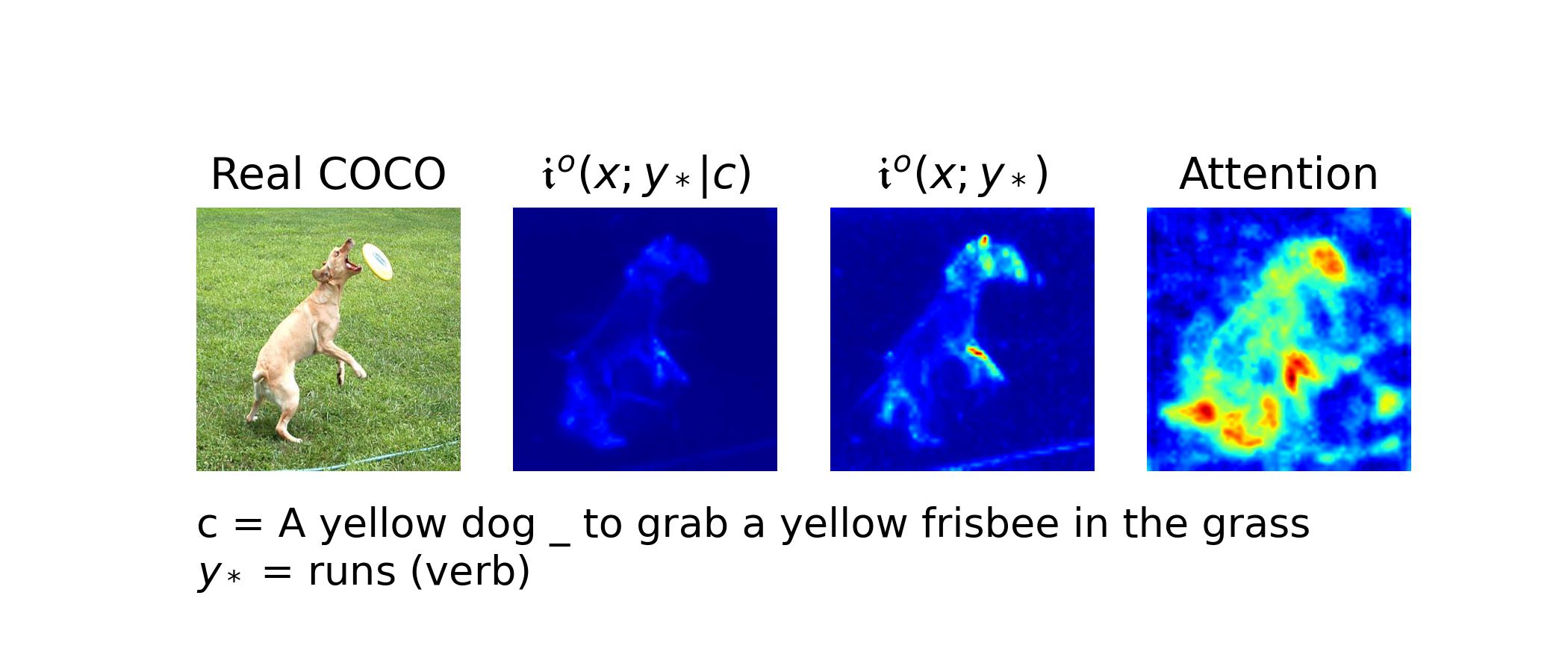} 
    \includegraphics[width=0.45\textwidth,trim={2cm 9mm 1.7cm 23mm},clip]{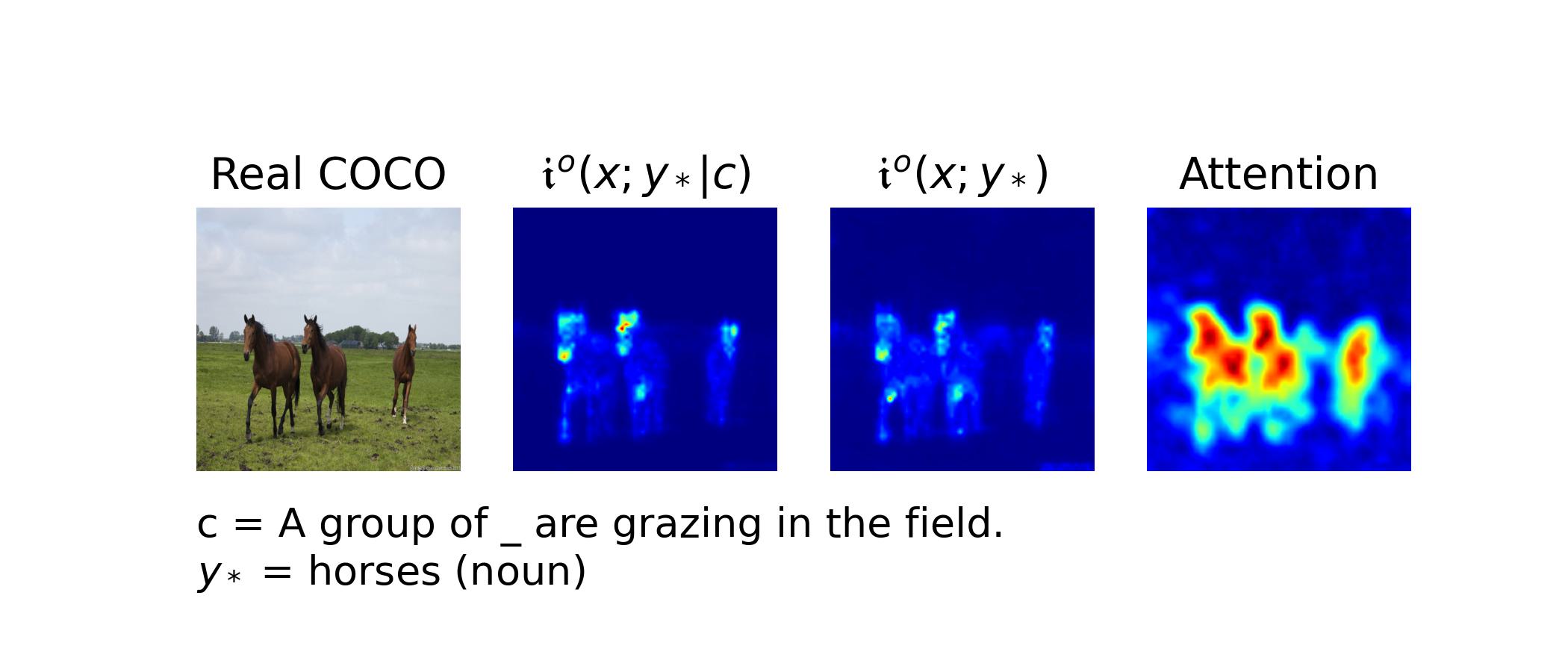} 
    \includegraphics[width=0.45\textwidth,trim={2cm 9mm 1.7cm 23mm},clip]{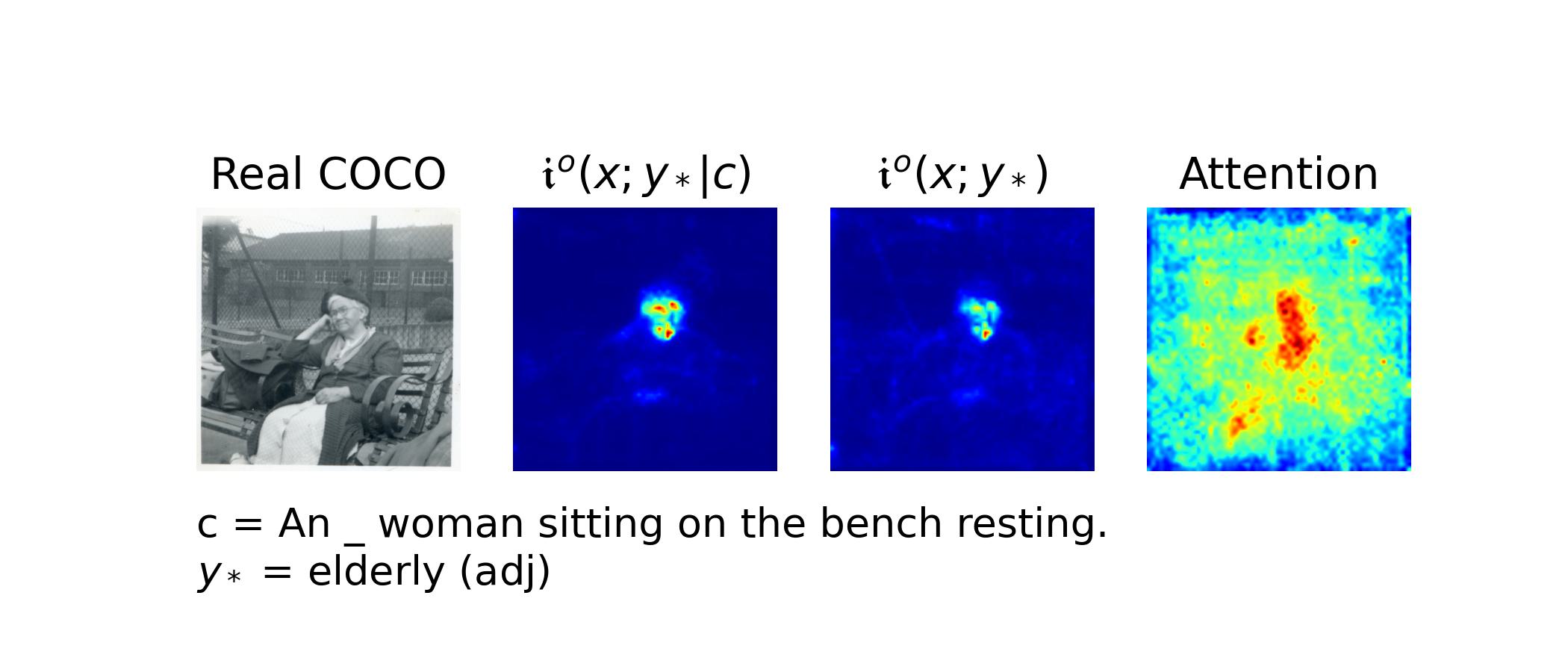} 
    \includegraphics[width=0.45\textwidth,trim={2cm 9mm 1.7cm 23mm},clip]{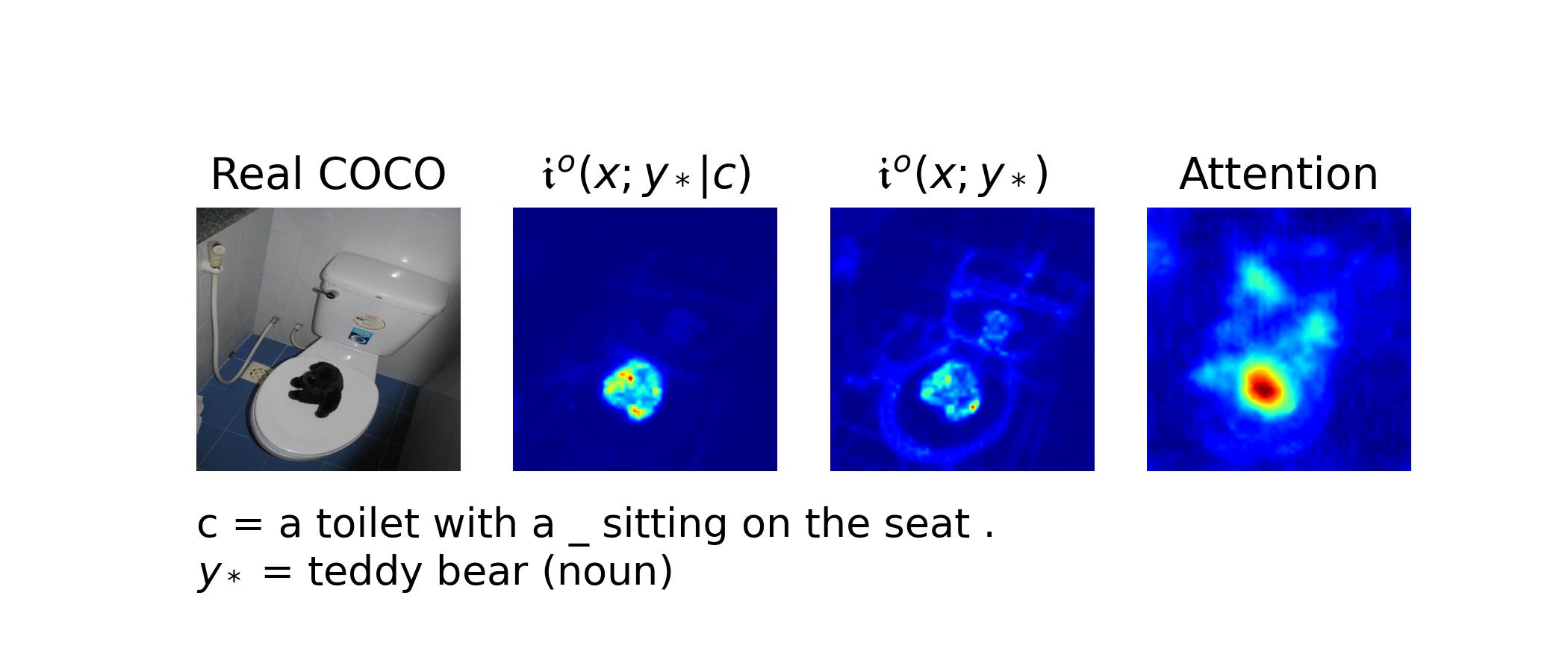} 
    \includegraphics[width=0.45\textwidth,trim={2cm 9mm 1.7cm 23mm},clip]{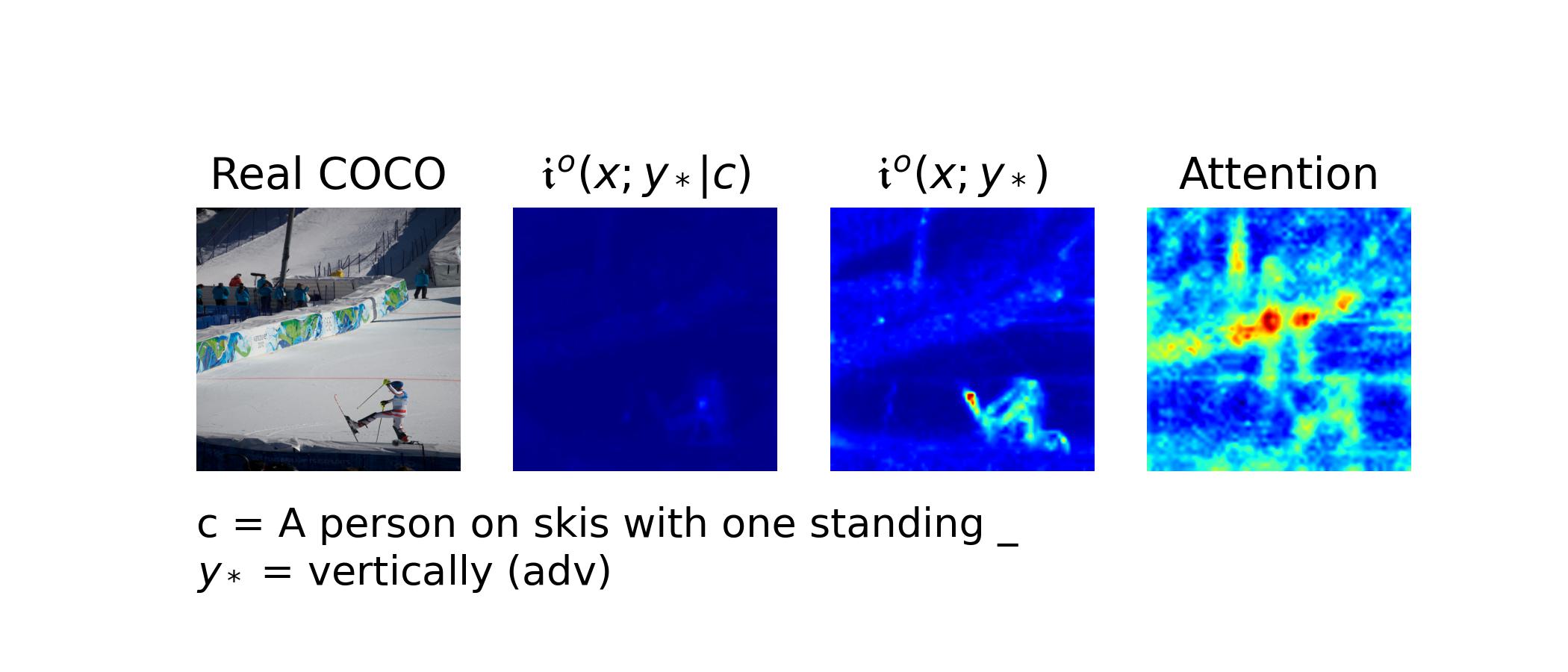} 
    \includegraphics[width=0.45\textwidth,trim={2cm 9mm 1.7cm 23mm},clip]{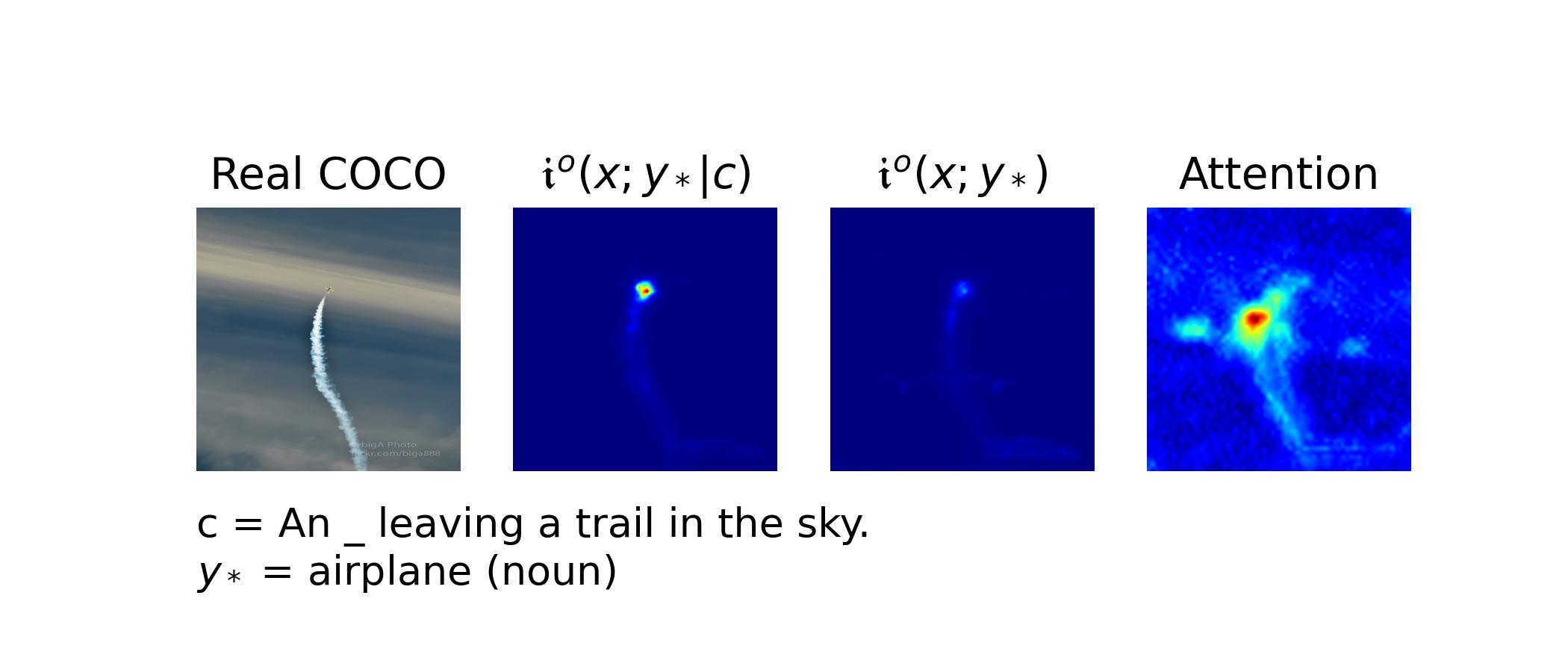} 
    \includegraphics[width=0.45\textwidth,trim={2cm 9mm 1.7cm 23mm},clip]{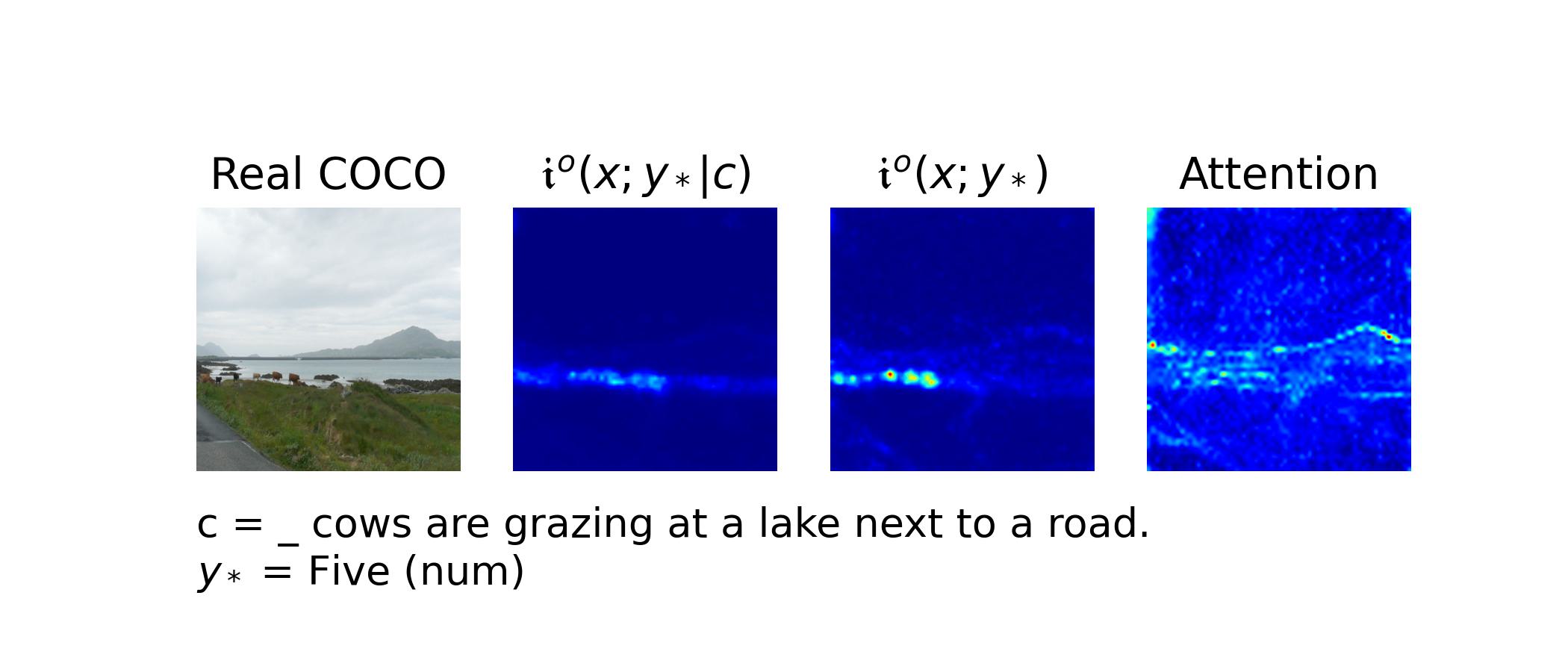} 
     \includegraphics[width=0.45\textwidth,trim={2cm 9mm 1.7cm 23mm},clip]{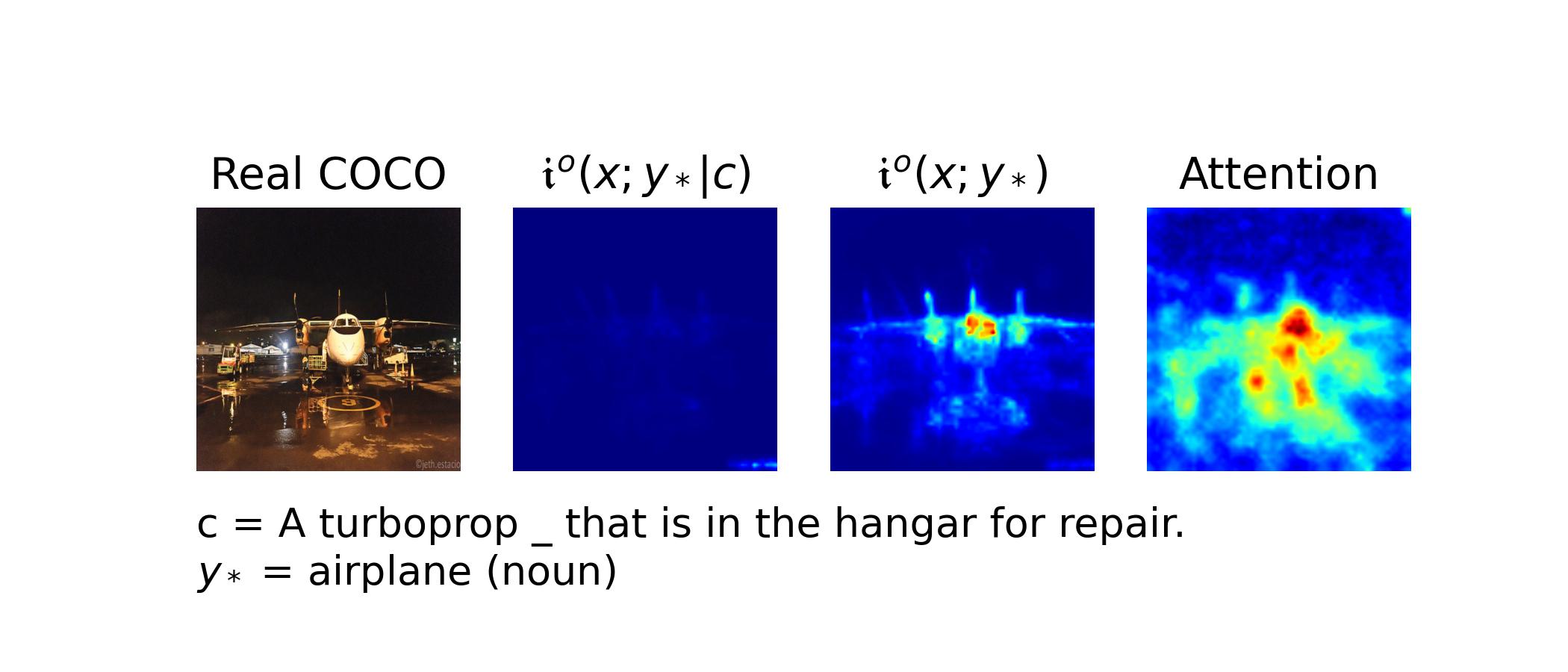} 
    \includegraphics[width=0.45\textwidth,trim={2cm 9mm 1.7cm 23mm},clip]{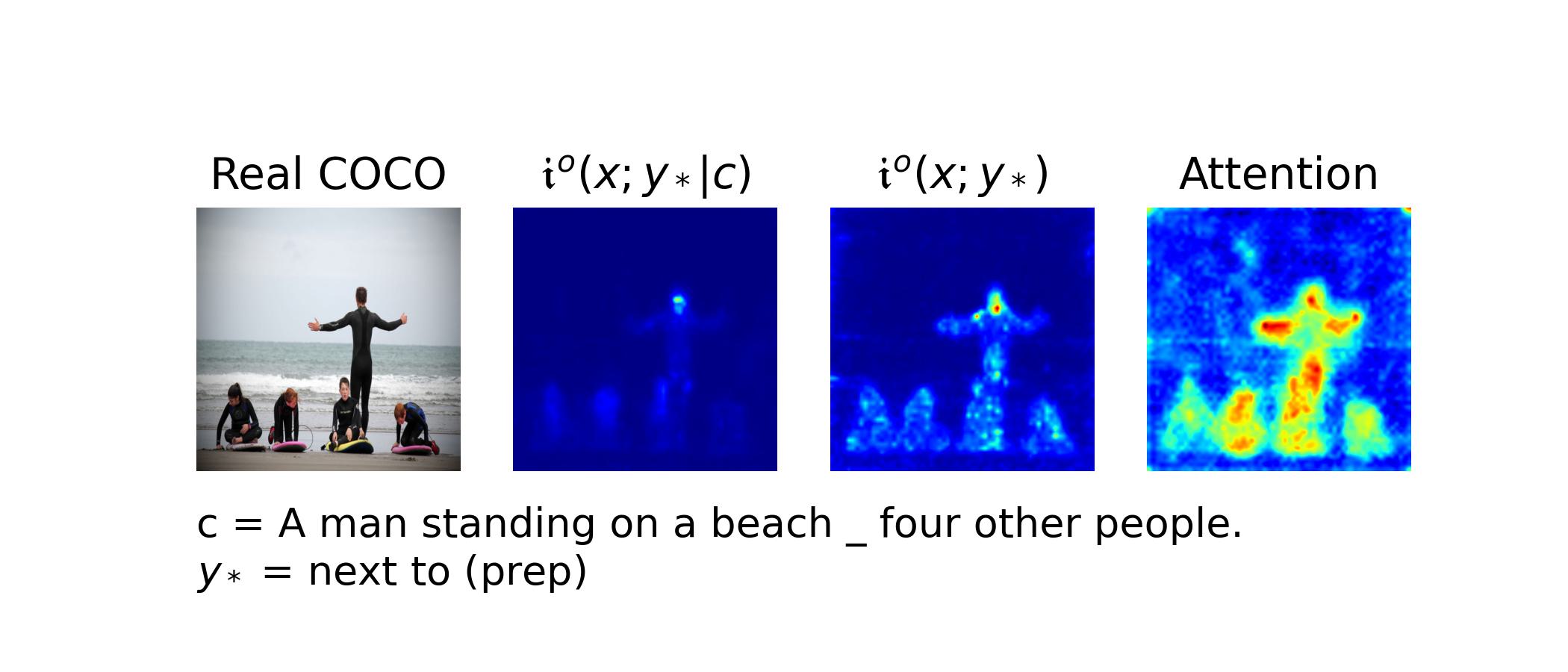} 
    \caption{Examples of localizing different types of words in images. The left half presents noun words, while the right half displays abstract words.}
    \label{fig:2d_mi_cmi_attn}
\end{figure}

Given an image-text pair $(\vx, \vy)$ where $\vy = \{y_*, \vc \}$, we compute pixel-level mutual information as $\ii^{o}_j(\vx; y_*)$, and the conditional one as $\ii^{o}_j(\vx;y_*|\vc)$ for pixel $j$, from Eq.~\ref{eq:pixel}. Eight cases are displayed in Fig. \ref{fig:2d_mi_cmi_attn}, and we put more examples in \S\ref{app:2d_mi_cmi_attn_noun} and \S\ref{app:2d_mi_cmi_attn_7}.

In Fig. \ref{fig:2d_mi_cmi_attn}, the visualizations of pixel-level MI and CMI are presented using a common color scale. Upon comparison, it becomes evident that CMI has a greater capacity to accentuate objects while simultaneously diminshing the background. This is attributed to the fact that MI solely computes the relationship between {noun} word $y_*$ and pixels, whereas CMI factors out context-related information from the complete prompt-to-pixel information.
However, there are occasional instances of the opposite effect, as observed in the example where `$y_*$ = airplane' (left bottom in Fig. \ref{fig:2d_mi_cmi_attn}). In this case, it appears that CMI fails to highlight pixels related to `airplane', while MI succeeds. This discrepancy arises from the presence of the word `turboprop' in the context. Therefore the context, $\vc$, accurately describes the content in the image and `airplane' adds no additional information. Compared to attention, CMI and MI qualitatively appear to focus more on fine details of objects, like the eyes, ears, and tusks of an elephant, or the face of horses. For object words, we can use a segmentation task to make more quantitative statements in \S\ref{sec:segment}. 

We also explore whether pixel-level CMI or MI can provide intriguing insights for other types of entities. Fig. \ref{fig:2d_mi_cmi_attn} (right) presents four different types of entities besides nouns, including verbs, adjectives, adverbs, numbers and prepositions. These words are quite abstract, and even through manual annotation, it can be challenging to precisely determine the corresponding pixels for them. The visualization results indicate that MI gives intuitively plausible results for these abstract items, especially adjectives and adverbs, that highlight relevant finer details of objects more effectively than attention. 
Interestingly, MI's ability to locate abstract words within images aligns with the findings presented in \S\ref{sec:aro} regarding relation testing (Table \ref{appendix:table-relation-results}).

\subsubsection{Localizing word information in images}\label{sec:segment}
The visualizations provided above offer an intuitive demonstration of how pixel-wise CMI relates parts of an image to parts of a caption. Therefore, our curiosity naturally extends to whether this approach can be applied to word localization within images. Currently, the prevalent evaluation involves employing attention layers within Stable Diffusion models for object segmentation \citep{tang2022daam, tian2023diffuse, wang2023diffusion}. These methodologies heavily rely on attention layers and meticulously crafted heuristic heatmap generation. It's worth highlighting that during image generation, the utilization of multi-scale cross-attention layers allows for the rapid computation of the Diffuse Attention Attribution Map (DAAM) \citep{tang2022daam}. This not only facilitates segmentation but also introduces various intriguing opportunities for word localization analyses. We opted for DAAM as our baseline choice due to its versatile applicability, and the detailed experimental design is documented in the \S\ref{app:locword}.
\begin{table}[h]
\caption{Unsupervised Object Segmentation mIoU (\%) Results on COCO-IT}
\label{tab:segment}
\begin{center}
\small
\begin{tabular}{lccc}
\hline
\multicolumn{1}{c}{\bf Method}  & \bf 50 steps & \bf 100 steps & \bf 200 steps
\\ \hline 
Whole Image Mask & 14.94 & 14.94 & 14.94\\
Attention \citep{tang2022daam} & 34.52 & 34.90 & 35.35\\
Conditional Mutual Info.            & 32.31 & 33.24 & 33.63\\
Attention+Information         & \textbf{42.46} & \textbf{42.71} & \textbf{42.84}\\
\hline
\end{tabular}
\end{center}
\end{table}
We use mean Intersection over Union (mIoU) as the evaluation metric for assessing the performance of pixel-level object segmentation. Table \ref{tab:segment} illustrates that, in contrast to attention-based methods, pixel-wise CMI proves less effective for object segmentation, with an error analysis appearing in Table \ref{tab:segment_std}. The attention mechanism in DAAM combines high and low resolution image features across the multi-scale attention layers, akin to Feature Pyramid Networks (FPN) \citep{lin2017feature}, facilitating superior feature fusion. CMI tends to focus more on the specific details that are unique to the target object rather than capturing the overall context of it. Although pixel-level CMI did not perform exceptionally well in object segmentation, the results from `Attention+Information' clearly demonstrate that the information-theoretic diffusion process \citep{kong22} does enhance the capacity of attention layers to capture features. 

\paragraph{Discussion: attention versus information versus segmentation}
 Looking at the heatmaps, it is clear that some parts of an object can be more informative than others, like faces or edges. Conversely, contextual parts of an image that are not part of an object can still be informative about the object. For instance, we may identify an airplane by its vapor trail (Fig. \ref{fig:2d_mi_cmi_attn}). Hence, we see that neither attention nor information perfectly aligns with the goal of segmentation.
 One difference between attention and mutual information is that when a pixel pays ``attention'' to a word in the prompt, it \emph{does not} imply that modifications to the word would modify the prompt. 
This is best highlighted with conditional mutual information, where we see that an informative word (``jet'') may contribute little information in a larger context (``turboprop jet''). 
To highlight this difference, we propose an experiment where we test whether intervening on words changes a generated image. Our hypothesis is that if a word has low CMI, then its omission should not change the result. For attention, on the other hand, this is not necessarily true. 
\vspace{-1mm}
\subsection{Selective Image Editing via Prompt Intervention} \label{sec:intervention}

Diffusion models have gained widespread adoption by providing non-technical users with a natural language interface to create diverse, realistic images.
Our focus so far has been on how well diffusion models understand structure in \emph{real} images.
We can connect real and generated images by studying how well we can \emph{modify} a real image, which is a popular use case for diffusion models discussed in \S\ref{sec:related}.
We can validate our ability to measure informative relationships by seeing how well the measures align with effects under prompt intervention.

For this experiment, we adopt the perspective that diffusion models are equivalent to continuous, invertible normalizing flows~\citep{diffusion_sde}. In this case, the denoising model is interpreted as a score function, and an ODE depending on this score function smoothly maps the data distribution to a Gaussian, or vice versa. The solver we use for this ODE is the 2nd order deterministic solver with 100 steps from \citet{karras}. 
We start with a real image and prompt, then use the (conditional) score model to map the image to a Gaussian latent space. In principle, this mapping is invertible, so if we reverse the dynamics we will recover the original image. We see in \S\ref{app:intervention} that the original image is almost always recovered with high fidelity. 

\begin{figure}[htbp]
    \begin{adjustbox}{valign=c} 
    \begin{minipage}[b]{0.63\columnwidth}
        \includegraphics[width=0.95\columnwidth,trim={50mm 8mm 38mm 5mm},clip]{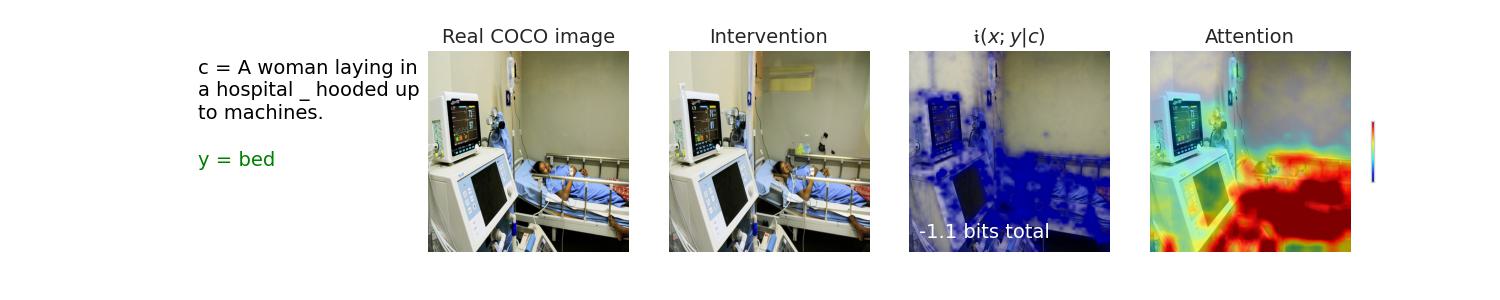} 
        \includegraphics[width=0.95\columnwidth,trim={50mm 8mm 38mm 12mm},clip]{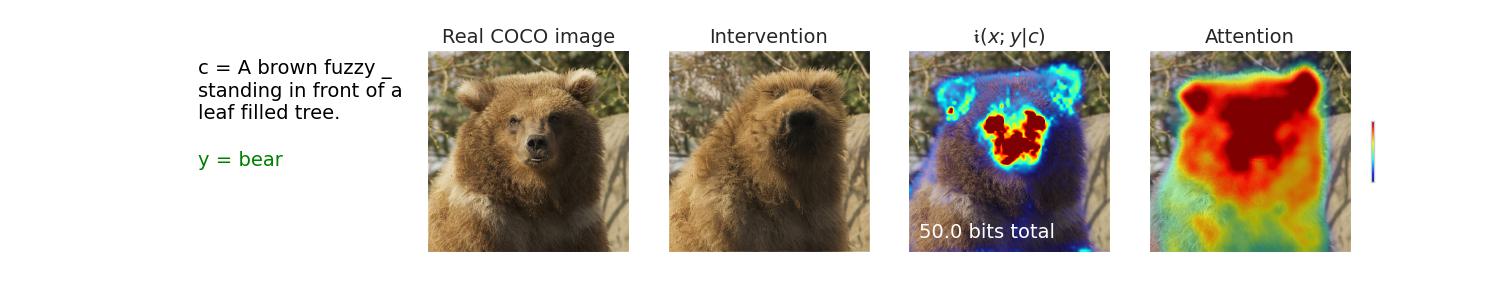} 
    \end{minipage}%
    \end{adjustbox}
    \hspace{-5mm}
    \begin{adjustbox}{valign=c} 
    \begin{minipage}[b]{0.4\columnwidth}
        \footnotesize
        \captionof{table}{Pearson Correlation with Image Change}\label{tab:intervention}
        \scalebox{0.9}{\begin{tabular}{c c c}  
             & $\ii(x;y|c)$ & Attention\\
            \hline
           Image-level & \textbf{0.34} $(\pm .010)$ & 0.24 $(\pm .011)$\\
           Pixel-level & \textbf{0.27} $(\pm .002)$ & 0.21 $(\pm .002)$
        \end{tabular}}
    \end{minipage}
    \end{adjustbox}
        \caption{COCO images edited by omitting words from the prompt. Conditional mutual information better reflects the actual changes in the image after intervention. Table shows Pearson correlation between metrics (CMI or attention) versus L2 changes in image after intervention, at the image-level and pixel-level as discussed in \S\ref{sec:intervention}.}    \label{fig:plot_table}
        \vspace{-3mm}
\end{figure}

Next, we consider adding an intervention. While doing the reverse dynamics, we \emph{modify} the (conditional) score by changing the denoiser prompt in some way. We focus in experiments on the effects of omitting a word, or swapping a word for a categorically similar word (``bear'' $\rightarrow$ ``elephant'', for example). Typically, we find that much of the detail in the original image is preserved, with only the parts that relate to the modified word altered.  Interestingly, we also find that in some cases interventions to a word are seemingly ignored (see Fig.~\ref{fig:abstract}, \ref{fig:plot_table}). 

We want to explore whether attention or information measures predict the effect of interventions. When intervening by omitting a word, we consider the conditional (pointwise) mutual information $\ii(\vx;y|\vc)$ from \eqref{eq:pointwise}, and for word swaps we use a difference of CMIs with $y, y'$ representing the swapped word, as in Fig.~\ref{fig:abstract}. For attention, we aggregate the attention corresponding to a certain word during generation using the code from~\citet{tang2022daam}. 
We find that a word can be ignored \emph{even if the attention heatmap highlights an object related to the word}.  In other words, attention to a word does not imply that it affects the generated outcome. One reason for this is that a word may provide little information beyond that in the context. For instance, in Fig.~\ref{fig:plot_table}, a woman in a hospital is assumed to be on a bed, so omitting this word has no effect. CMI, on the other hand, correlates well with the effect of intervention, and ``bed'' in this example has low CMI. 

To quantify our observation that conditional mutual information better correlates with the effect of intervention than attention models, we measure the Pearson correlation between a score (CMI or attention heatmap) and the effect of the intervention using L2 distance between images before and after intervention. 
To get an image-level correlation, we correlated the aggregate scores per image across examples on COCO100-IT. We also consider the pixel-level correlations between L2 change per pixel and metrics, CMI and attention heatmaps. We average these correlations over all images and report the results in Table ~\ref{tab:intervention}.  
CMI is a much better predictor of changes at the image-level, which reflects the fact that it directly quantifies what additional information a word contributes after taking context into account. At the per-pixel level, CMI and attention typically correlated very well when changes were localized, but both performed poorly in cases where a small prompt change led to a global change in the image. 
Results visualizing this effect are shown along with additional experiments with word swap interventions in \S\ref{app:intervention}. Small dependence, as measured by information, correctly implied small effects from intervention. Large dependence, however, can lead to complex, global changes due to the nonlinearity of the generative process.  

\section{Related Work}\label{sec:related}

\textbf{Visual perception via diffusion models}: Diffusion models' success in image generation has piqued interest in their text-image understanding abilities. A variety of pipeline designs are emerging in the realm of visual perception and caption comprehension, offering fresh perspectives and complexities. Notably, ODISE introduced a pipeline that combines diffusion models with discriminative models, achieving excellent results in open-vocabulary segmentation \citep{xu2023openvocabulary}. Similarly, OVDiff uses diffusion models to sample support images for specific textual categories and subsequently extracts foreground and background features \citep{karazija2023diffusion}. It then employs cosine similarity, often known as the CLIP \citep{radford2021learning} filter, to segment objects effectively. MaskDiff \citep{le2023maskdiff} and DifFSS \citep{tan2023diffss} introduce a new breed of conditional diffusion models tailored for few-shot segmentation. DDP, on the other hand, takes the concept of conditional diffusion models and applies it to dense visual prediction tasks \citep{ji2023ddp}. VDP incorporates diffusion for a range of downstream visual perception tasks such as semantic segmentation and depth estimation \citep{zhao2023unleashing}. Several methods have begun to explore the utilization of attention layers in diffusion models for understanding text-image relationships in object segmentation \citep{tang2022daam, tian2023diffuse, wang2023diffusion, zhang2023diffusionengine, ma2023diffusionseg}.
Diffusion models have also been explored as a way to generate synthetic datasets for segmentation \citep{ge2023beyond}.

\textbf{Image editing via diffusion}: Editing real images is an area of growing interest both academically and commercially with approaches using natural text, similar images, and latent space modifications to achieve desired effects~\citep{glide,mao2023guided,concept_discovery,balaji2022ediffi,imagic,mokady2023null}. \citet{ddib}'s approach most closely resembles the procedure used in our intervention experiments. Unlike previous work, we focus not on edit quality but on using edits to validate the learned dependencies as measured by our information estimators.

\textbf{Interpretable ML}: Traditional methods for attribution based on gradient sensitivity rather than attention~\citep{integrated_gradients,gradient_shap} are seldom used in computer vision due to the well-known phenomenon that (adversarial) perturbations based on gradients are uncorrelated with human perception~\citep{intriguing}. 
Information-theoretic approaches for interpretability based on information decomposition~\citep{williamsbeer} are mostly unexplored because no canonical approach exists~\citep{kolchinsky2022novel} and existing approaches are completely intractable for high-dimensional use cases~\citep{reing_aistats21}, though there are some recent attempts to decompose \emph{redundant information} with neural networks \citep{rine,factorizedcl}. Our decomposition decomposes an information contribution from each variable, but does not explicitly separate unique and redundant components. ``Interpretability'' in machine learning is a fuzzy concept which should be treated with caution~\citep{mythos}. We adopted an \emph{operational interpretation} from information theory, which considers $\vy \rightarrow \vx$ as a noisy channel and asks how many bits of information are communicated, using diffusion models to characterize the channel. 

\vspace{-1mm}
\section{Conclusion}

The eye-catching image generation capabilities of diffusion models have overshadowed the equally important and underutilized fact that they also represent the state-of-the-art in density modeling~\citep{vdm}. 
Using the tight link between diffusion and information, we were able to introduce a novel and tractable information decomposition.  This significantly expands the usefulness of neural information estimators~\citep{belghazi2018mine,poole2019variational,brekelmans2023improving}
by giving an interpretable measure of fine-grained relationships at the level of individual samples and variables.
While we focused on vision tasks for ease of presentation and validation, information decomposition can be particularly valuable in biomedical applications like gene expression where we want to identify informative relationships for further study~\citep{pepke}.
Another promising application relates to contemporaneous works on \emph{mechanistic interpretability} \citep[\textit{inter alia}]{wang2022interpretability, hanna2023does} which seeks to identify ``circuits"---a subgraph of a neural network responsible for certain behaviors---by ablating individual network components to observe performance differences. For language models, differences are typically measured as the change in total probability of interested vocabularies; whereas metric design remains an open question for diffusion models. Our analyses indicate that the CMI estimators are apt for capturing compositional understanding and localizing image edits. For future work, we are interested in exploring their potential as metric candidates for identifying relevant circuits in diffusion models. 

\section{Ethics Statement}

Recent developments in internet-scale image-text datasets have raised substantial concerns over their lack of privacy, stereotypical representation of people, and political bias \citep[\textit{inter alia}]{birhane2021multimodal, peng2021mitigating, schuhmann2021laion}. Dataset contamination entails considerable safety ramifications. \underline{First}, models trained on these datasets are susceptible to memorizing similar pitfalls. \underline{Second}, complex text-to-image generation models pose significant challenges to interpretability and system monitoring \citep{hendrycks2023overview}, especially given the increasing popularity in black-box access to these models \citep{dalle2}. These risks are exacerbated by these models' capability to synthesize photorealistic images, forming an ``echo chamber" that reinforces existing biases in dataset collection pipelines. We, as researchers, shoulder the responsibility to analyze, monitor, and prevent the risks of such systems at scale.

We believe the application of our work has meaningful ethical implications. Our work aims to characterize the fine-grained relationship between image and text. Although we primarily conduct our study on entities that do not explicitly entail societal implications, our approach can be conceivably adapted to attribute prompt spans responsible for generating images that amplify demographic stereotypes \citep{bianchi2023easily}, among many other potential risks. Aside from detecting known risks and biases, we should aim to design approaches that automatically identify system anomalies. In our image generation experiments, we observe changes in mutual information to be inconsistent during prompt intervention. It is tempting to hypothesize that these differences may be reflective of dataset idiosyncrasies. We hope that our estimator can contribute to a growing body of work that safeguards AI systems in high-stake application scenarios \citep{barrett2023identifying}, as diffusion models could extrapolate beyond existing text-to-image generation to sensitive domains such as protein design \citep{watson2023novo}, molecular structure discovery \citep{igashov2022equivariant}, and interactive decision making \citep{chi2023diffusion}.

Finally, it is imperative to address the ethical concerns associated with the use of low-wage labor for dataset annotation and evaluation \citep{Perrigo_2023}. Notably, existing benchmarks for assessing semantic understanding of image generation resort to manual evaluation \citep{conwell2022testing, liu2022compositional, imagen}. In \S\ref{sec:aro} we adapt our estimator for compositional understanding, and aim to develop an automated metric for evaluation on a broader spectrum of tasks.

\subsubsection*{Acknowledgments}
GV thanks participants of the DEMICS workshop hosted at MPI Dresden for valuable feedback on this project.

\bibliography{iclr2024_conference}
\bibliographystyle{iclr2024_conference}

\newpage
\appendix
\section{Derivations of the negative log-likelihood}
\label{app:proof}
For completeness, we provide a derivation for the negative log-likelihood \ref{eq:nll} from \citep{kong22}. We first introduce a seminal result from \citep{guo},
 \begin{equation}\label{eq:immse} 
 \ds I(\vx; \vz) = \half \mmse(\snr).
\end{equation}
This relationship admits a point-wise generalization,
\begin{equation}\label{eq:pimmse} 
 \ds \kl{p(\vz|\vx)}{p(\vz)}  = \half \mmse(\vx, \snr),
\end{equation}
The marginal is $p(\vz) = \int p(\vz|\vx) p(\vx) d\vx$, and the pointwise MMSE is defined as follows,
\begin{equation}\label{eq:pmmse} 
\mmse(\vx, \snr) \equiv \mathbb E_{p(\vz|\vx)} \big[ 
\norm{\vx - \xhat^*(\vz, \snr)} 
\big] .
\end{equation}
To obtain the desired result, we apply the thermodynamic integration trick introduced in \citep{vdm}, by first defining the point-wise gap function $f(\vx, \snr)$ as
$$
f(\vx, \snr) \equiv \kl{p(\vz|\vx)}{p_G(\vz)} - \kl{p(\vz|\vx)}{p(\vz)}.
$$
We denote $p_G(\vz) = \int p(\vz|\vx) p_G(\vx) d\vx$ as the marginal output distribution of the MMSE for the channel with Gaussian input as $\mmse_G(\snr)$. 
In the limit of zero SNR, we get $\lim_{\snr \rightarrow 0} f(\vx, \snr) = 0$. In the high SNR limit, \citep{kong22} prove that
\begin{equation}\label{eq:log_ratio}
\lim_{\snr \rightarrow \infty} f(\vx, \snr) = \log \frac{p(\vx)}{p_G(\vx)}.
\end{equation}
Combining this with ~\eqref{eq:pimmse}, we can write the log likelihood \emph{exactly} in terms of the log likelihood of a Gaussian and a one dimensional integral. 
\begin{align}\label{eq:density}
-\log p(\vx) &= -\log p_G(\vx) - \int_{0}^{\infty} d\snr \ds f(\vx, \snr) \nonumber \\
&= { -\log p_G(\vx)} - \half \int_{0}^{\infty} d\snr \left( {\mmse_G(\vx, \snr)} - \mmse(\vx, \snr) \right) 
\end{align}

This expresses density in terms of a Gaussian density and a correction that measures how much better we can denoise the target distribution than we could using the optimal decoder for Gaussian source data. 
The density can be further simplified by writing out the Gaussian expressions explicitly and simplifying with an identity given in,
  \begin{equation}\label{eq:density_simple} 
-\log p(\vx) = d/2 \log(2 \pi e) - \half \int_{0}^{\infty} d\snr \left(  \frac{d}{1+\snr} - \mmse(\vx, \snr) \right).
  \end{equation}
Observe that the first term in the integrand does not depend on $\vx$, which allows us to derive the desired result Eq. \ref{eq:nll}. We refer readers to \citep{kong22} for more detailed derivations.

\section{Derivations of pointwise information via the orthogonality principle }\label{sec:orthogonality}

Our goal is to show that the following expression,
$$\ii^o(\vx; \vy) \equiv \half \int \E_{p(\eps)} \left[ \norm{\epshat(\vz) - \epshat(\vz|\vy)} \right] d\logsnr,$$
is a pointwise information estimator, i.e., that it satisfies the identity,
$$I(X;Y) = \E_{p(\vx, \vy)} [ \ii^o(\vx; \vy) ] .$$
To show this fact, we first recall the definition of our optimal denoiser and optimal conditional denoiser. 
\begin{align*}
\epshat(\vx) &\equiv \arg \min_{\epsbar(\cdot)} \E_{p(\vx), p(\eps)} \left[ \norm{\eps - \epsbar(\vz)} \right] \\
\epshat(\vx|\vy) &\equiv \arg \min_{\epsbar(\cdot)} \E_{p(\vx|\vy), p(\eps)} \left[ \norm{\eps - \epsbar(\vz|\vy)} \right]
\end{align*}
For this optimal denoiser, the following expression holds exactly.
\be
-\log p(\vx) &= \half \int \E_{p(\eps)} \left[ \norm{\eps - \epshat(\vz)} \right] d\logsnr + const \\
-\log p(\vx|\vy) &= \half \int \E_{p(\eps)} \left[ \norm{\eps - \epshat(\vz|\vy)} \right] d\logsnr + const
\ee
Therefore, we have that, 
\begin{align*}
I(X;Y) &= \E_{p(\vx, \vy)} [ \log p(\vx|\vy) - \log p(\vx ] \\
&= \E_{p(\vx, \vy)} \Big[ \half \int \E_{p(\eps)} \left[ \norm{\eps - \epshat(\vz)} - \norm{\eps - \epshat(\vz|\vy)} \right] d\logsnr \Big]
\end{align*}
Rearranging we have the following. 
\begin{align*}
I(X;Y) 
&= \E_{p(\vx, \vy)} \Big[ \overbrace{\half \int \E_{p(\eps)} \left[ \norm{\epshat(\vz) - \epshat(\vz|\vy)} \right] d\logsnr}^{\ii^o(\vx; \vy) } \Big]  \\
& \qquad \qquad + 2 \E_{p(\vy)} \Big[ \half \int \underbrace{ \color{red} \E_{p(\vx | \vy),p(\eps)} \left[ (\epshat(\vz) - \epshat(\vz|\vy)) \cdot (\epshat(\vz|\vy) - \eps) \right]  }_{\equiv \mathcal{O}} d\logsnr \Big]
\end{align*}
What remains is to show that the term in red is zero, $\mathcal{O} = 0$, and therefore the whole second term is equal to zero. 
This fact follows from the orthogonality principle~\citep{kay1993fundamentals}, which states the slightly more general result that,
$$\forall \vf, ~~ \E_{p(\vx | \vy) p(\eps)} [ \vf(\vz,\vy) \cdot (\epshat(\vz|\vy) - \eps)] = 0.$$ 
Note that this is stated in a slightly different way, as we have used $\vz \equiv \vz(\vx, \eps)$ to write the noisy channel that our MMSE estimator is attempting to use to recover $\eps$. 
The term $(\epshat(\vz|\vy) - \eps)$ is recognized as the error of the MMSE estimator. This error must be orthogonal to any estimator, $\vf$. If it isn't, then we can use it to build an estimator with lower MSE than $\epshat(\vz|\vy)$, contradicting our assumption that $\epshat(\vz|\vy)$ is the MMSE estimator. 
A similar result to the orthogonality principle can be shown in a more general way using Bregman divergences~\citep{banerjee2005clustering}.

Therefore, we finally have the desired result that $I(X;Y) = \E_{p(\vx, \vy)} [ \ii^o(\vx; \vy) ]$. Note that this pointwise estimator has a slightly different interpretation from the standard one, $\ii^s$, as it is not equal to a log-likelihood ratio pointwise, though it is still in expectation. On the other hand, it has several nice properties. It is non-negative, which is convenient for visualizing heatmaps. It is clear that if mutual information is zero, then the optimal denoiser should learn to ignore $\vy$, so $\epshat(\vz|\vy) = \epshat(\vz)$ and our information estimate is then zero.

\section{Additional Results}
\subsection{Relationship between image-level MI and CMI}
\label{app:rela_mi_cmi}
\begin{figure}[h]
	\centering
	\subfigure{
		\begin{minipage}[b]{0.46\textwidth}
		\includegraphics[width=1\textwidth, trim=0cm 0cm 1cm 0cm, clip]{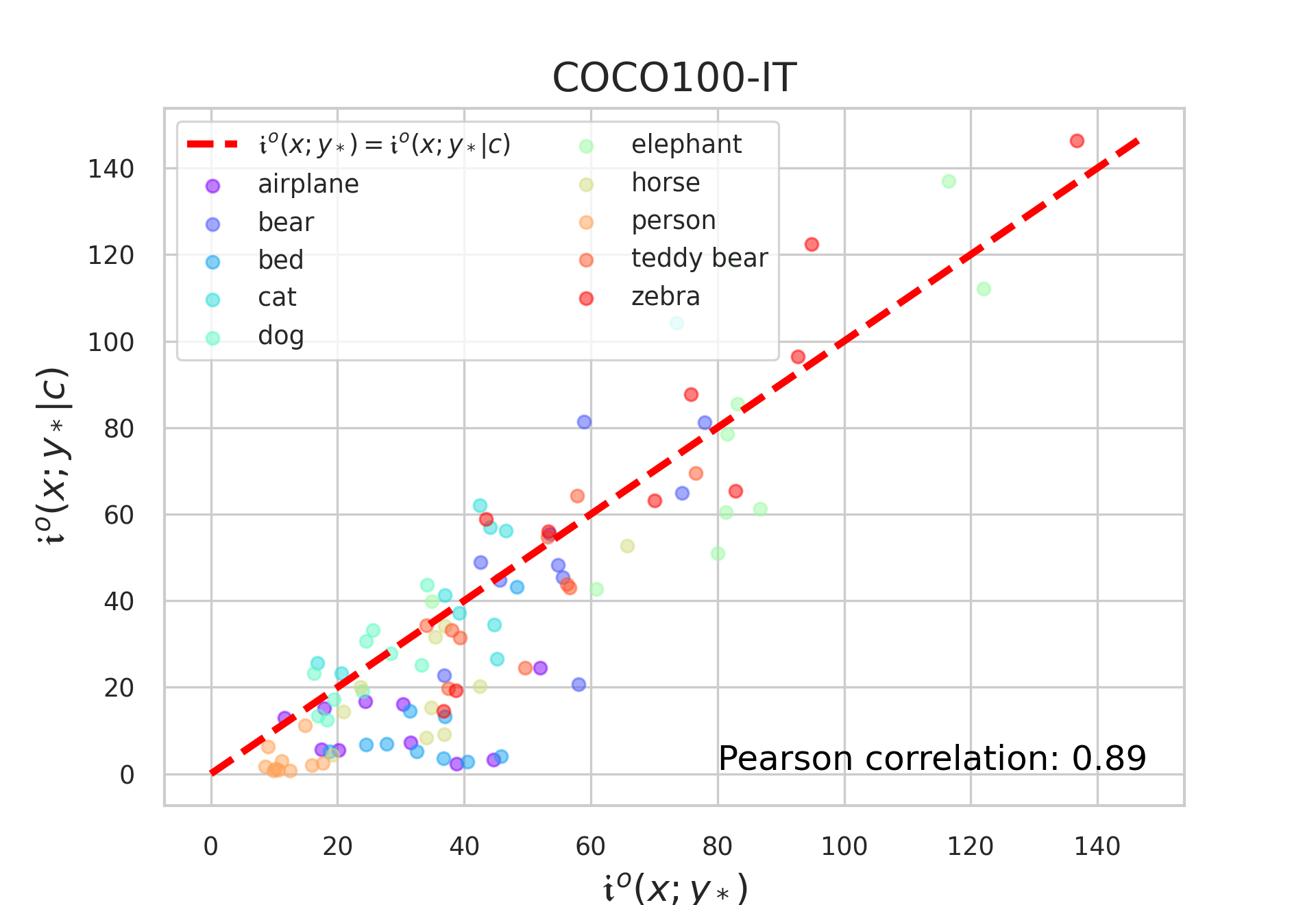}
		\end{minipage}
	}
    	\subfigure{
    		\begin{minipage}[b]{0.46\textwidth}
   		 \includegraphics[width=1\textwidth, trim=0cm 0cm 1cm 0cm, clip]{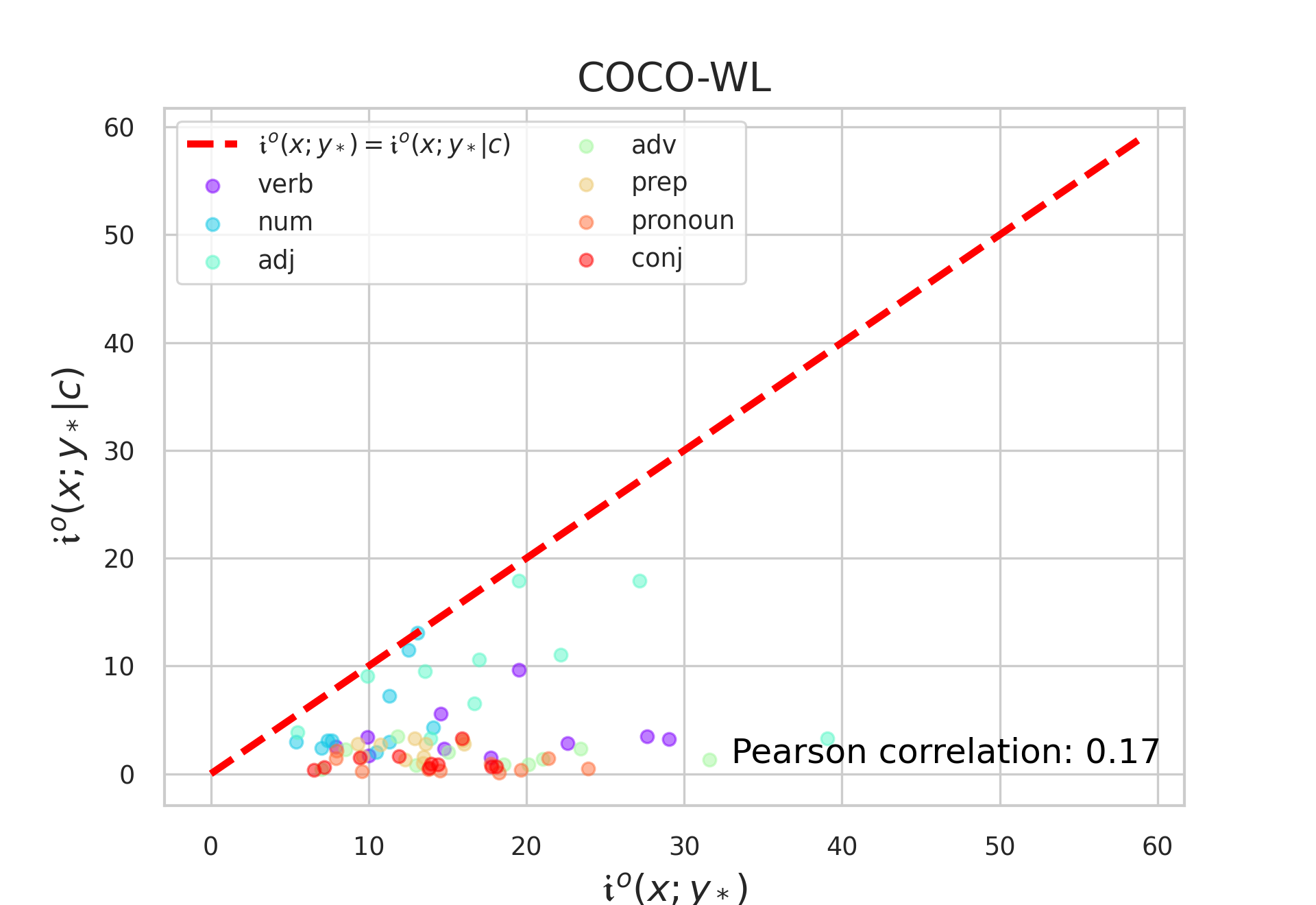}
    		\end{minipage}
    	}
	\caption{Scatter plot for correlation between MI and CMI.}
	\label{fig:scatter}
\end{figure}

On both the COCO100-IT and COCO-WL datasets, we conducted further calculations for image-level MI and CMI, presenting the results in scatterplots in Fig. \ref{fig:scatter}. These quantitative findings align with our pixel-level visual analysis (\S\ref{sec:word_loc}). MI and CMI exhibit strong consistency for noun words, with a high Pearson correlation coefficient of 0.89. In most cases, MI values remain higher than CMI, primarily due to MI containing more information from the background context. However, for abstract words, the Pearson coefficient drops to 0.17, and notably, MI is consistently larger than CMI (with most cases being nearly zero), indicating MI's superior capability in capturing information involving abstract words compared to CMI. This signals a high degree of redundancy between abstract words and context~\citep{williamsbeer}.

\subsection{Image-level MMSE curves and pixel-level MMSE visualizaiton}
\label{app:MMSE}
\begin{figure}[h]
	\centering
	\includegraphics[width=0.99\textwidth, trim=5cm 1cm 5cm 1cm, clip]{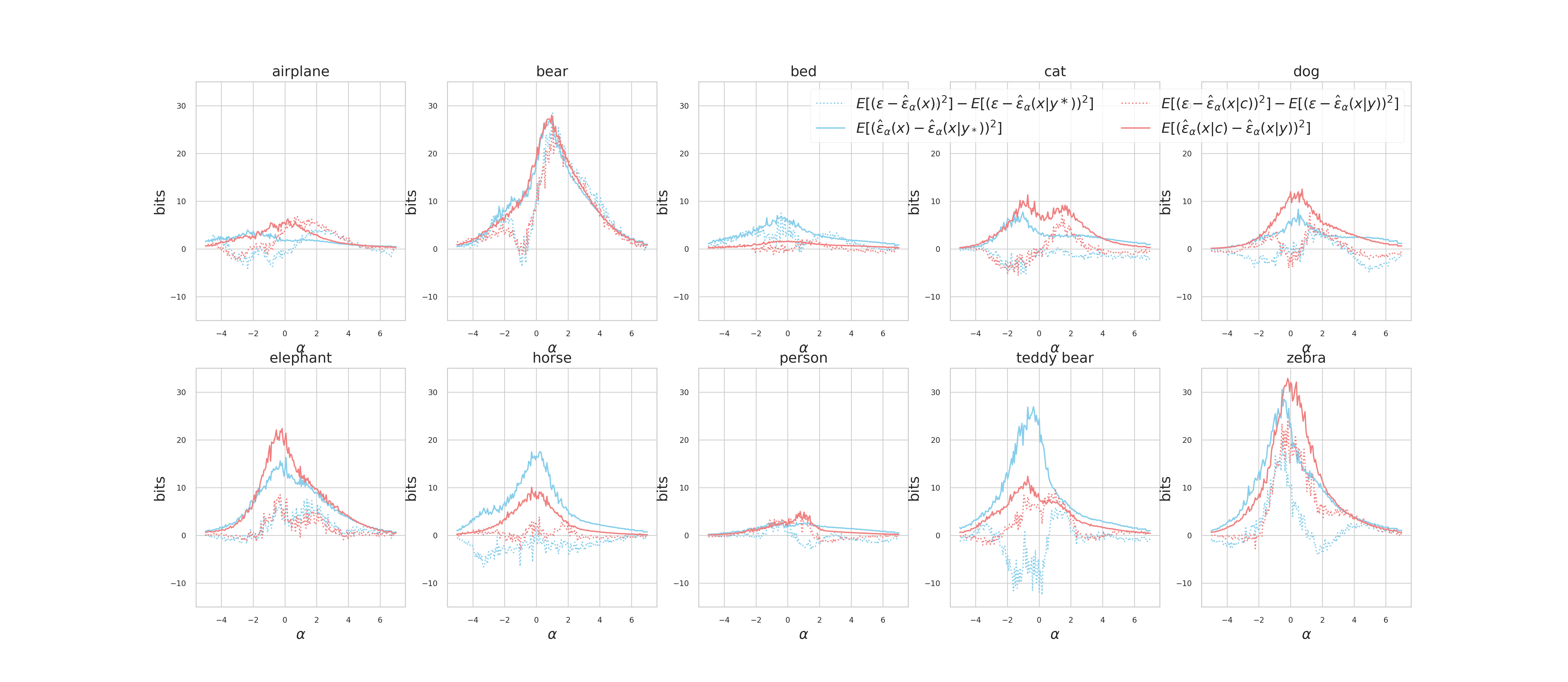}
	\caption{MMSE curves examples for 10 categories.}
	\label{fig:mmses}
\end{figure}

\begin{figure}[t]
    \centering
    \includegraphics[width=0.99\textwidth,trim={3.8cm 9mm 2.8cm 1cm},clip]{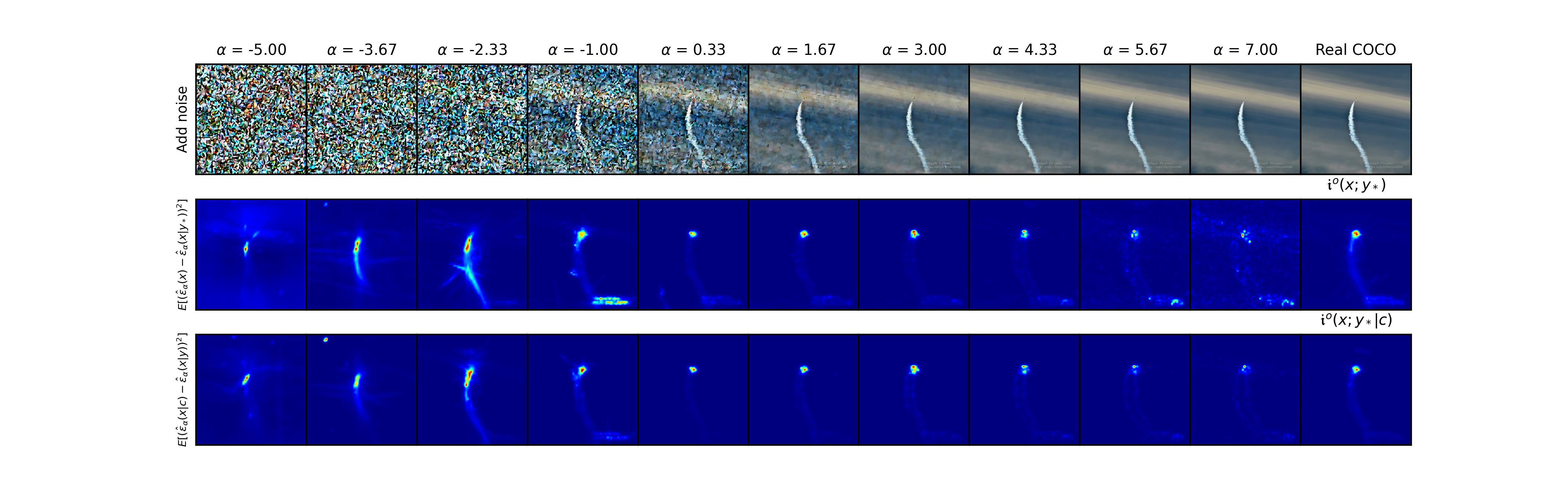}
    \includegraphics[width=0.99\textwidth,trim={3.8cm 9mm 2.8cm 1cm},clip]{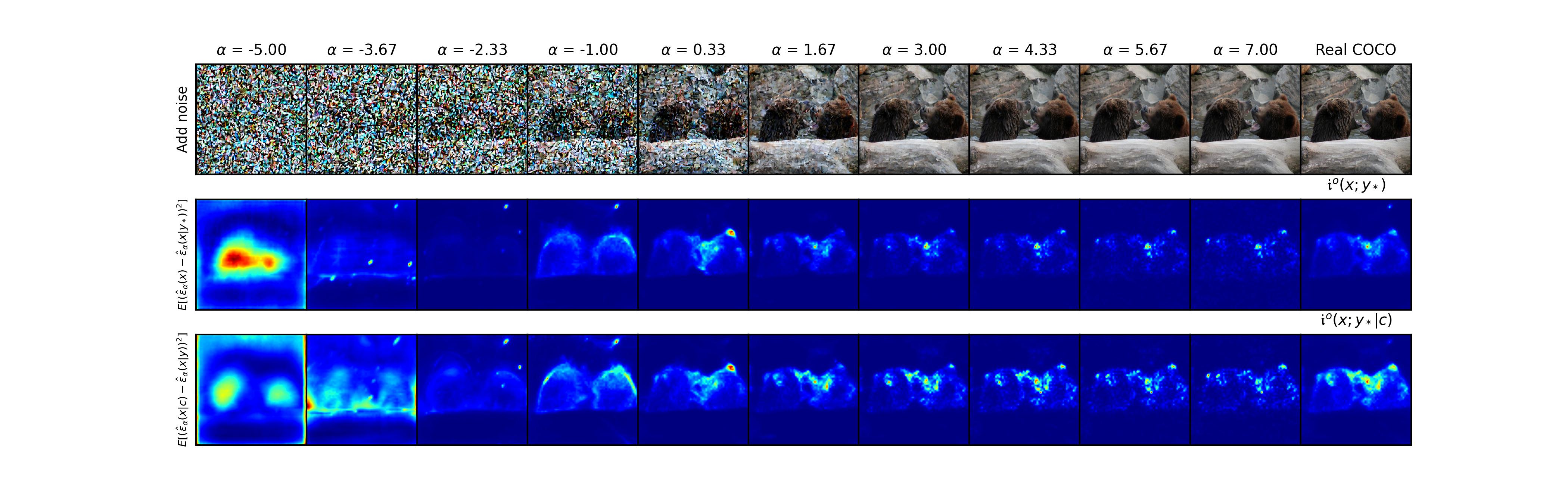}
    \includegraphics[width=0.99\textwidth,trim={3.8cm 9mm 2.8cm 1cm},clip]{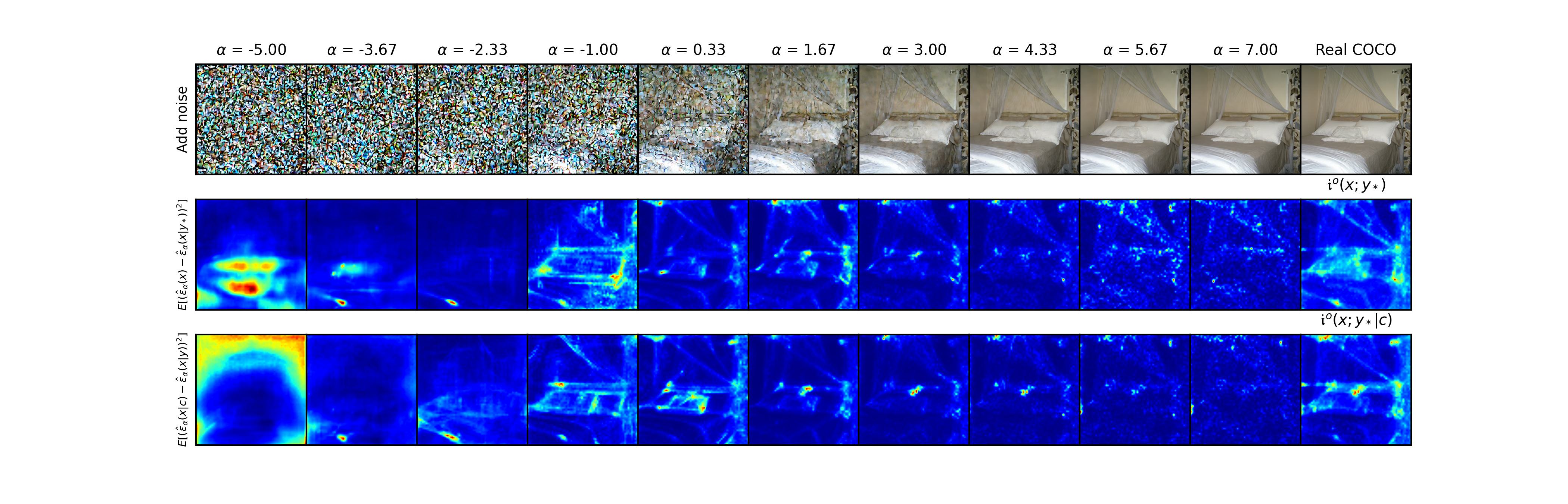}
    \includegraphics[width=0.99\textwidth,trim={3.8cm 9mm 2.8cm 1cm},clip]{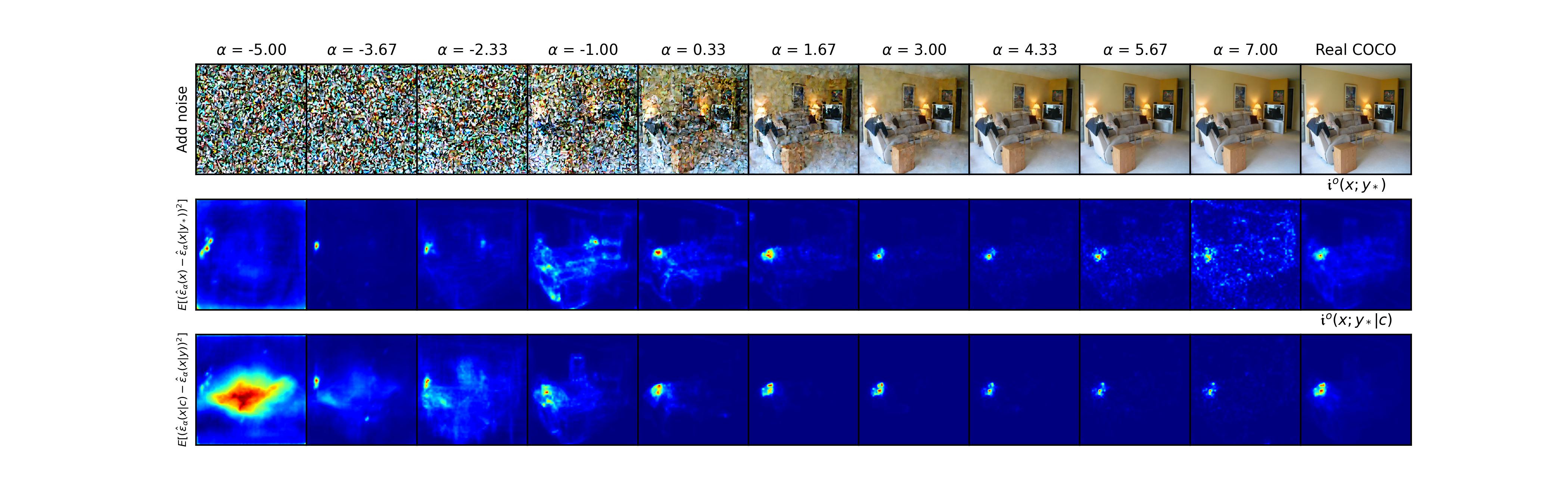}
    \includegraphics[width=0.99\textwidth,trim={3.8cm 9mm 2.8cm 1cm},clip]{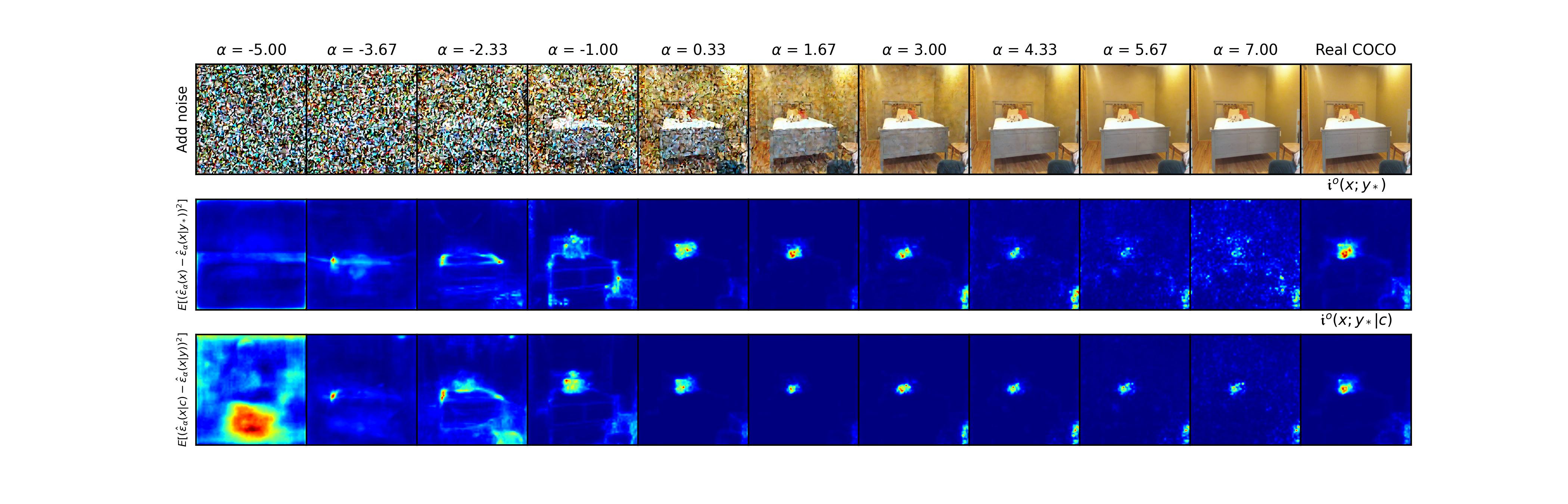}
    \caption{Examples of pixel-level MMSE visualization.}
    \label{fig:mmse-2D-1}
\end{figure}

\begin{figure}[t]
    \centering
    \includegraphics[width=0.99\textwidth,trim={3.8cm 9mm 2.8cm 1cm},clip]{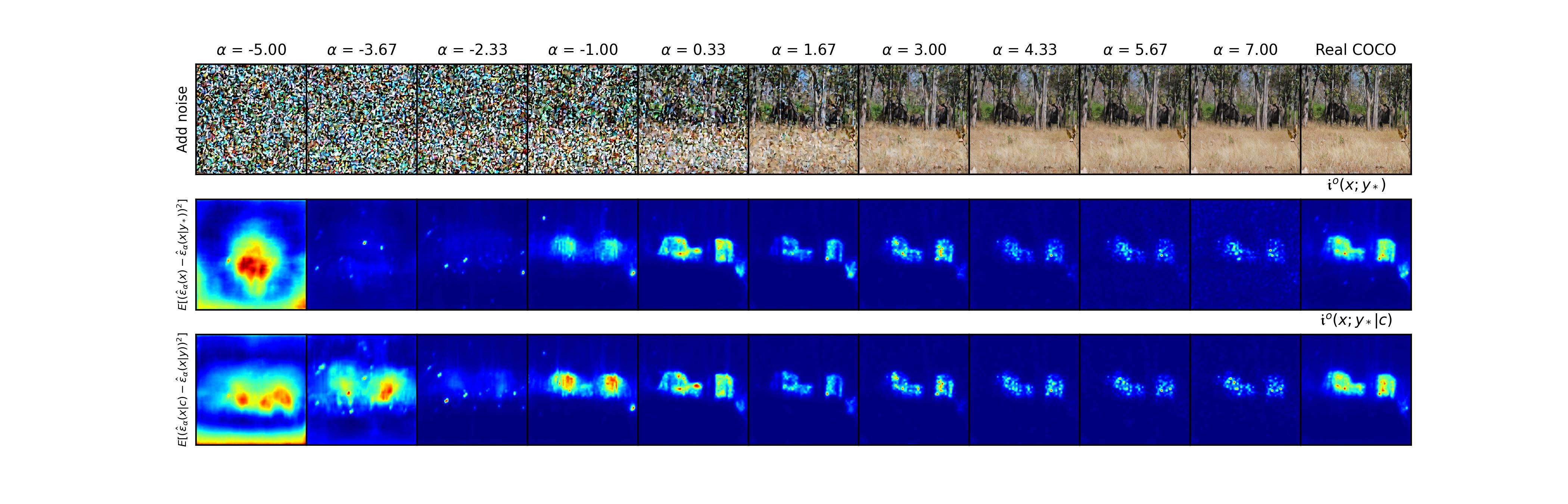}
    \includegraphics[width=0.99\textwidth,trim={3.8cm 9mm 2.8cm 1cm},clip]{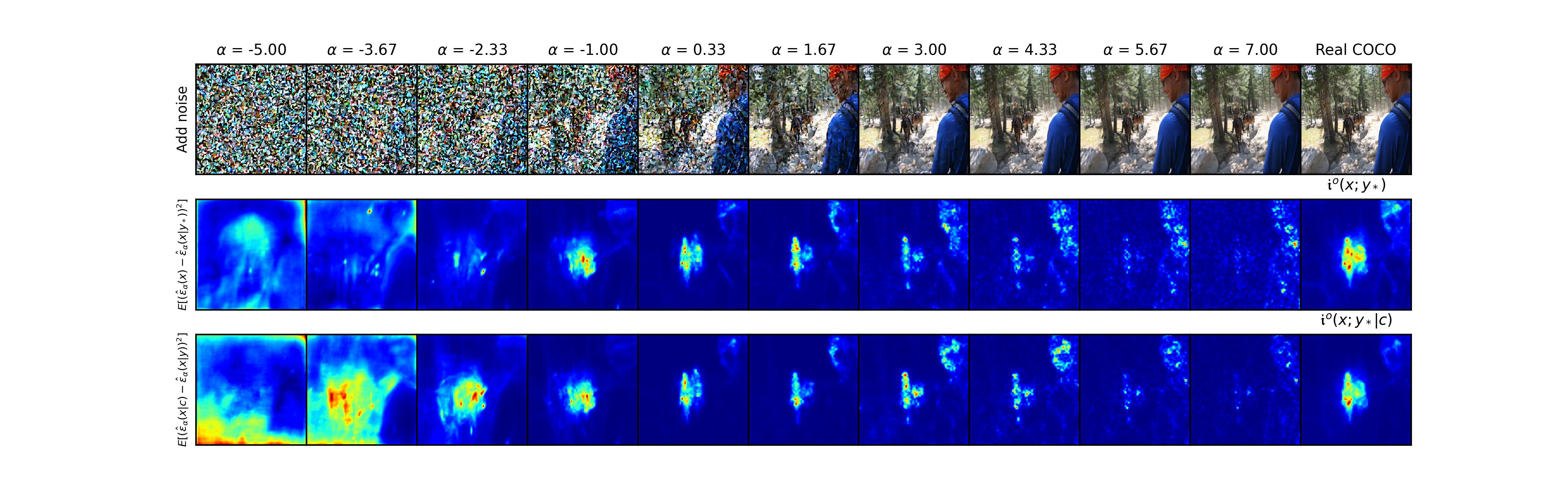}
    \includegraphics[width=0.99\textwidth,trim={3.8cm 9mm 2.8cm 1cm},clip]{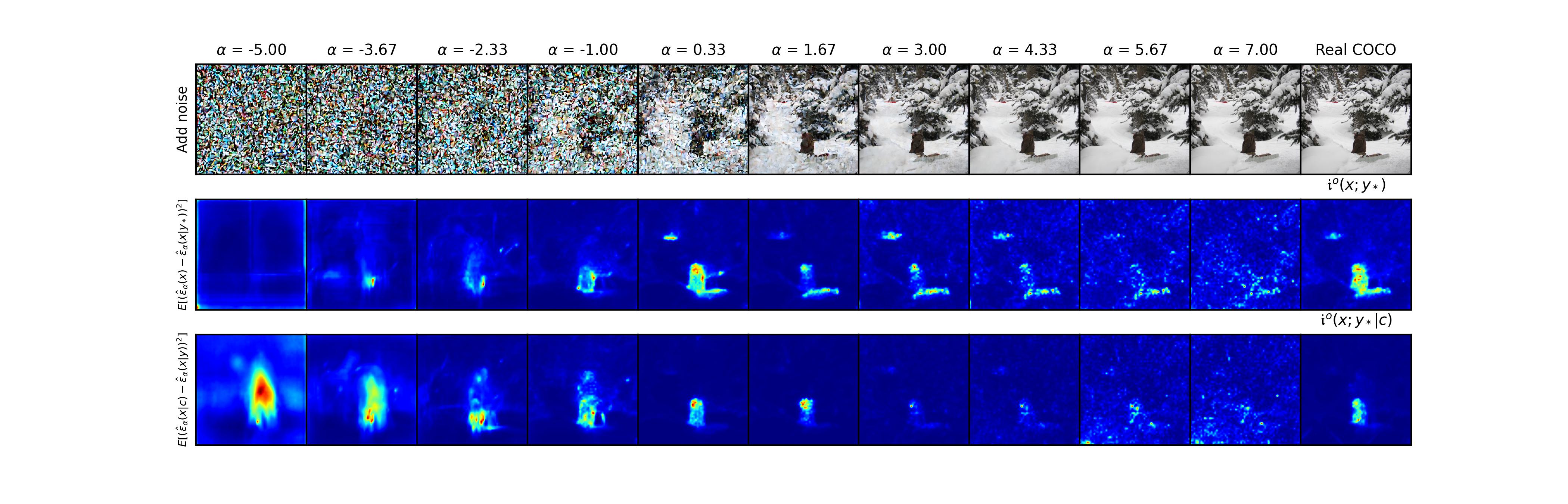}
    \includegraphics[width=0.99\textwidth,trim={3.8cm 9mm 2.8cm 1cm},clip]{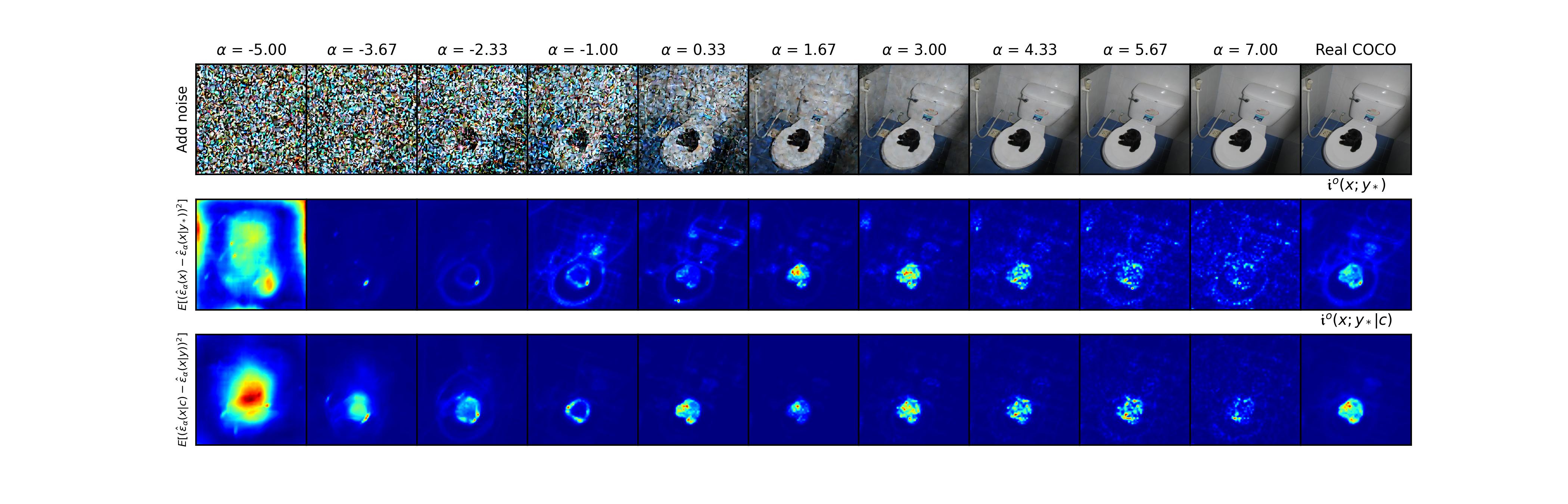}
    \includegraphics[width=0.99\textwidth,trim={3.8cm 9mm 2.8cm 1cm},clip]{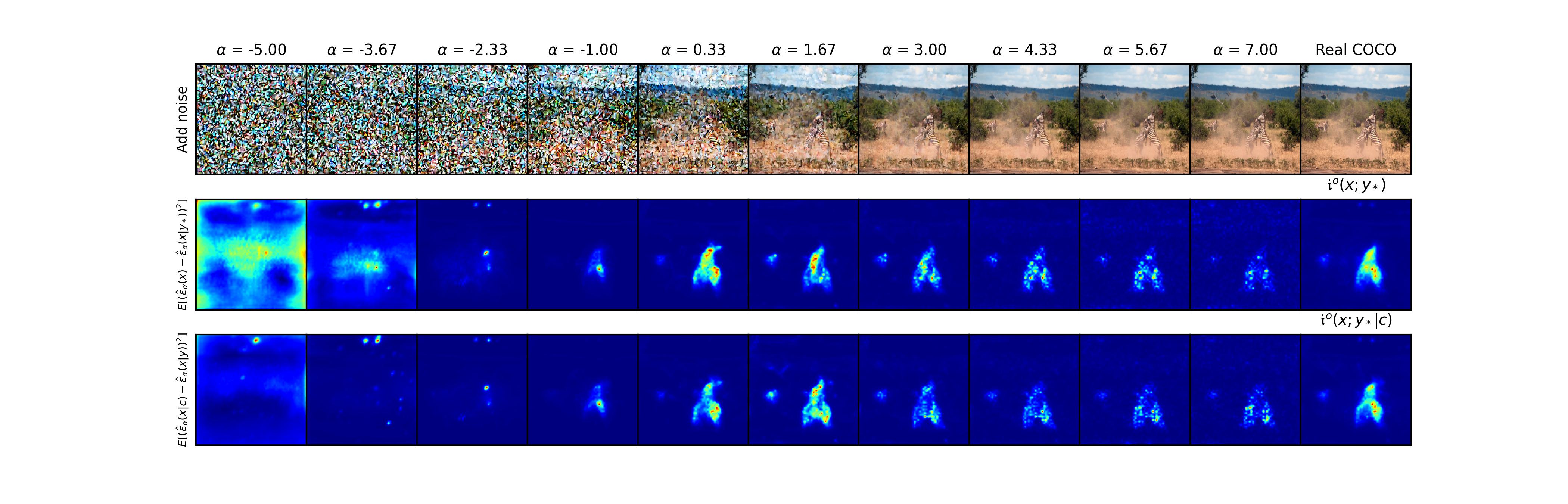}
    \caption{Examples of pixel-level MMSE visualization.}
    \label{fig:mmse-2D-2}
\end{figure}

We analyze the image-level (Fig. \ref{fig:mmses}) and pixel-level (Fig. \ref{fig:mmse-2D-1} and \ref{fig:mmse-2D-2}) MMSE for 10 cases in COCO100-IT. To fully harness the capabilities of ITD \citep{kong22}, we configured the diffusion steps to be 200. We conducted 50 samples under the same $\logsnr$, and the MMSE results are derived from the average denoising of these samples. However, for the purpose of pixel-level visualization, we selected only 20 steps. 

From Fig. \ref{fig:mmses}, it becomes evident that as $\alpha$ varies, the orthogonal approximation exhibits greater stability with fewer zigzag patterns compared to the standard version. Furthermore, the orthogonal method enhances the consistency between MMSE and conditional MMSE, leading to synchronized peaks and similar distributions. The diffusion process reveals that the optimal performance for locating object-related pixels in the image coincides with the appearance of peaks in Fig \ref{fig:mmses}. When the $\logsnr$ is too high, the highlighted pixels gradually become sparse, while excessively low $\logsnr$ values lead to chaos in MMSE.

\subsection{Examples of word location for object nouns} \label{app:2d_mi_cmi_attn_noun}
\begin{figure}[t]
    \centering
    \includegraphics[width=0.49\textwidth,trim={2cm 9mm 1.7cm 15mm},clip]{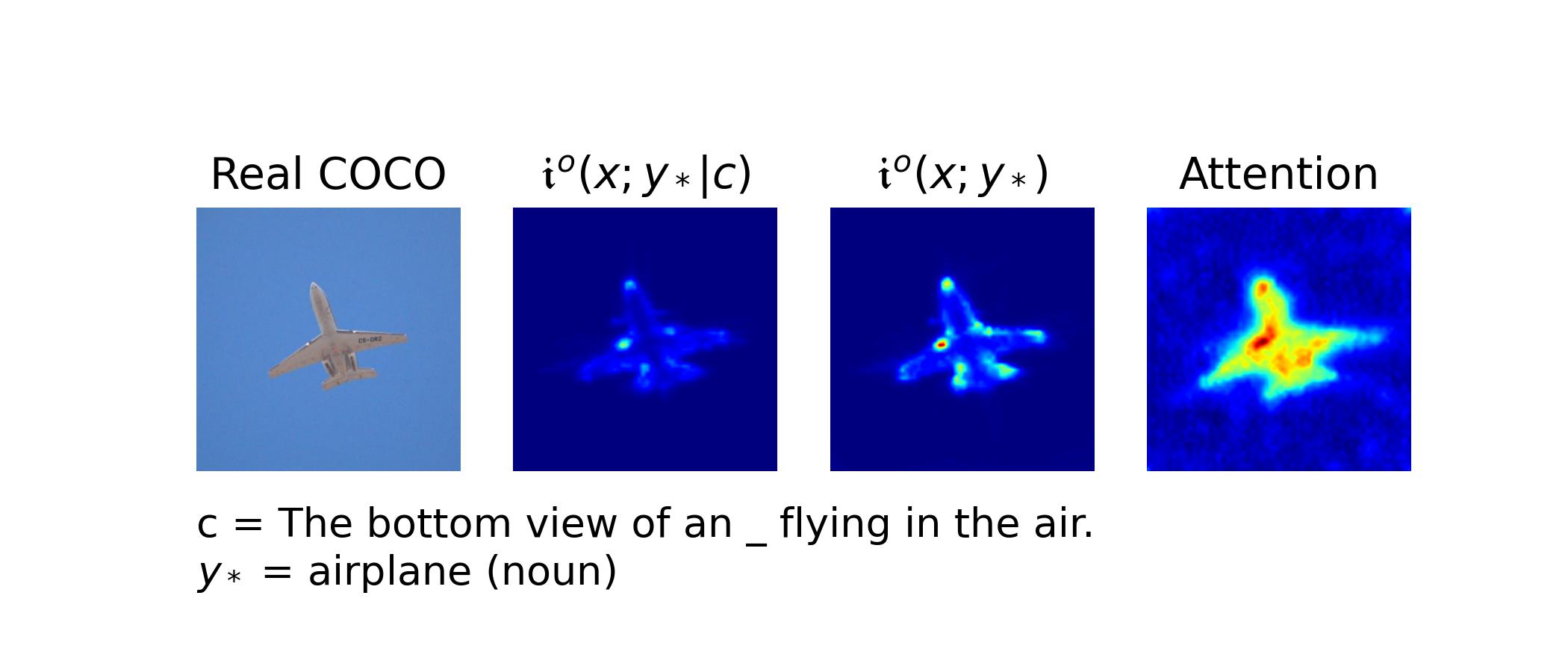} 
    \includegraphics[width=0.49\textwidth,trim={2cm 9mm 1.7cm 15mm},clip]{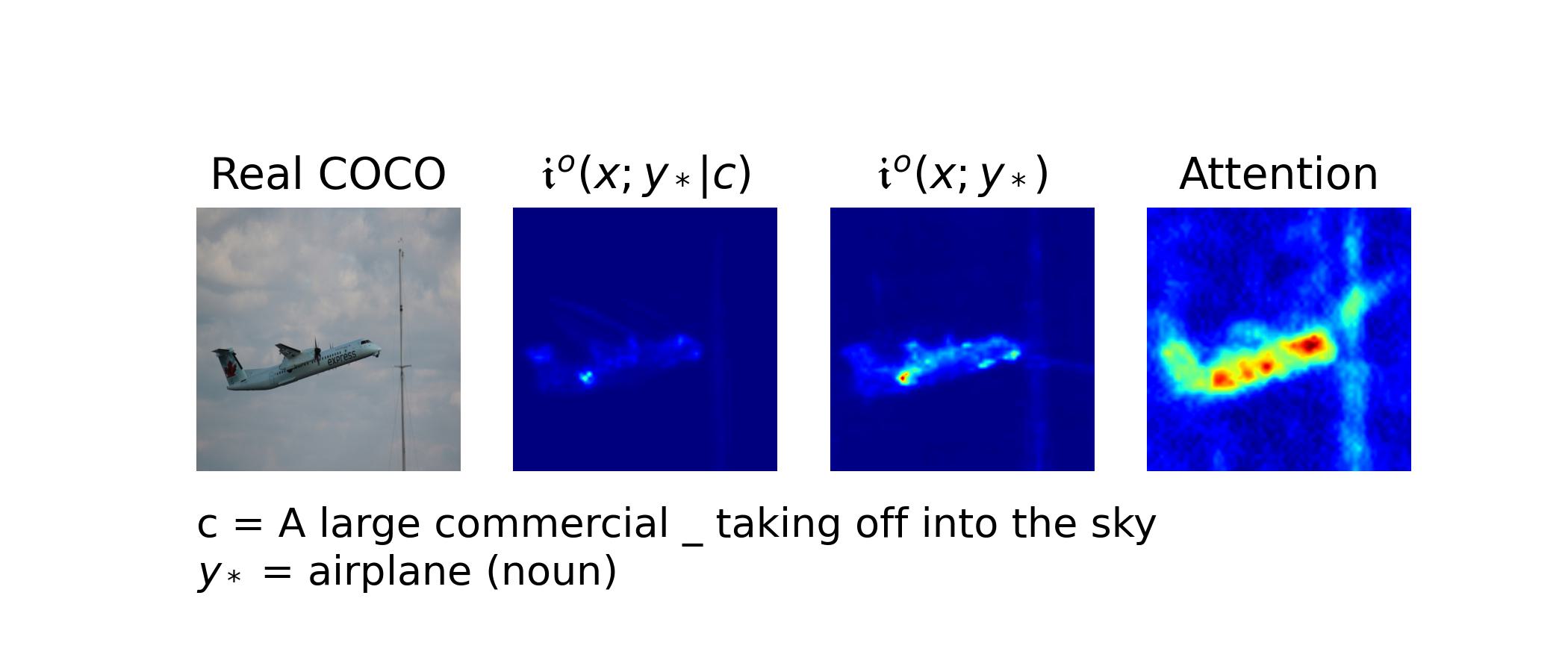} 
    \includegraphics[width=0.49\textwidth,trim={2cm 6mm 1.7cm 15mm},clip]{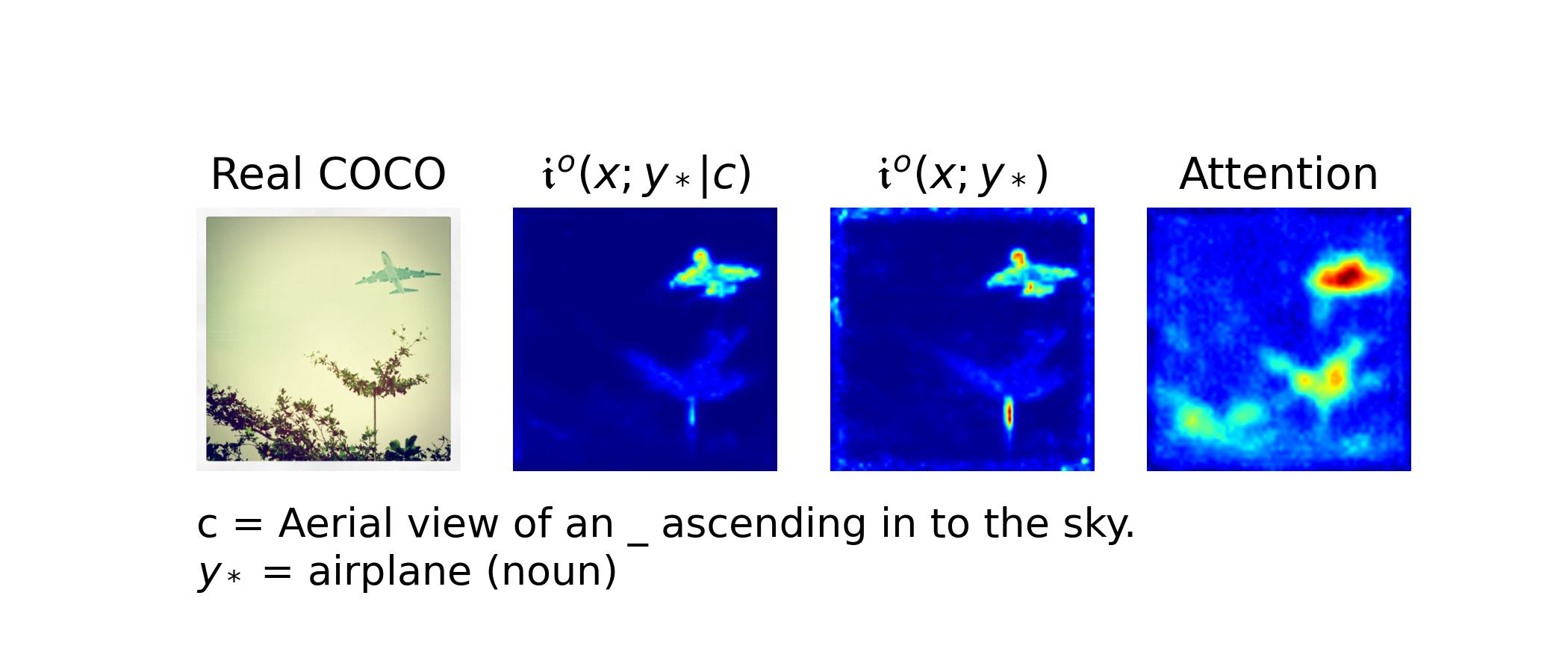} 
    \includegraphics[width=0.49\textwidth,trim={2cm 6mm 1.7cm 15mm},clip]{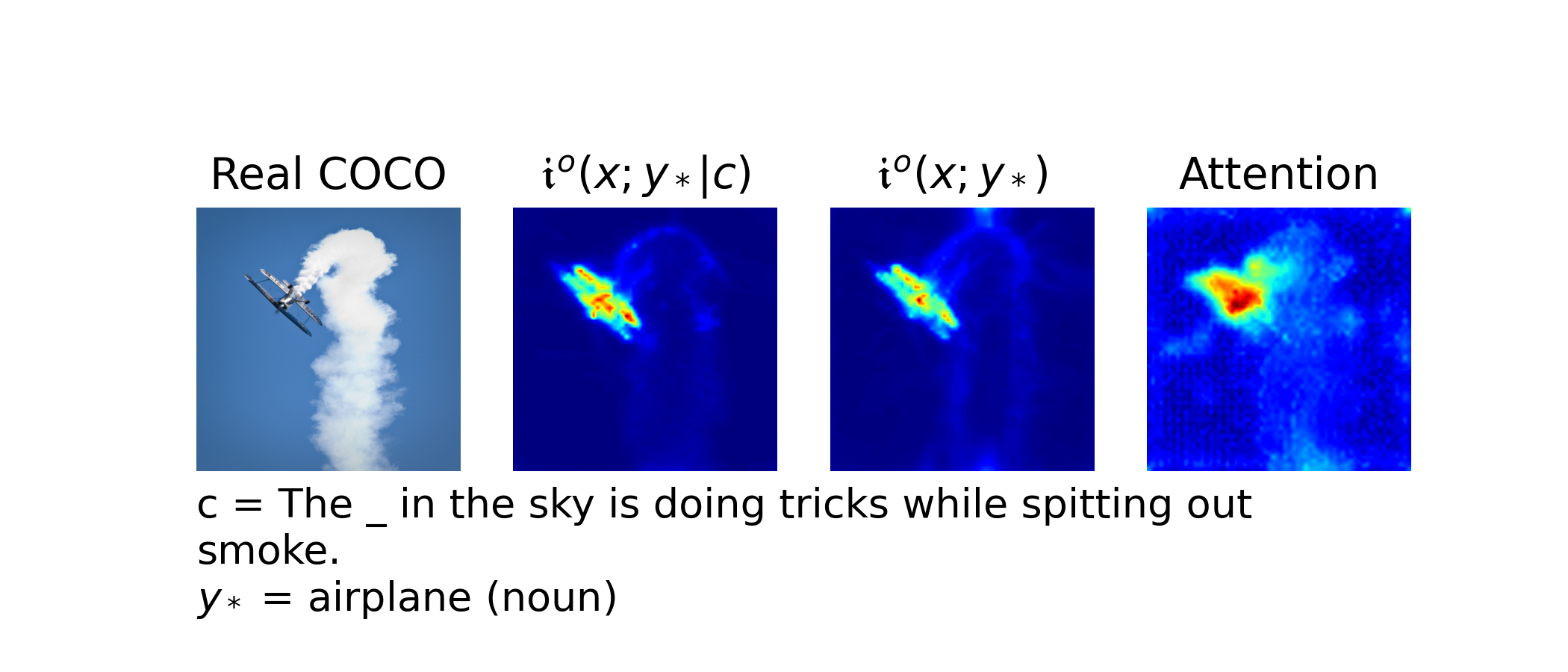} 
    \includegraphics[width=0.49\textwidth,trim={2cm 9mm 1.7cm 15mm},clip]{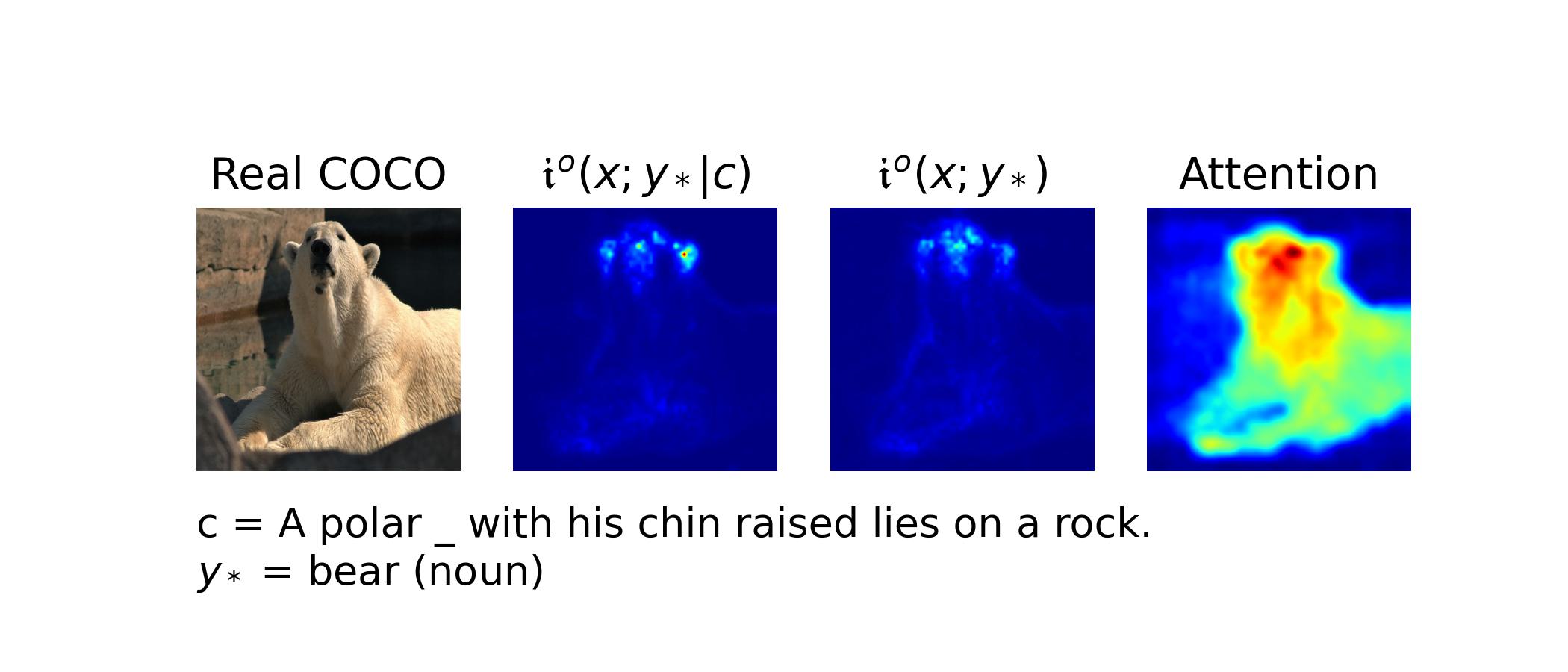} 
    \includegraphics[width=0.49\textwidth,trim={2cm 9mm 1.7cm 15mm},clip]{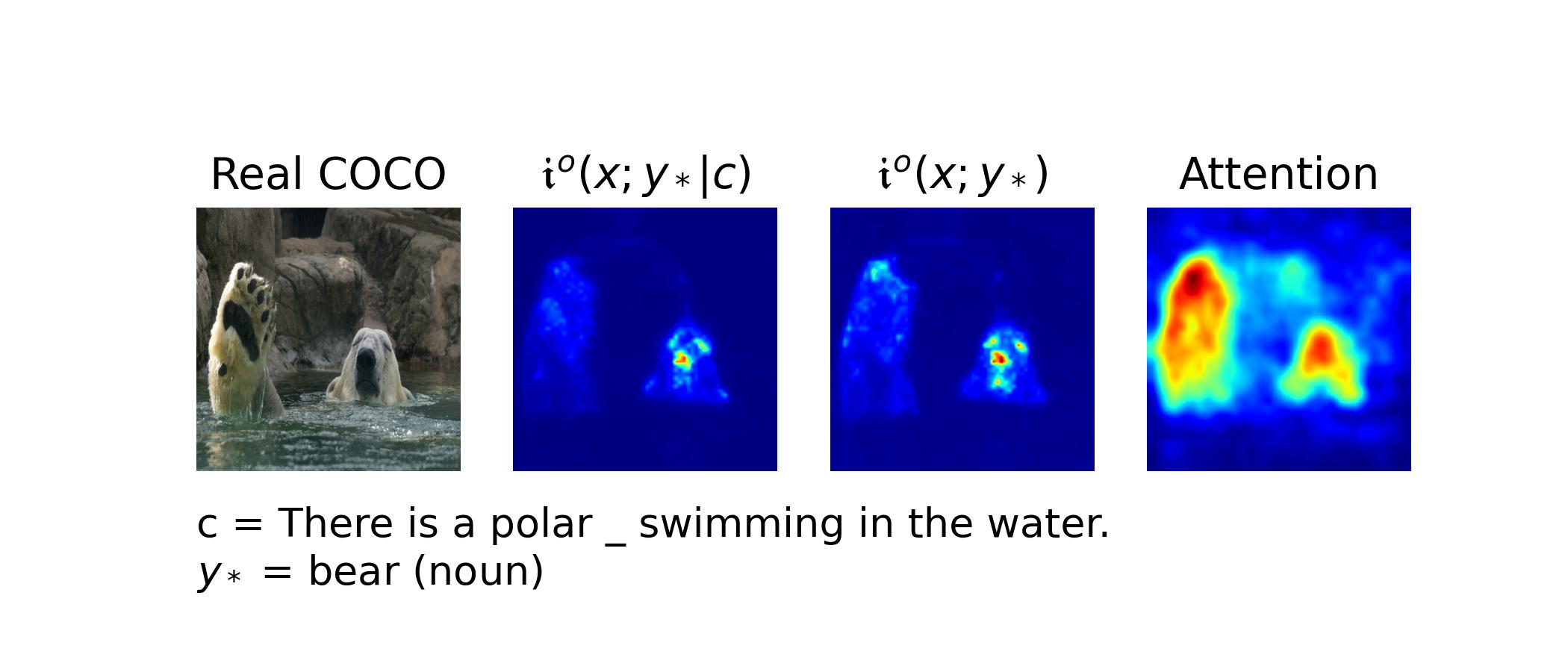} 
    \includegraphics[width=0.49\textwidth,trim={2cm 6mm 1.7cm 15mm},clip]{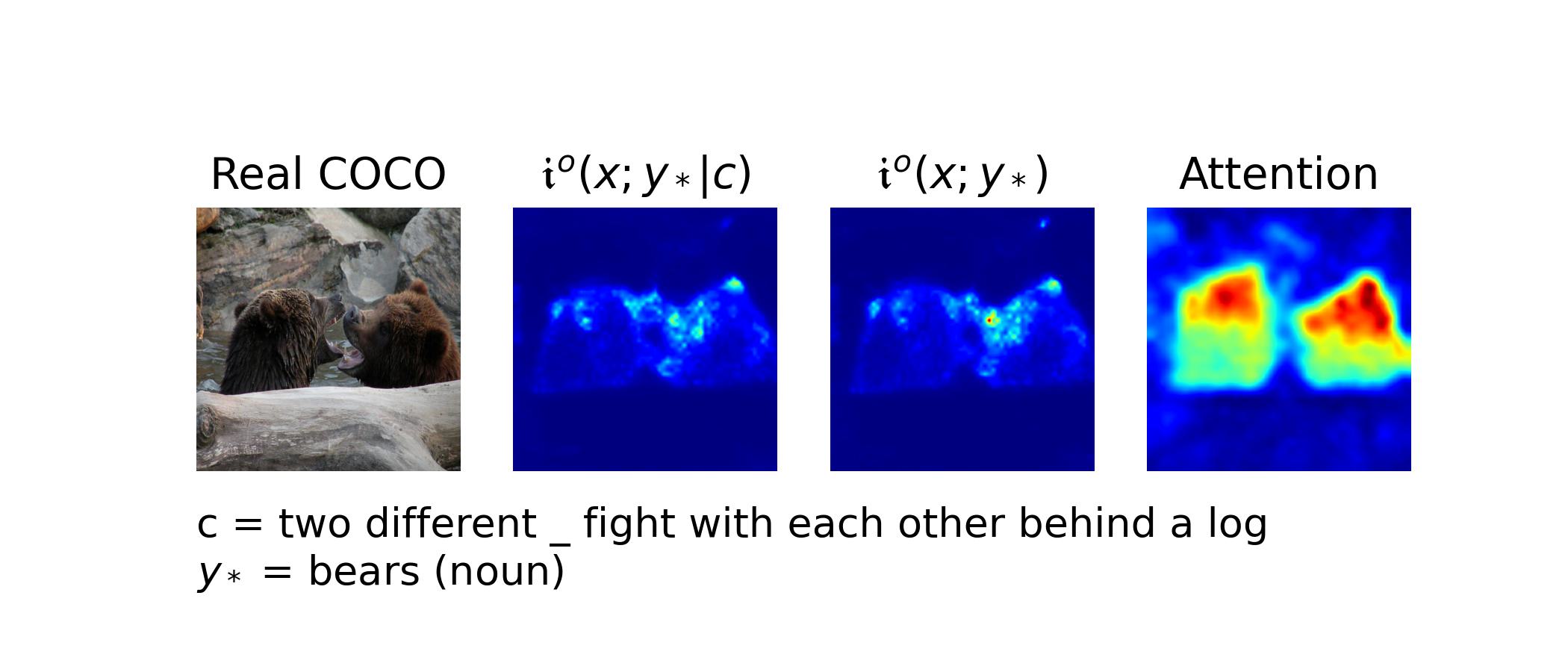} 
    \includegraphics[width=0.49\textwidth,trim={2cm 6mm 1.7cm 15mm},clip]{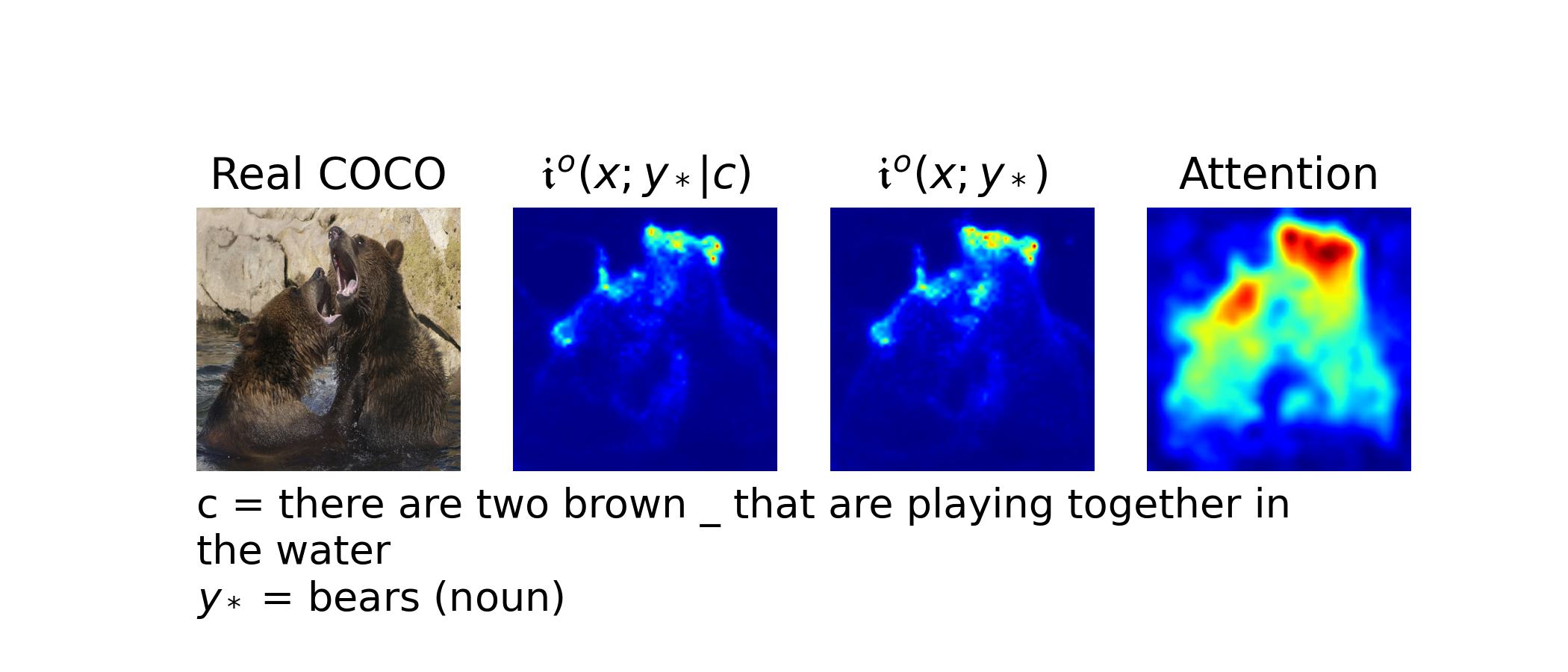} 
    \includegraphics[width=0.49\textwidth,trim={2cm 6mm 1.7cm 15mm},clip]{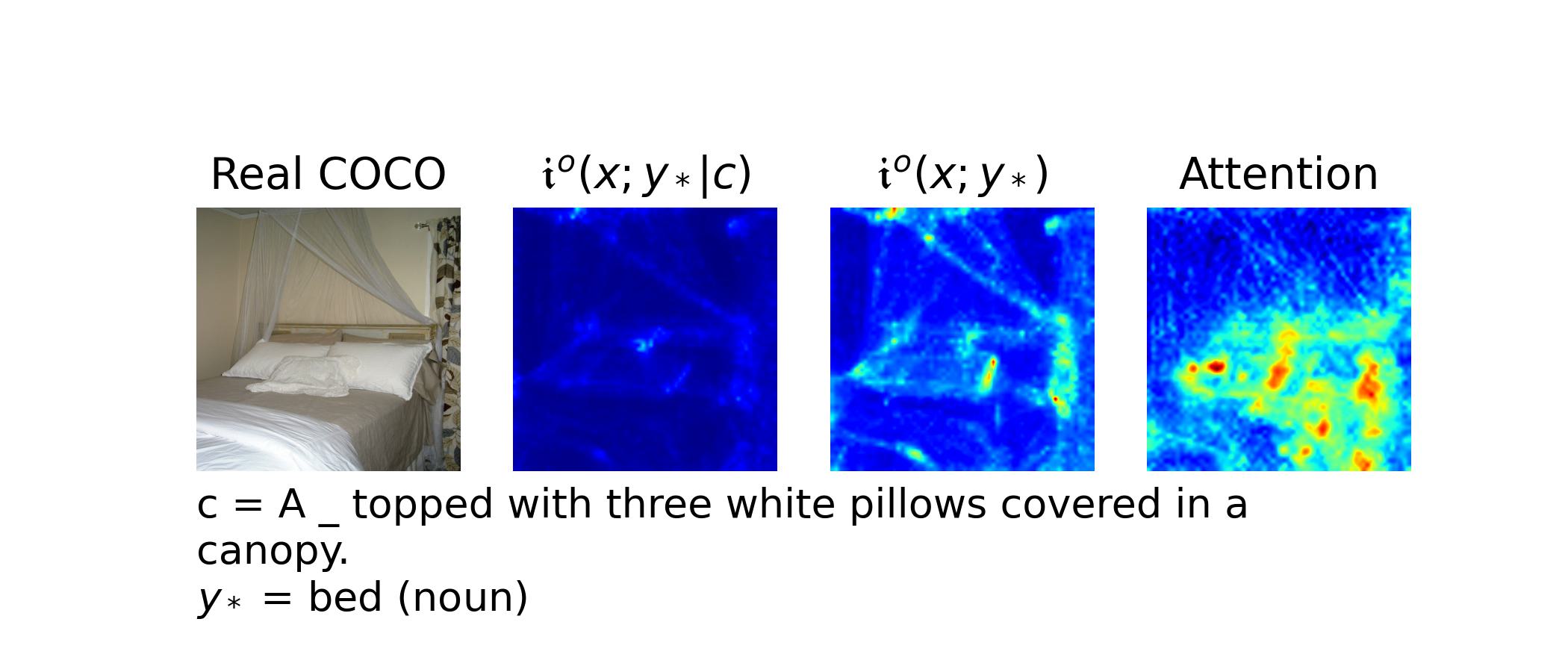} 
    \includegraphics[width=0.49\textwidth,trim={2cm 6mm 1.7cm 15mm},clip]{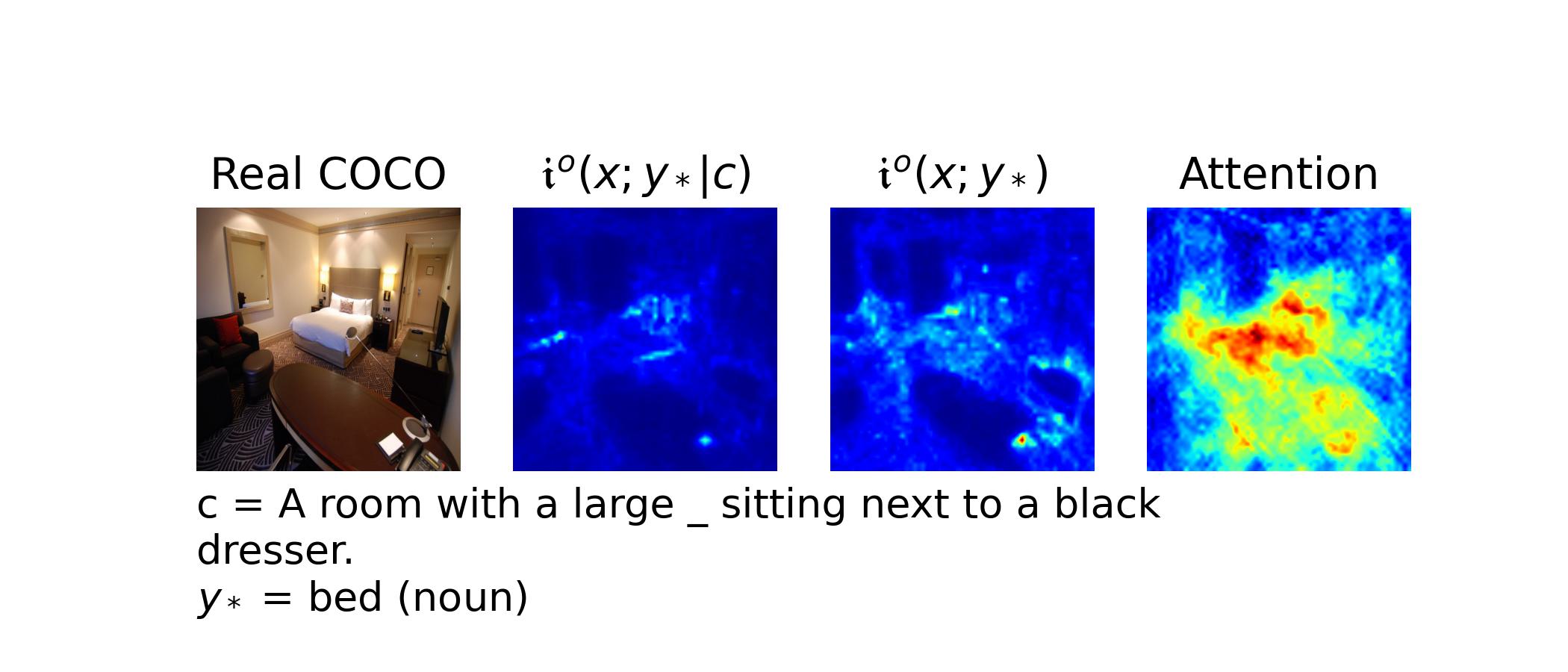} 
    \includegraphics[width=0.49\textwidth,trim={2cm 9mm 1.7cm 15mm},clip]{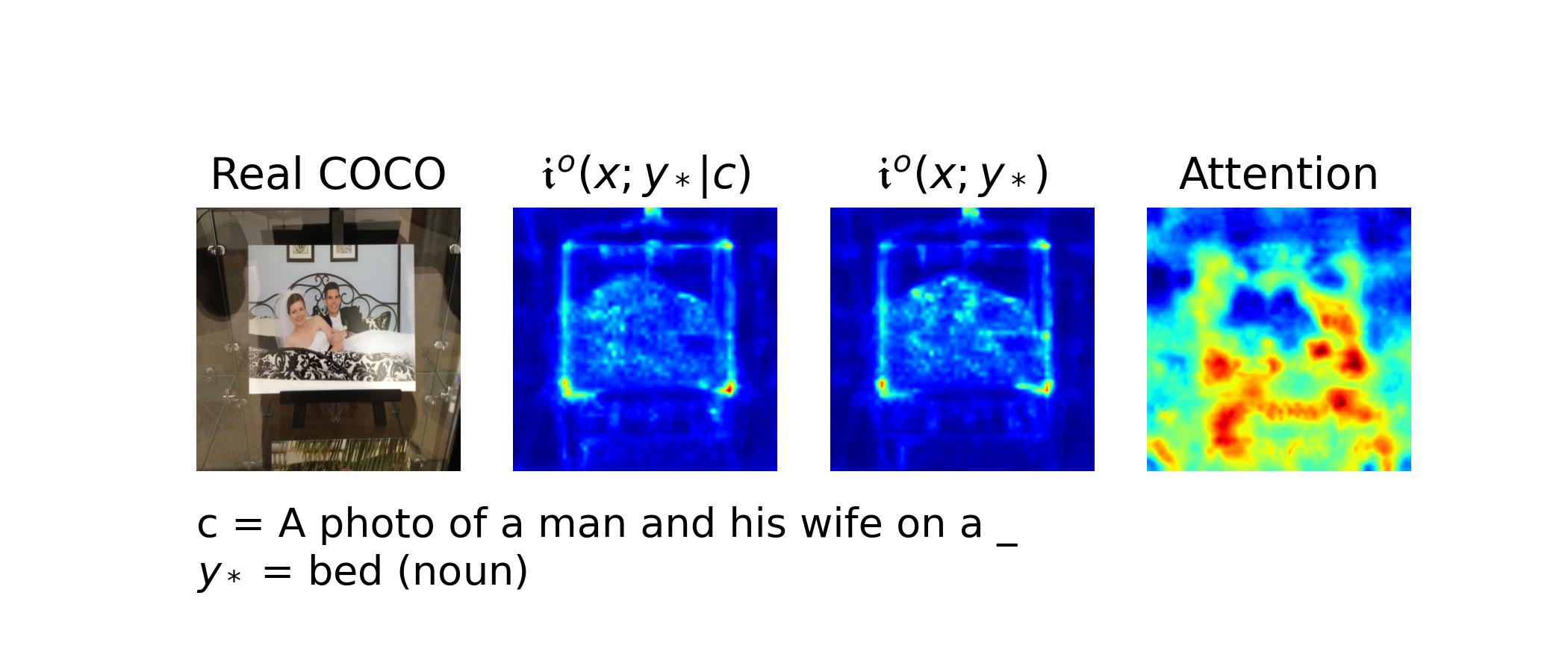} 
    \includegraphics[width=0.49\textwidth,trim={2cm 9mm 1.7cm 15mm},clip]{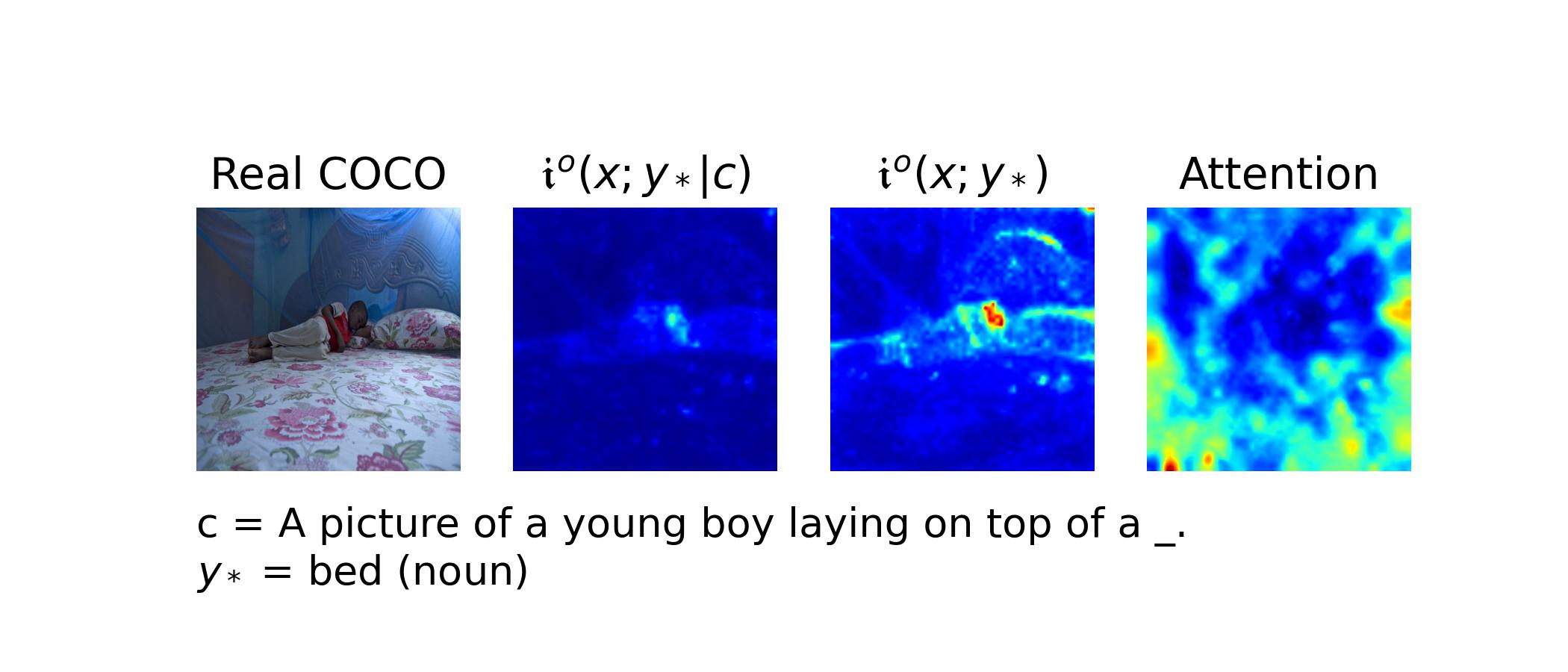} 
    \includegraphics[width=0.49\textwidth,trim={2cm 9mm 1.7cm 15mm},clip]{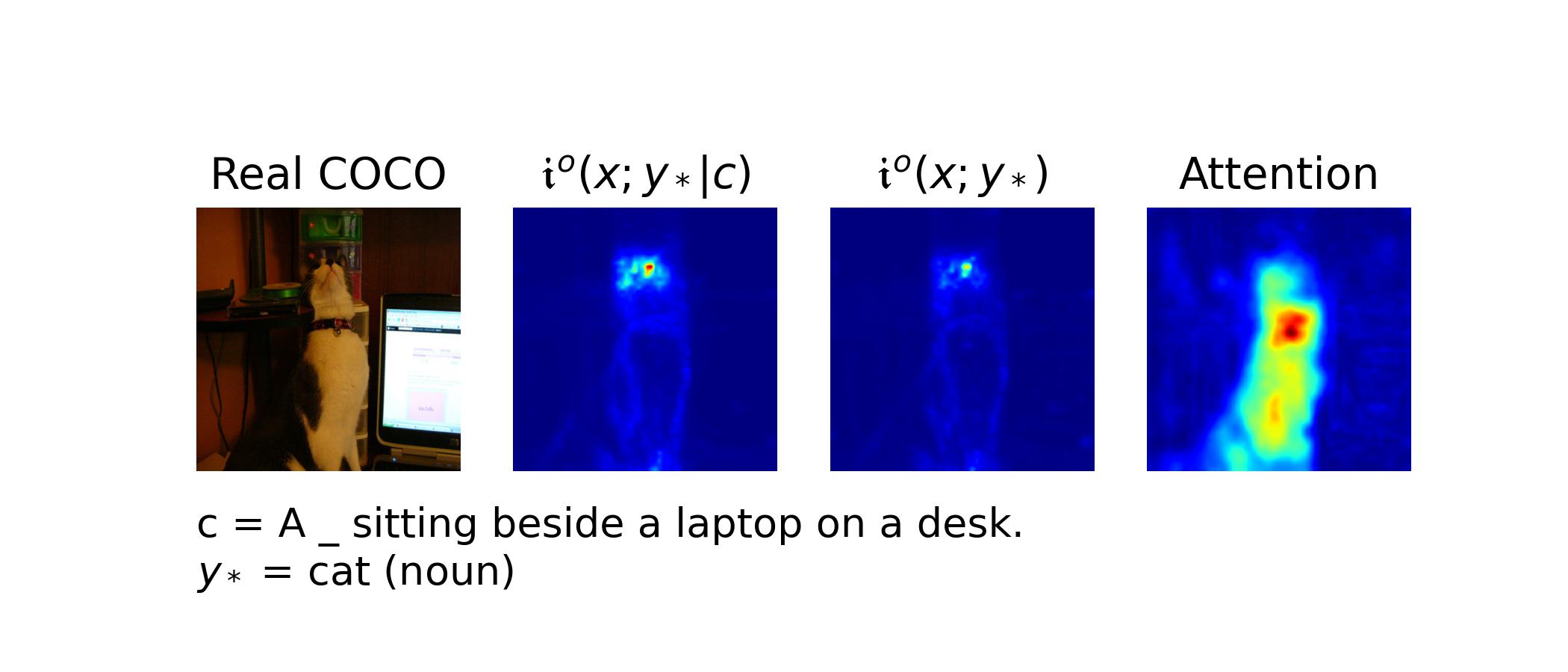} 
    \includegraphics[width=0.49\textwidth,trim={2cm 9mm 1.7cm 15mm},clip]{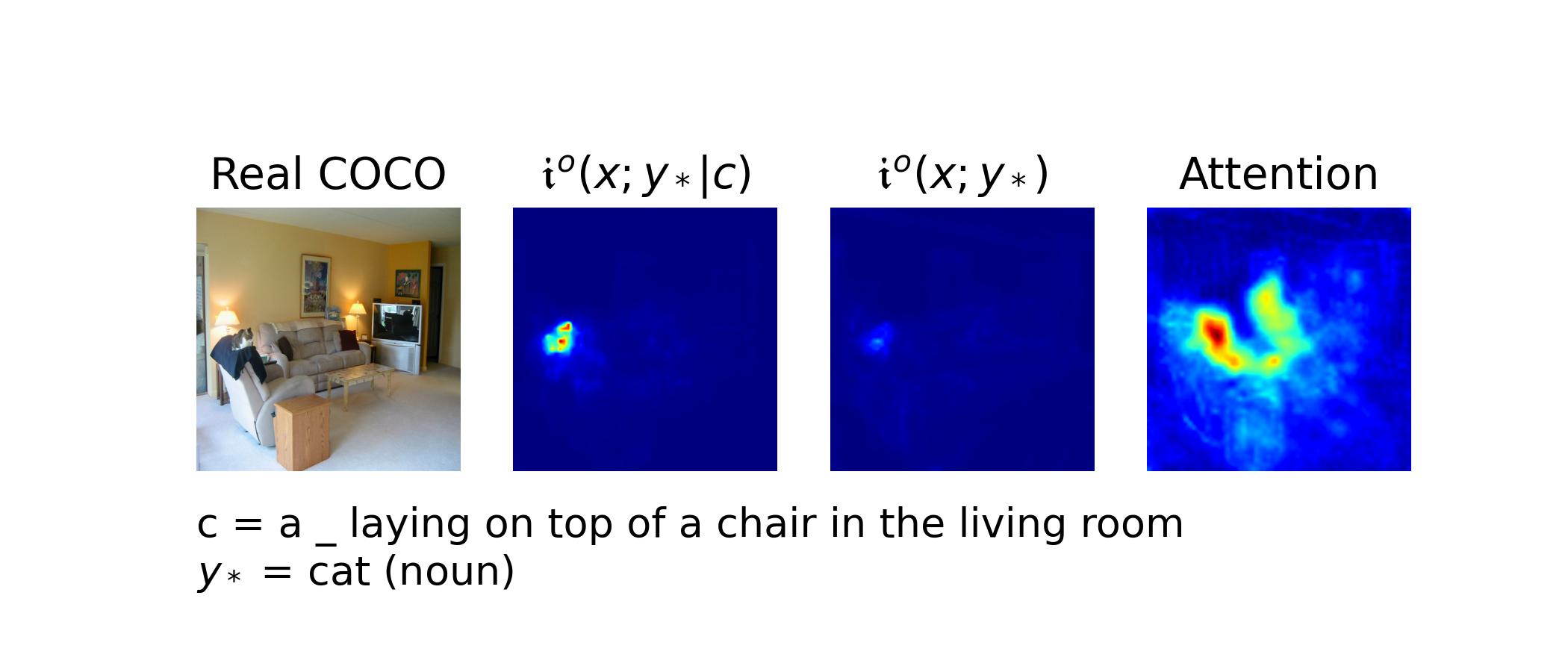} 
    \includegraphics[width=0.49\textwidth,trim={2cm 9mm 1.7cm 15mm},clip]{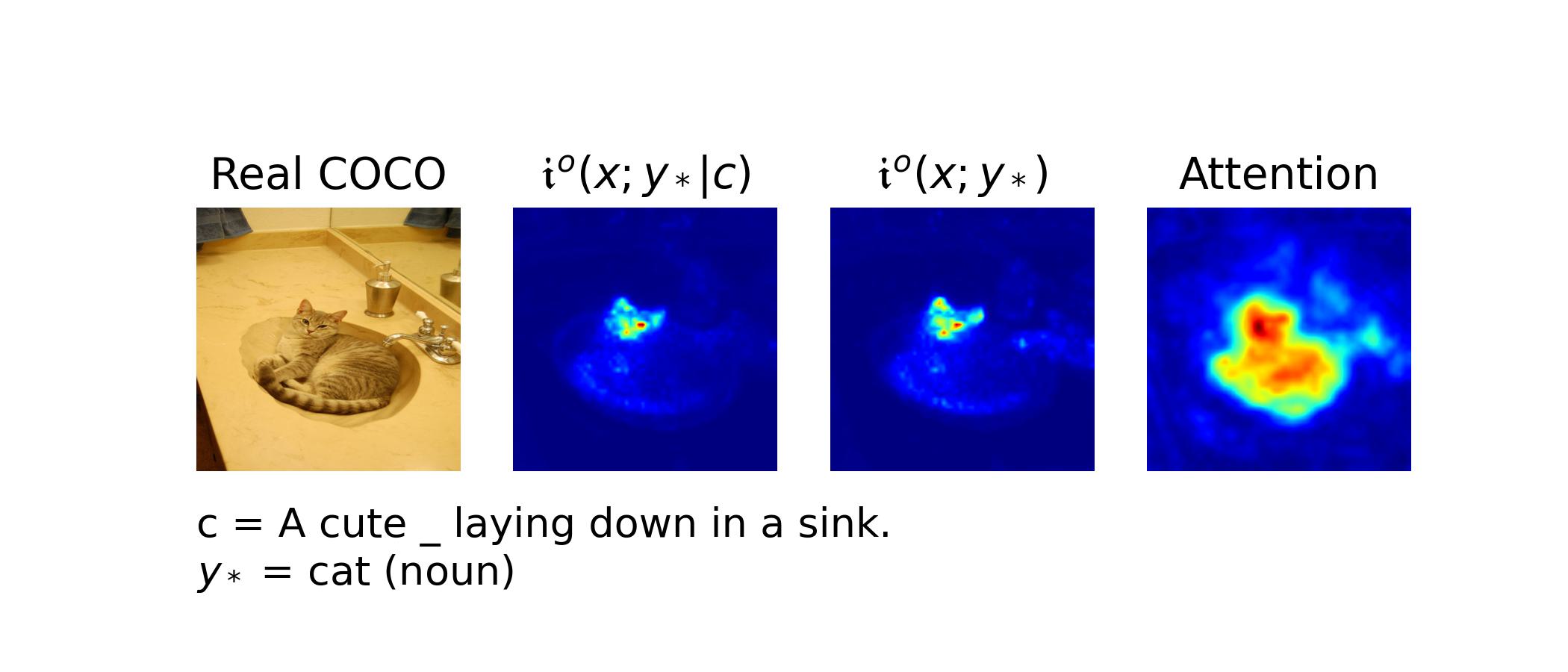} 
    \includegraphics[width=0.49\textwidth,trim={2cm 9mm 1.7cm 15mm},clip]{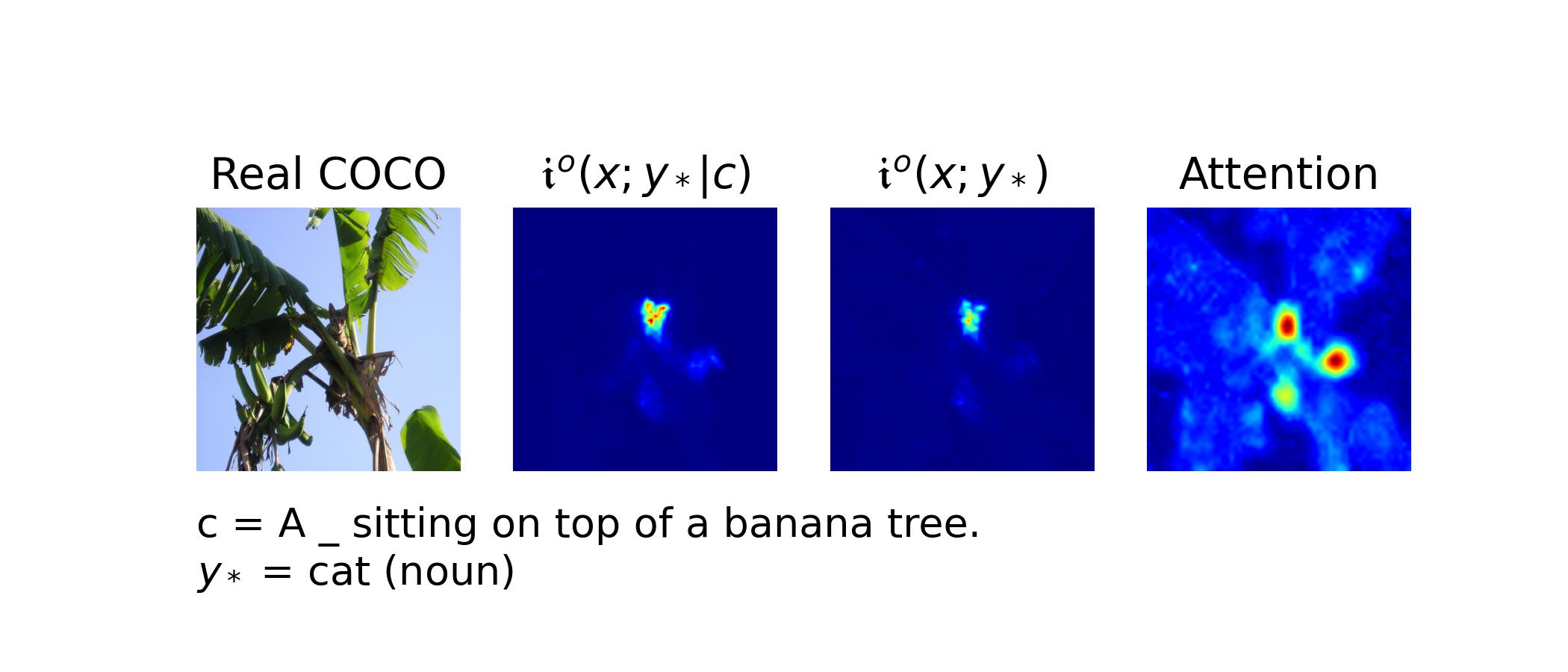} 
    \caption{Examples of localizing noun words in images.}
    \label{fig:2d_mi_cmi_attn_noun_1}
\end{figure}

\begin{figure}[t]
    \centering
    \includegraphics[width=0.49\textwidth,trim={2cm 6mm 1.7cm 15mm},clip]{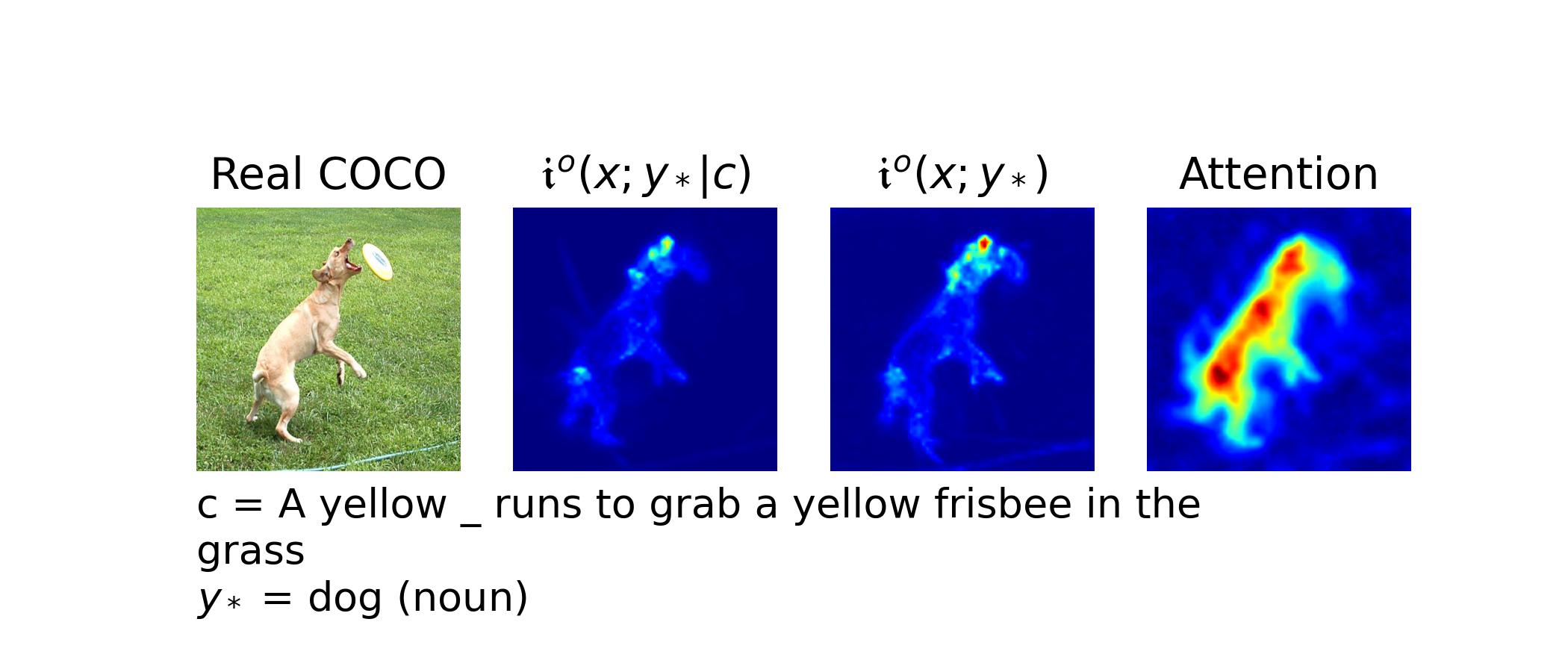} 
    \includegraphics[width=0.49\textwidth,trim={2cm 6mm 1.7cm 15mm},clip]{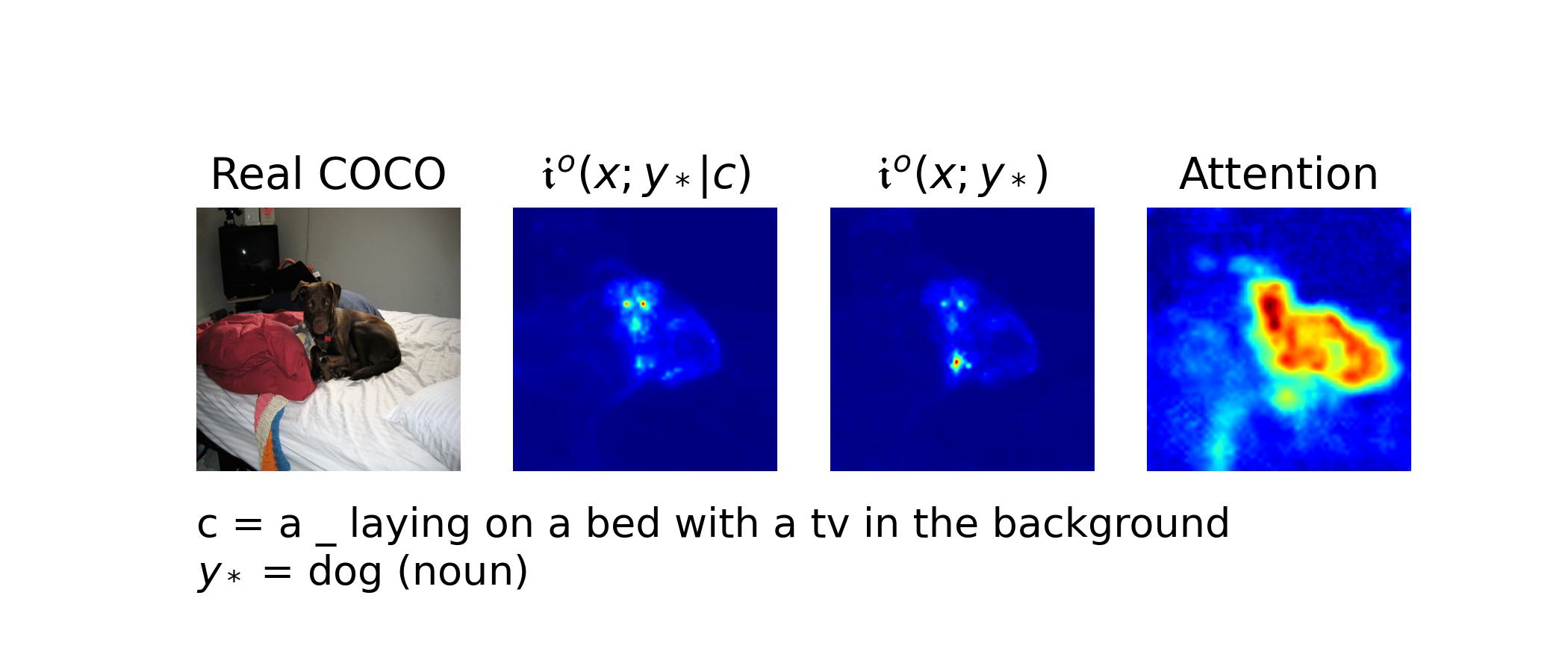} 
    \includegraphics[width=0.49\textwidth,trim={2cm 6mm 1.7cm 15mm},clip]{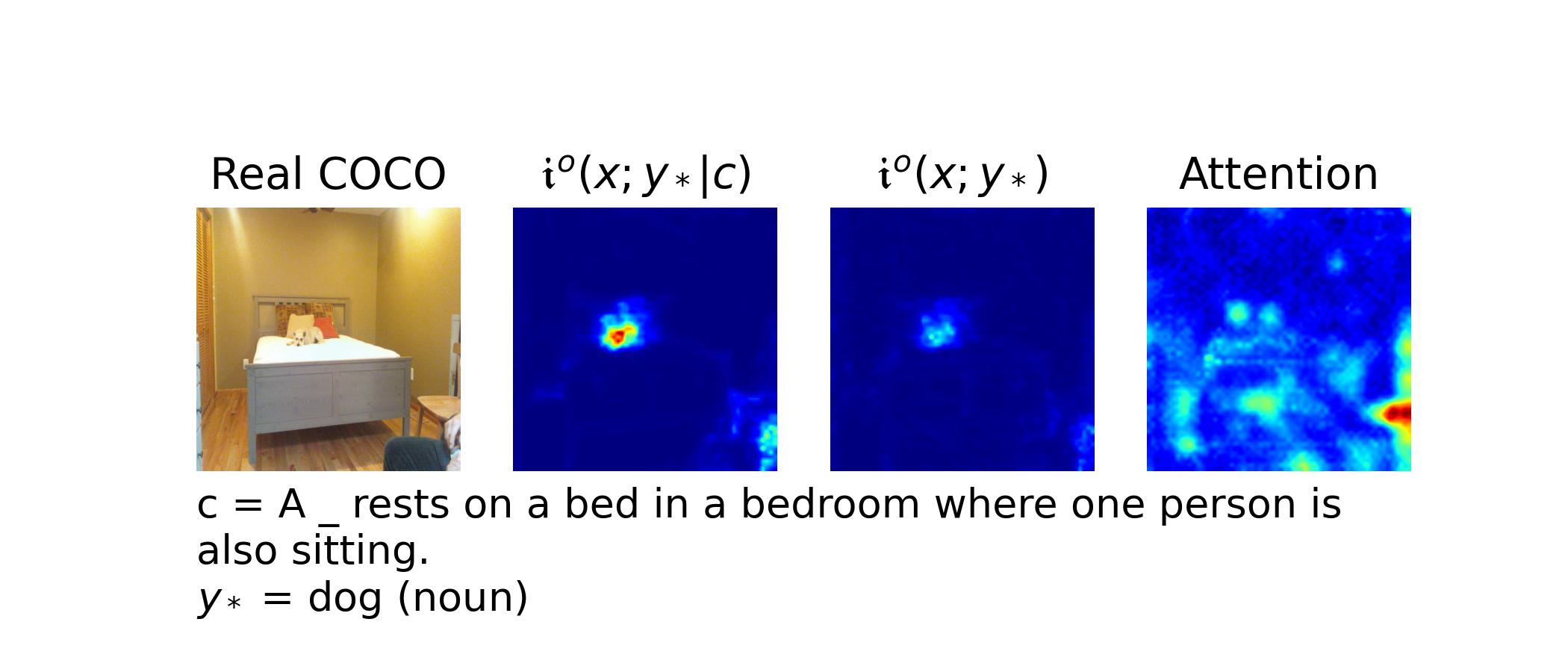} 
    \includegraphics[width=0.49\textwidth,trim={2cm 6mm 1.7cm 15mm},clip]{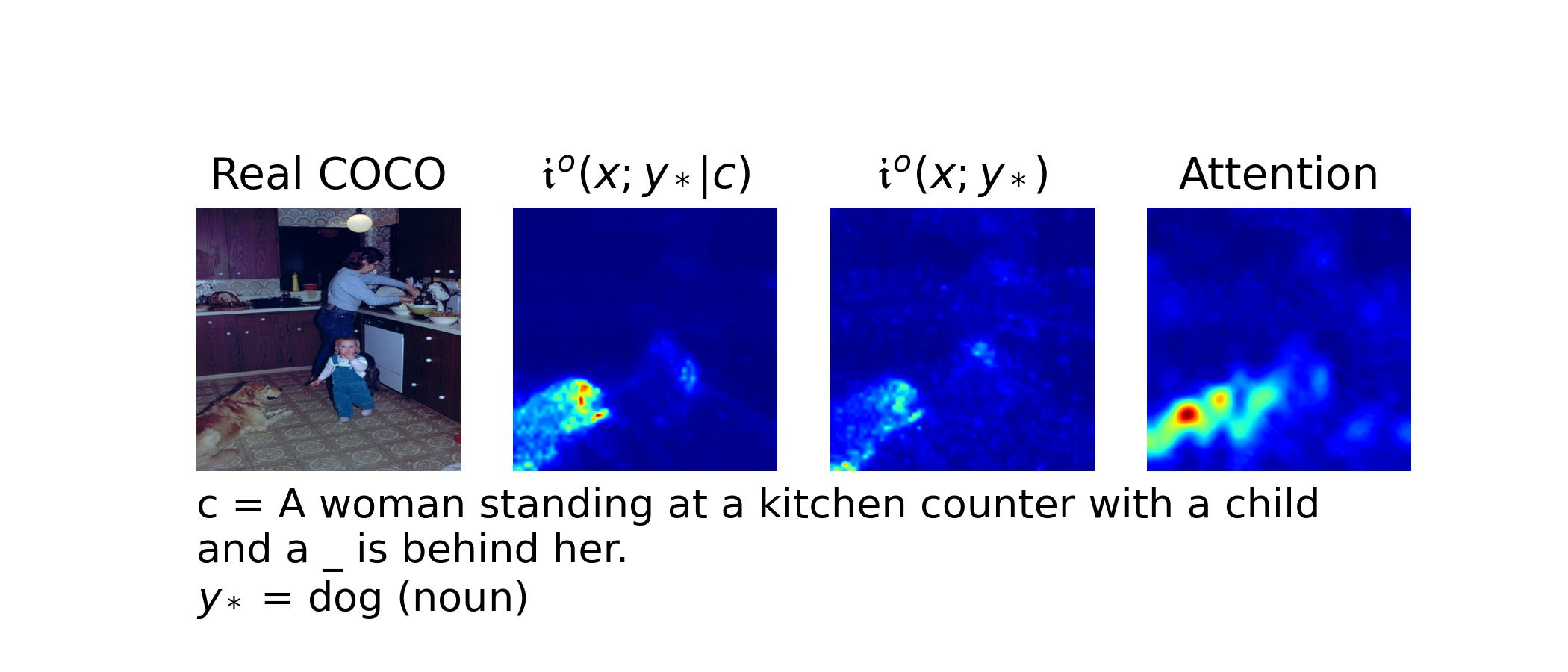} 
    \includegraphics[width=0.49\textwidth,trim={2cm 6mm 1.7cm 15mm},clip]{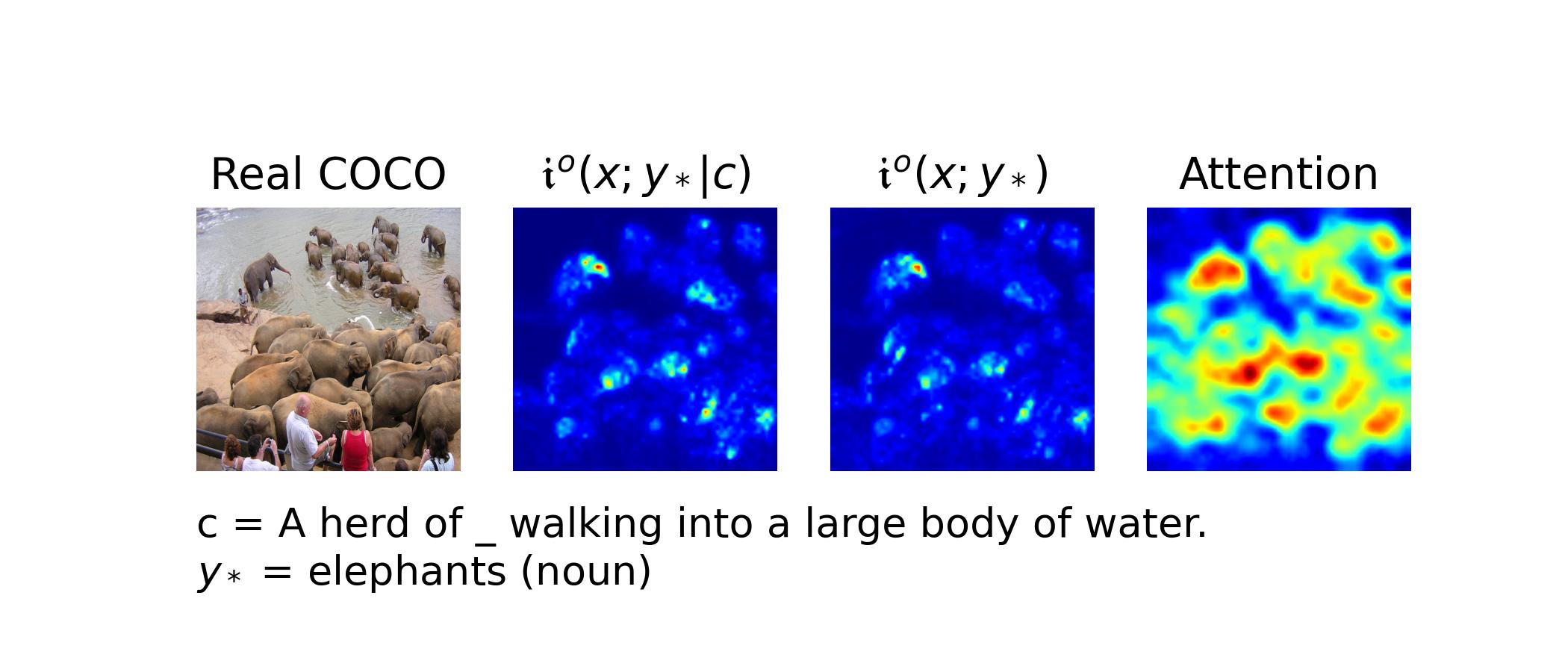} 
    \includegraphics[width=0.49\textwidth,trim={2cm 6mm 1.7cm 15mm},clip]{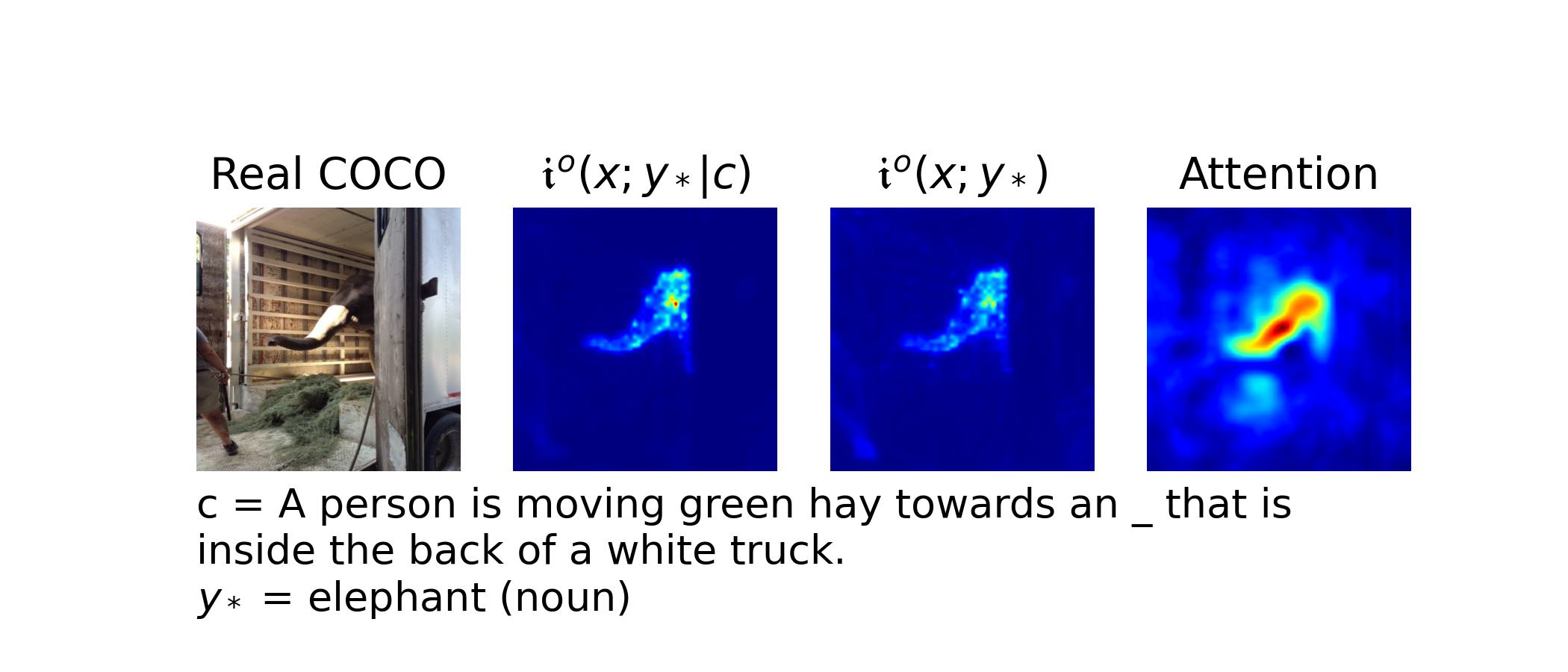} 
    \includegraphics[width=0.49\textwidth,trim={2cm 9mm 1.7cm 15mm},clip]{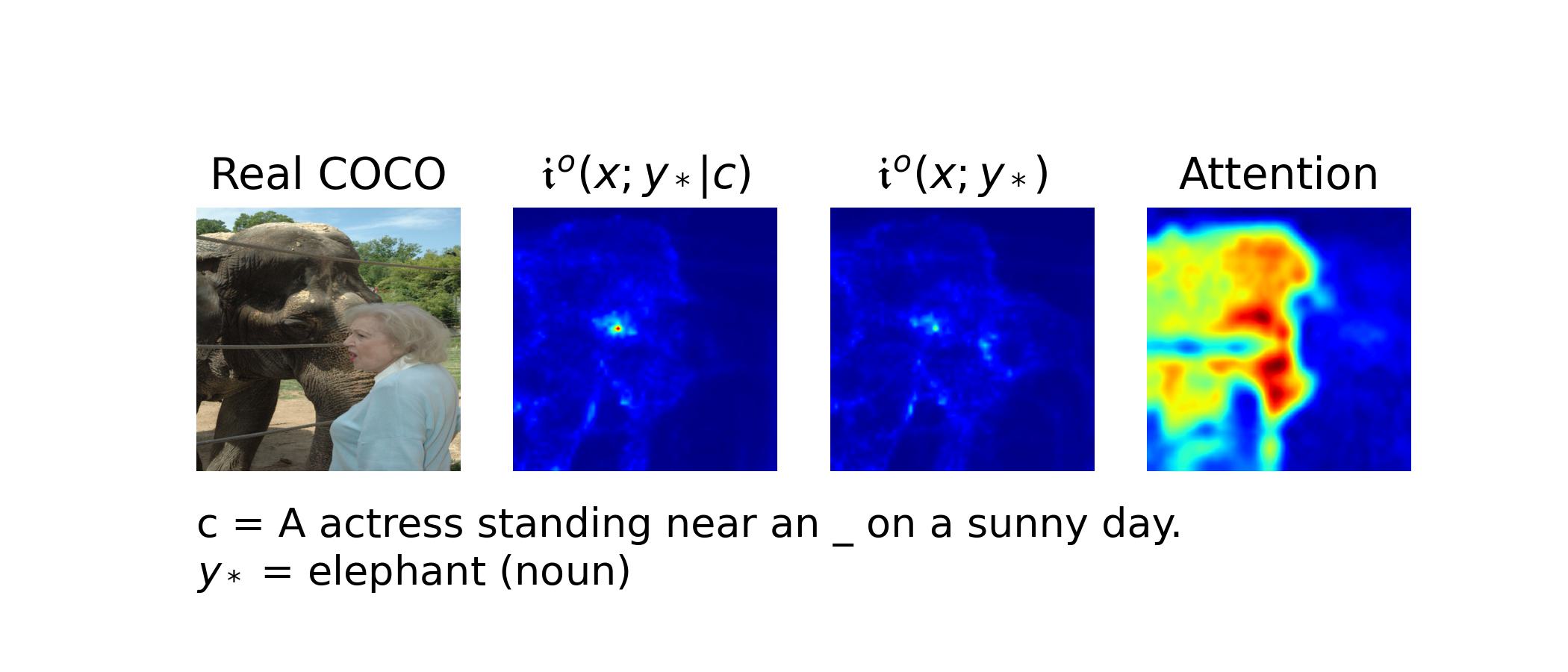} 
    \includegraphics[width=0.49\textwidth,trim={2cm 9mm 1.7cm 15mm},clip]{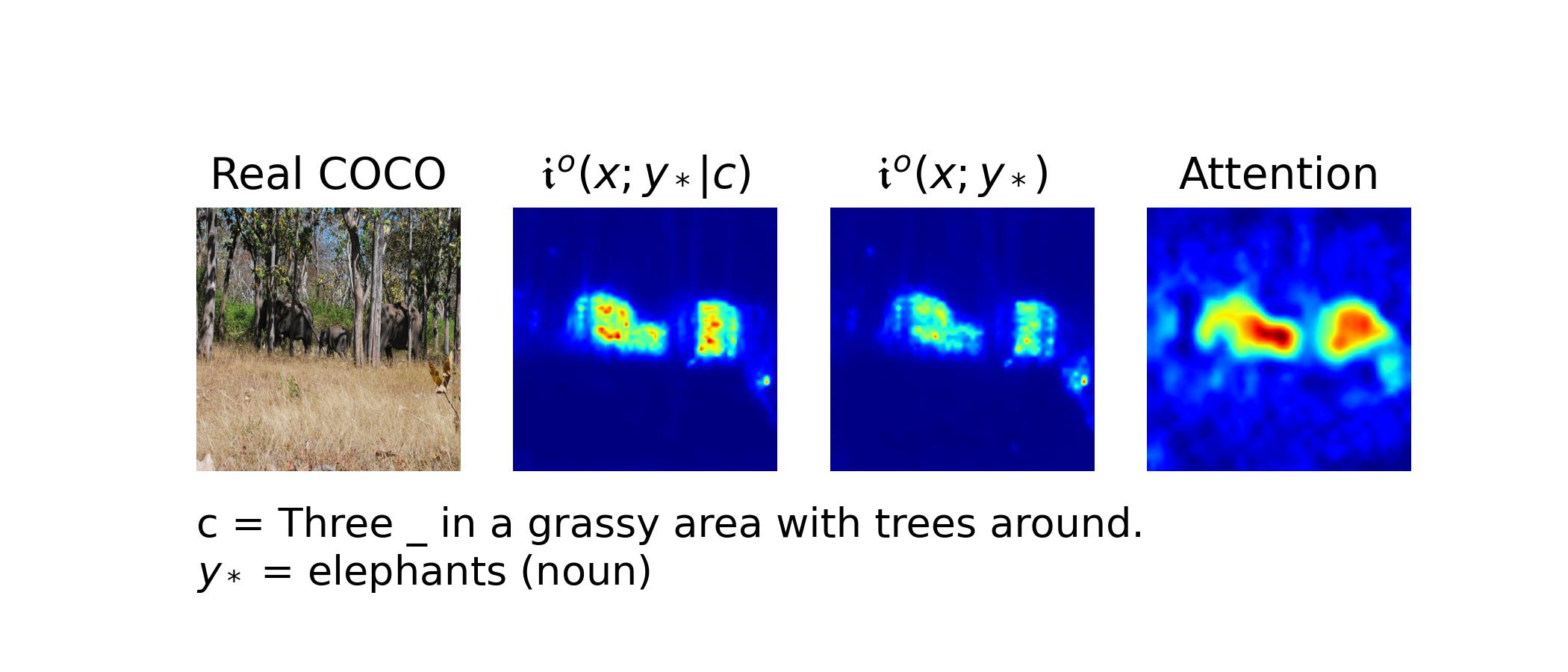} 
    \includegraphics[width=0.49\textwidth,trim={2cm 6mm 1.7cm 15mm},clip]{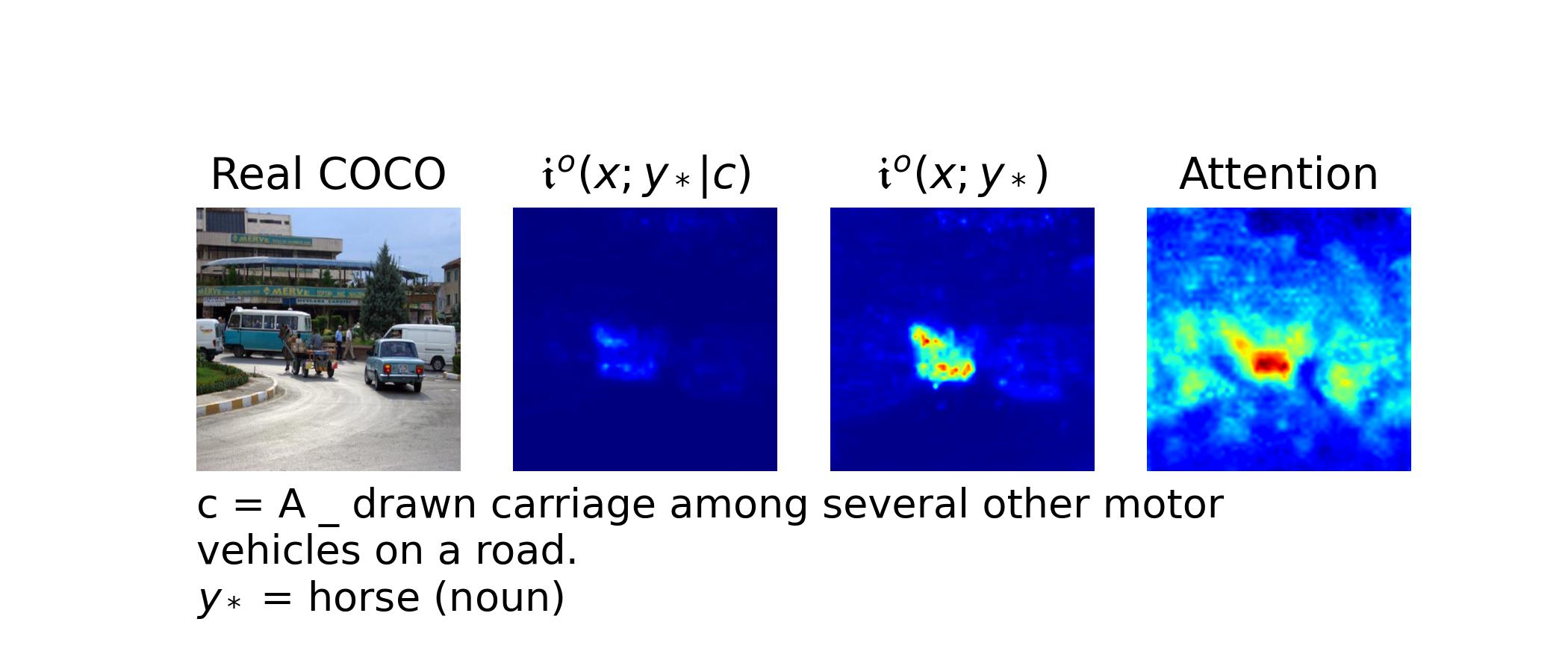} 
    \includegraphics[width=0.49\textwidth,trim={2cm 6mm 1.7cm 15mm},clip]{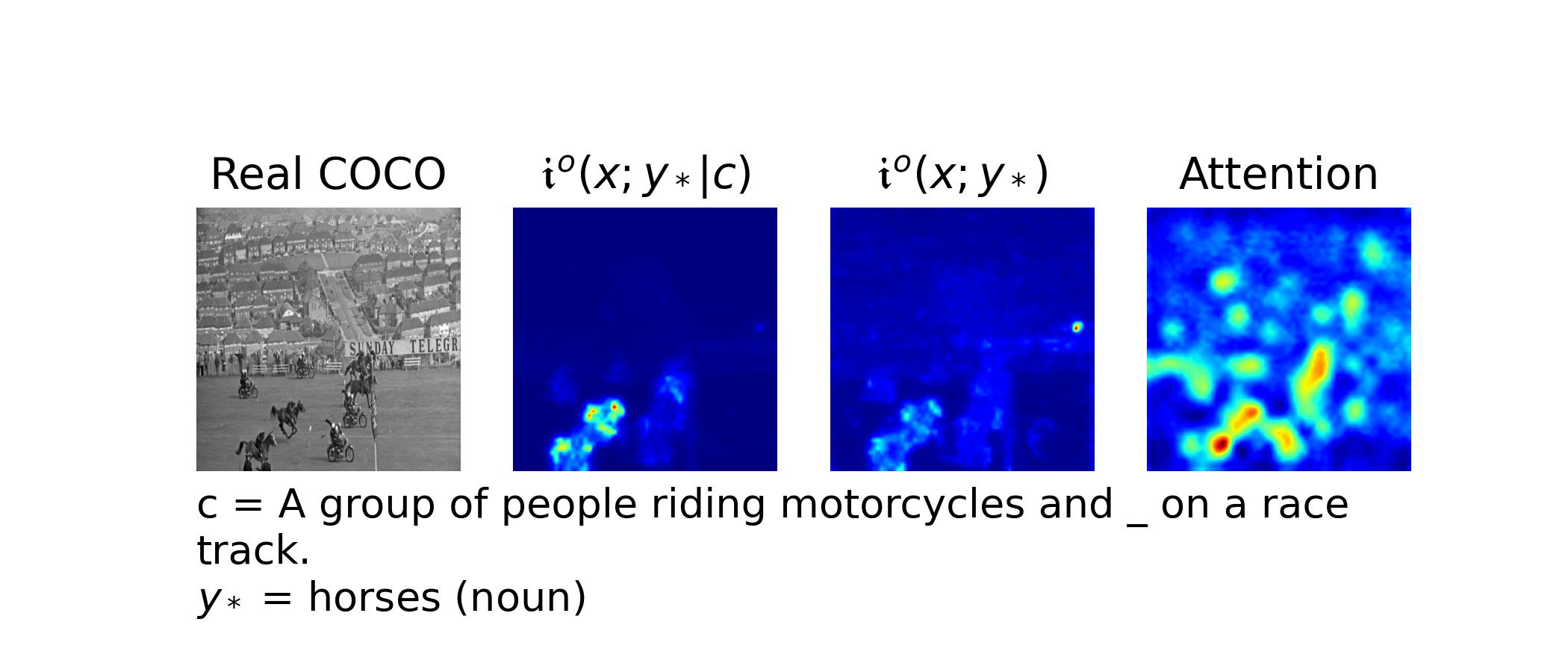} 
    \includegraphics[width=0.49\textwidth,trim={2cm 9mm 1.7cm 15mm},clip]{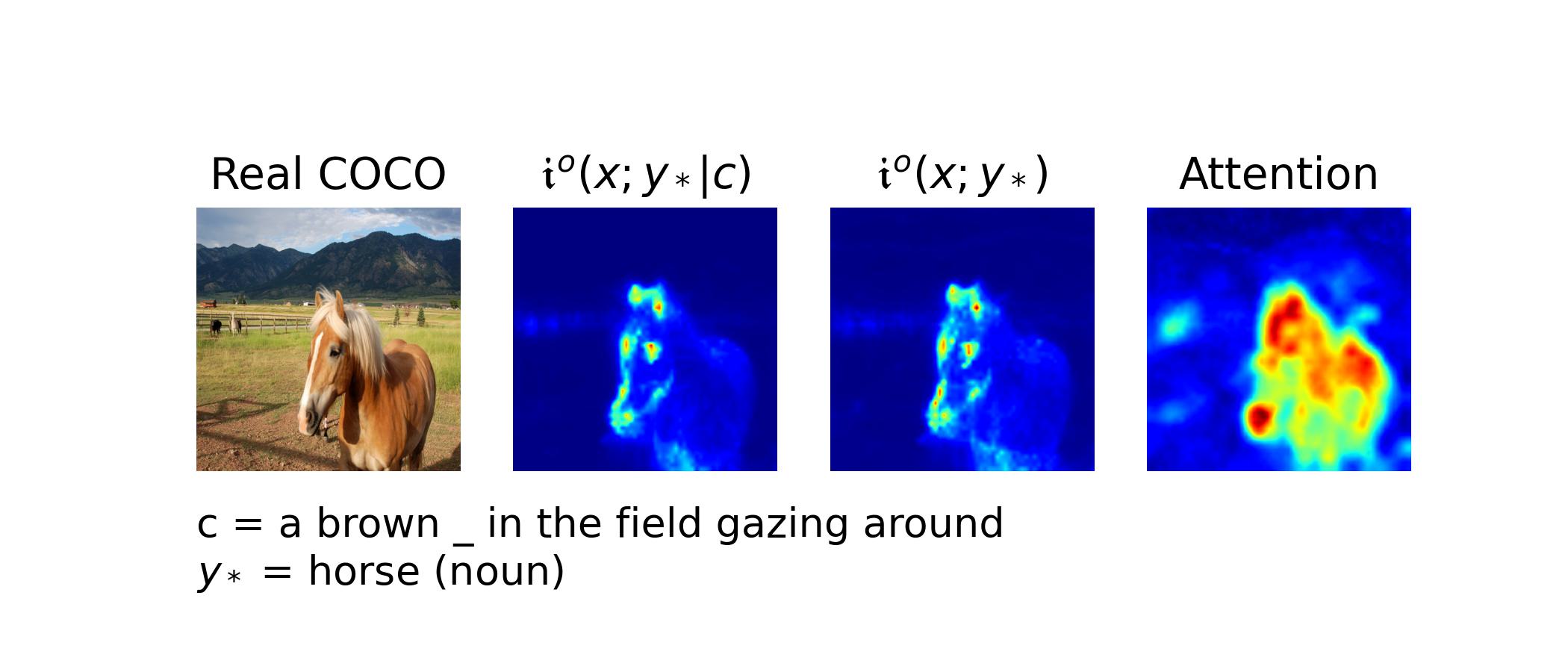} 
    \includegraphics[width=0.49\textwidth,trim={2cm 9mm 1.7cm 15mm},clip]{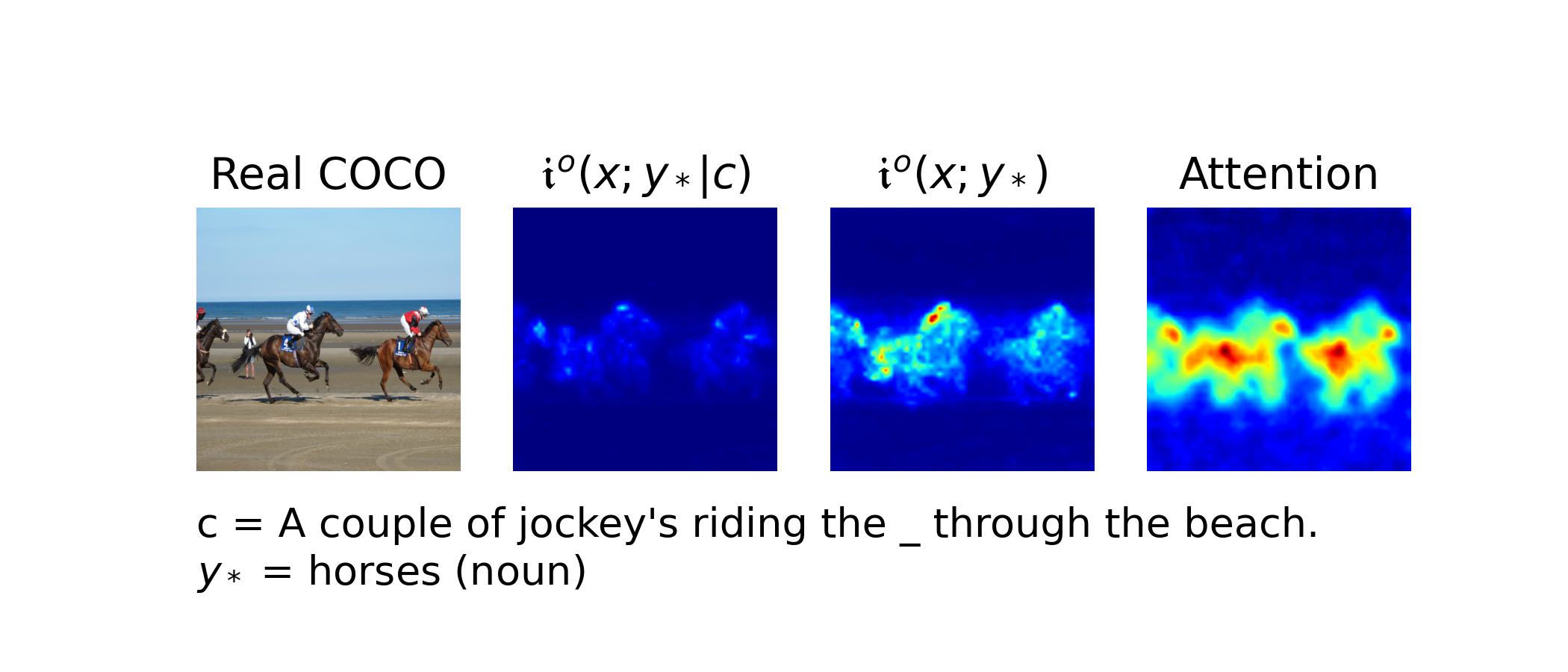} 
    \includegraphics[width=0.49\textwidth,trim={2cm 9mm 1.7cm 15mm},clip]{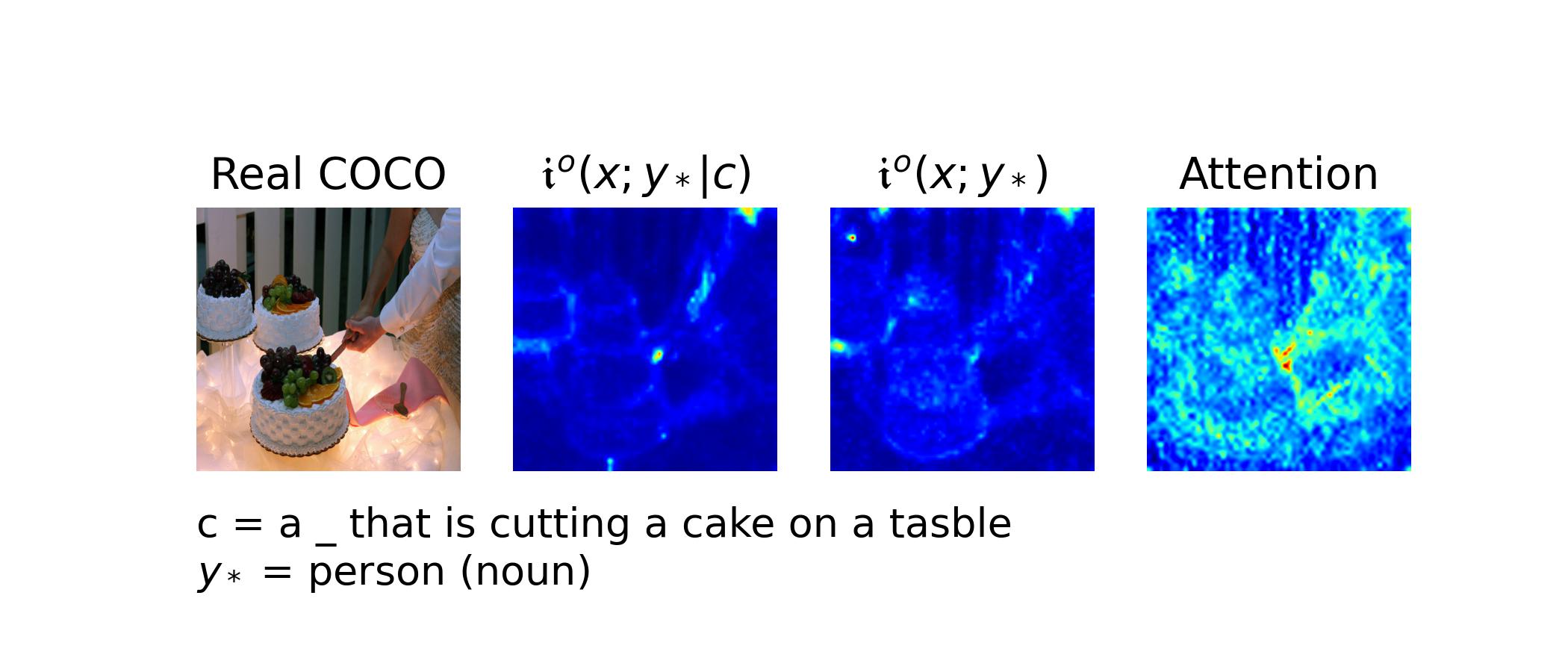} 
    \includegraphics[width=0.49\textwidth,trim={2cm 9mm 1.7cm 15mm},clip]{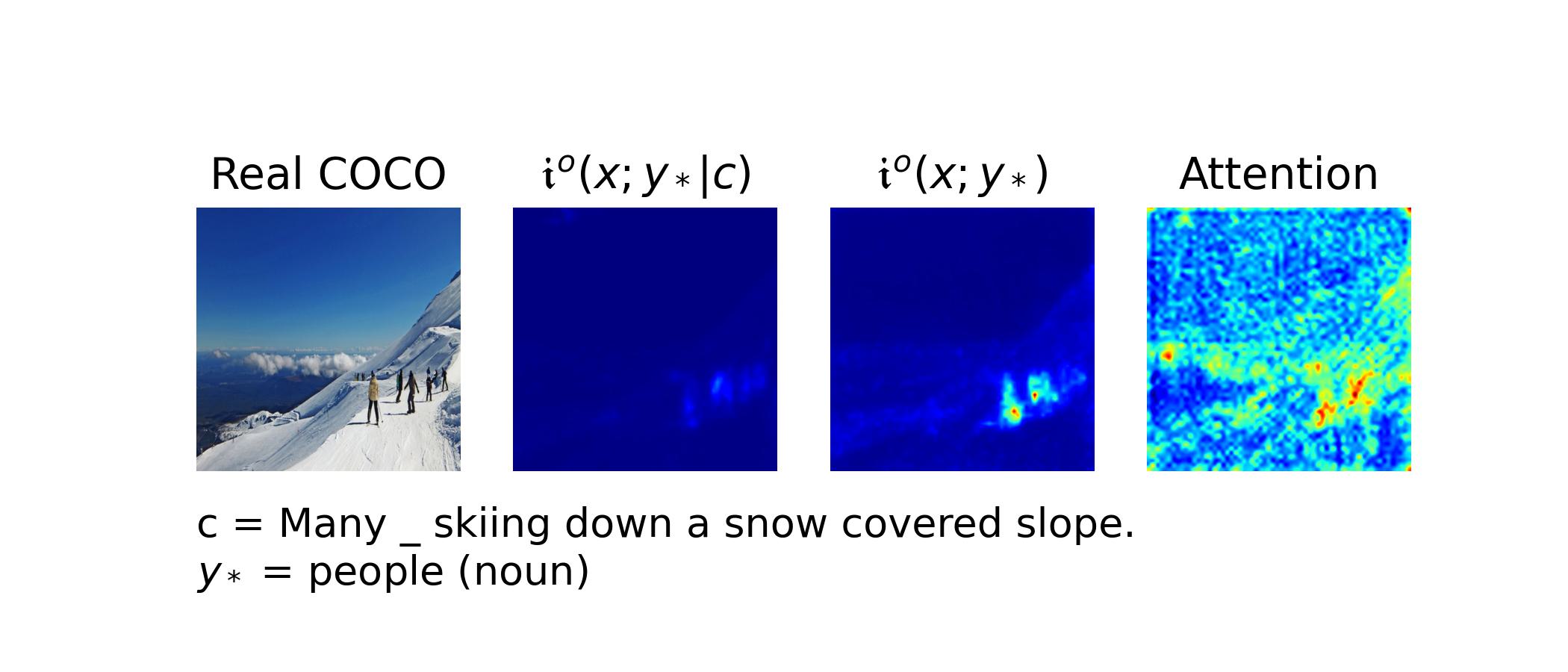} 
    \includegraphics[width=0.49\textwidth,trim={2cm 6mm 1.7cm 15mm},clip]{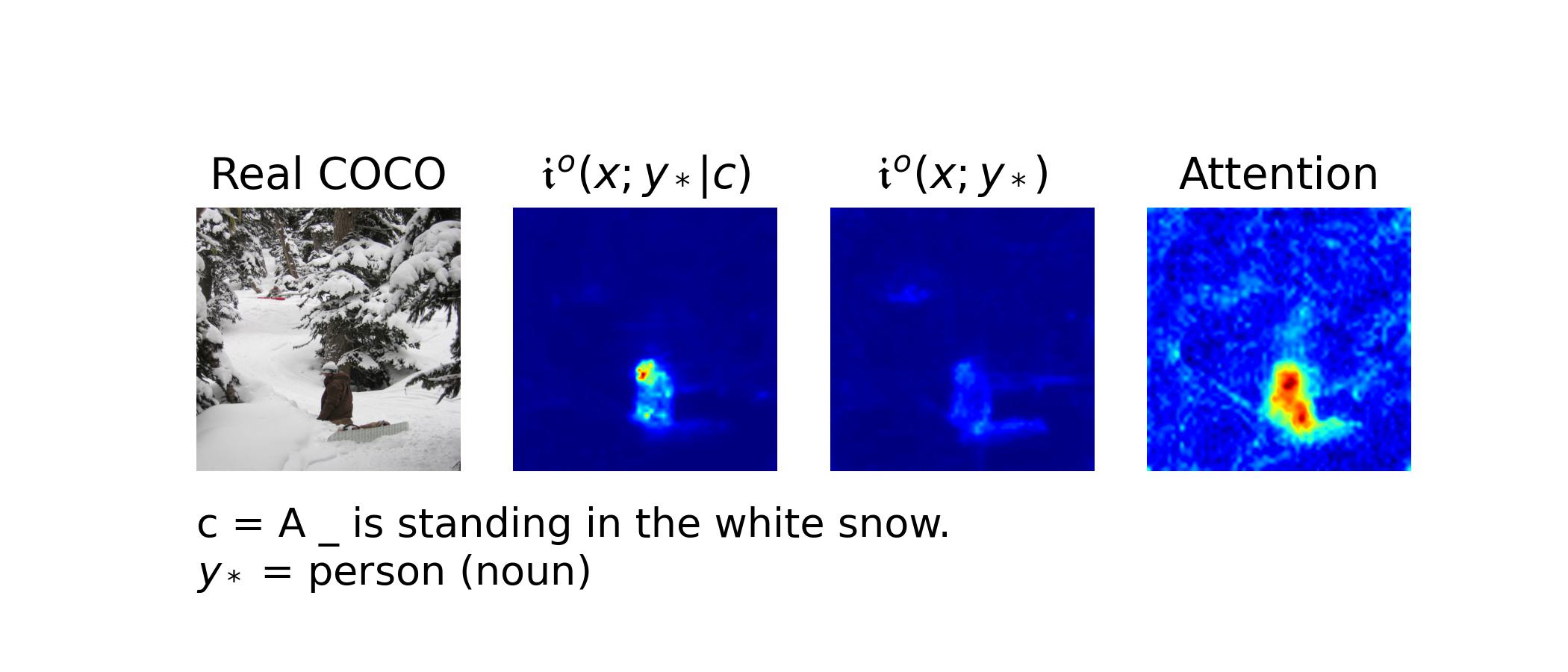} 
    \includegraphics[width=0.49\textwidth,trim={2cm 6mm 1.7cm 15mm},clip]{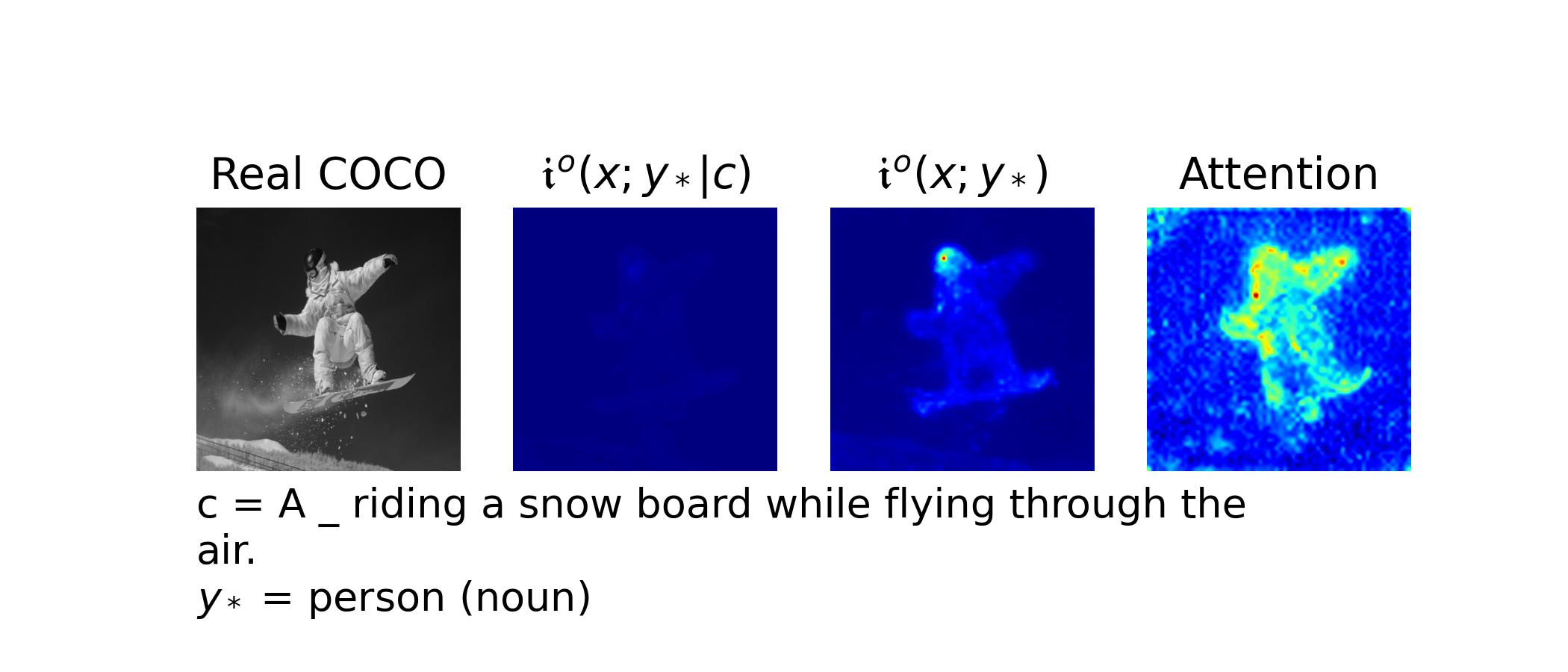} 
    \caption{Examples of localizing noun words in images.}
    \label{fig:2d_mi_cmi_attn_noun_2}
\end{figure}

\begin{figure}[t]
    \centering
    \includegraphics[width=0.49\textwidth,trim={2cm 9mm 1.7cm 15mm},clip]{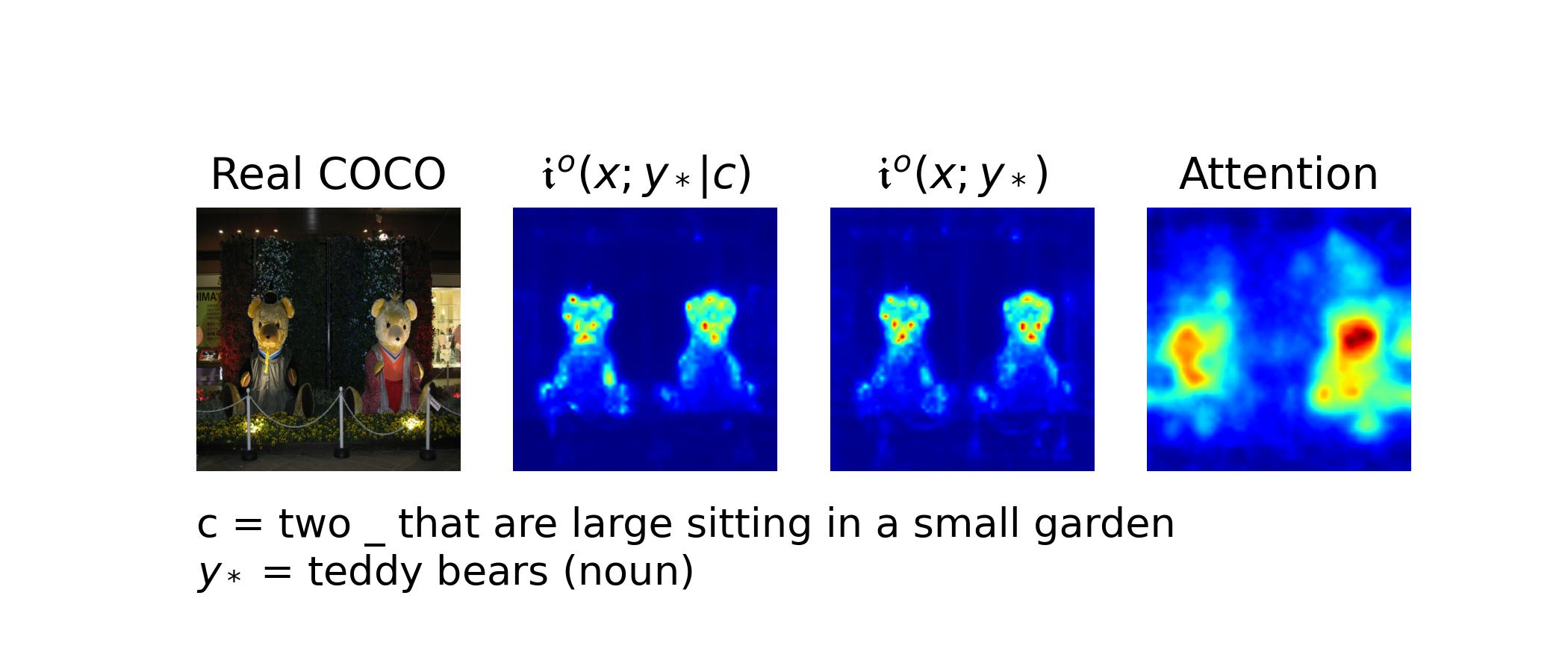} 
    \includegraphics[width=0.49\textwidth,trim={2cm 9mm 1.7cm 15mm},clip]{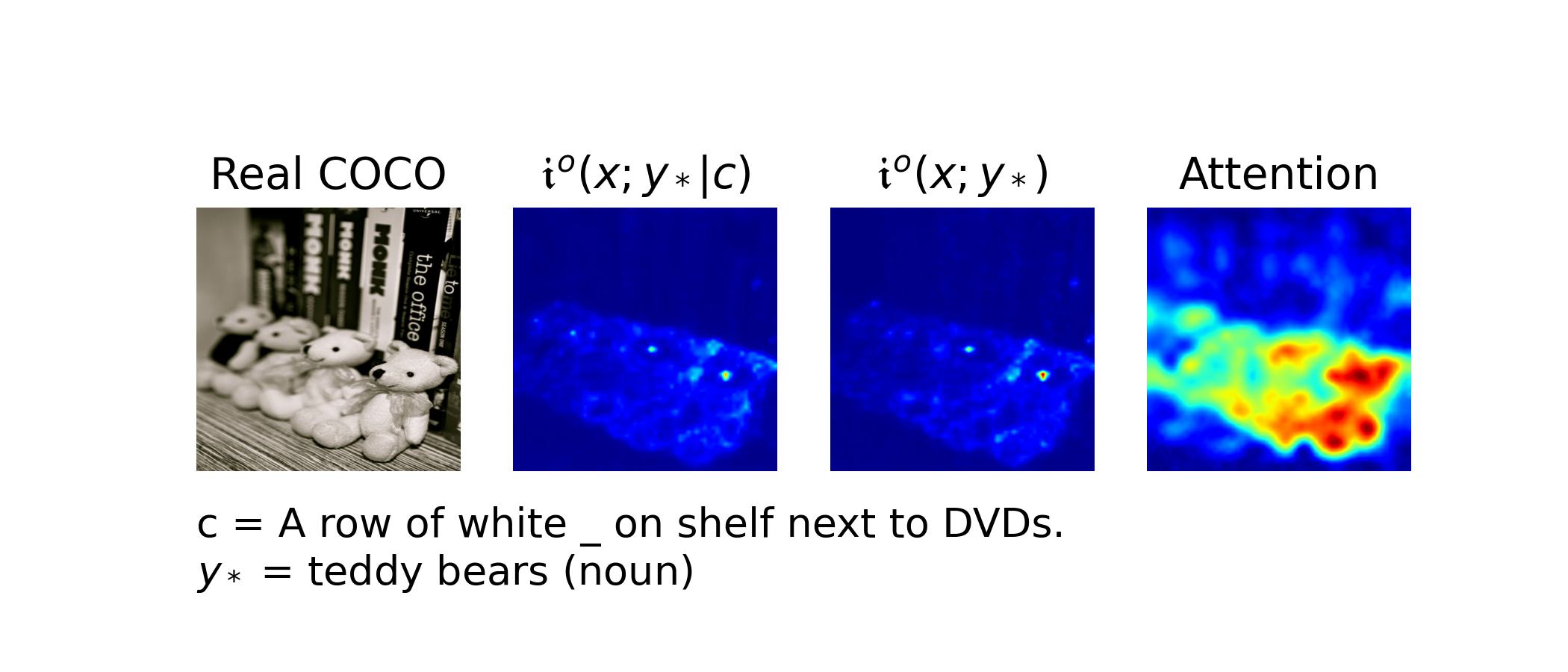} 
    \includegraphics[width=0.49\textwidth,trim={2cm 9mm 1.7cm 15mm},clip]{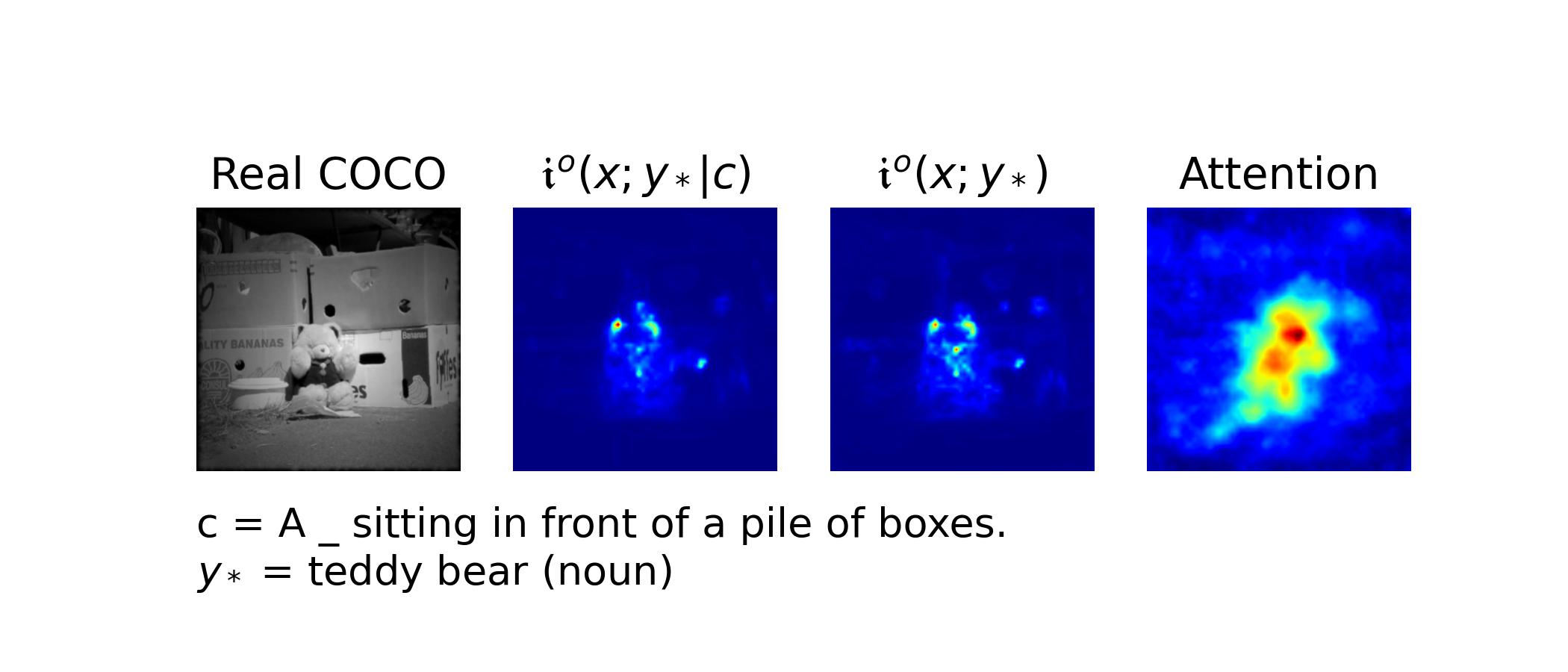} 
    \includegraphics[width=0.49\textwidth,trim={2cm 9mm 1.7cm 15mm},clip]{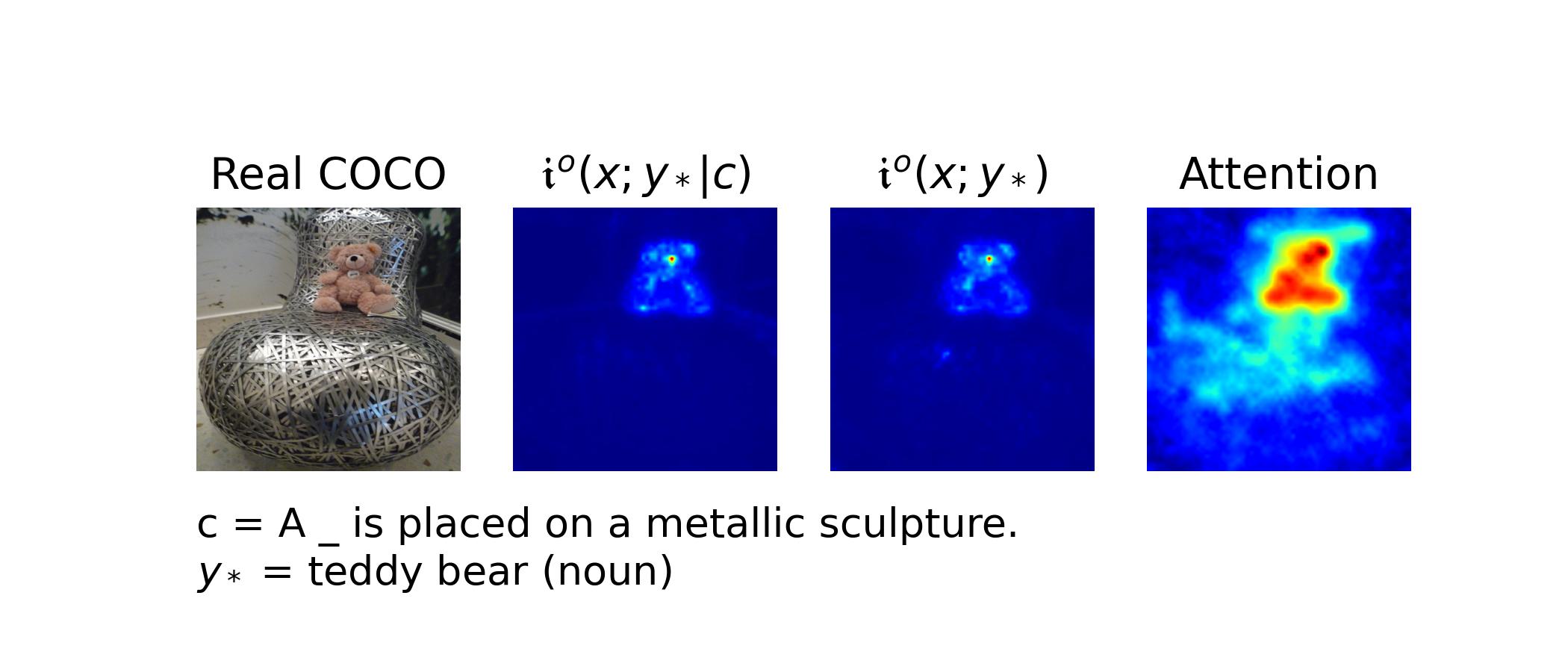} 
    \includegraphics[width=0.49\textwidth,trim={2cm 9mm 1.7cm 15mm},clip]{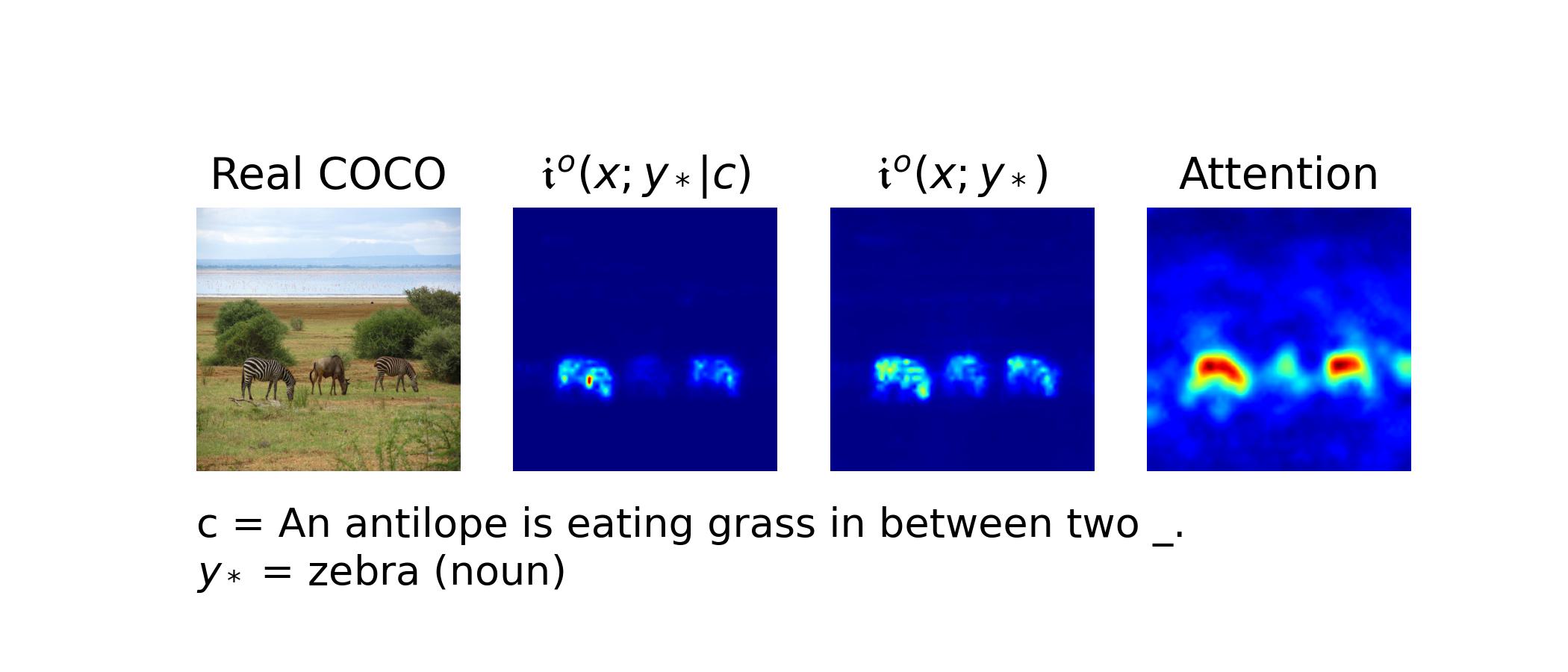} 
    \includegraphics[width=0.49\textwidth,trim={2cm 9mm 1.7cm 15mm},clip]{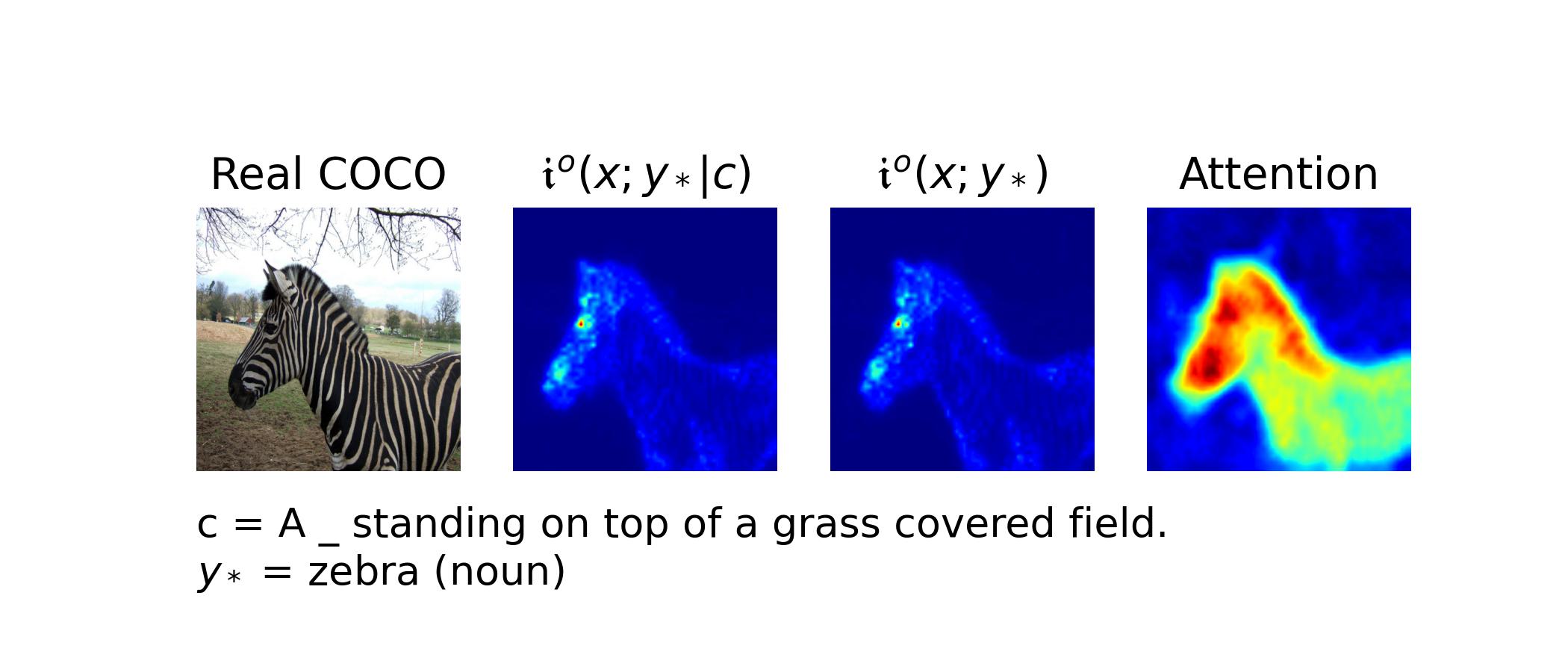} 
    \includegraphics[width=0.49\textwidth,trim={2cm 9mm 1.7cm 15mm},clip]{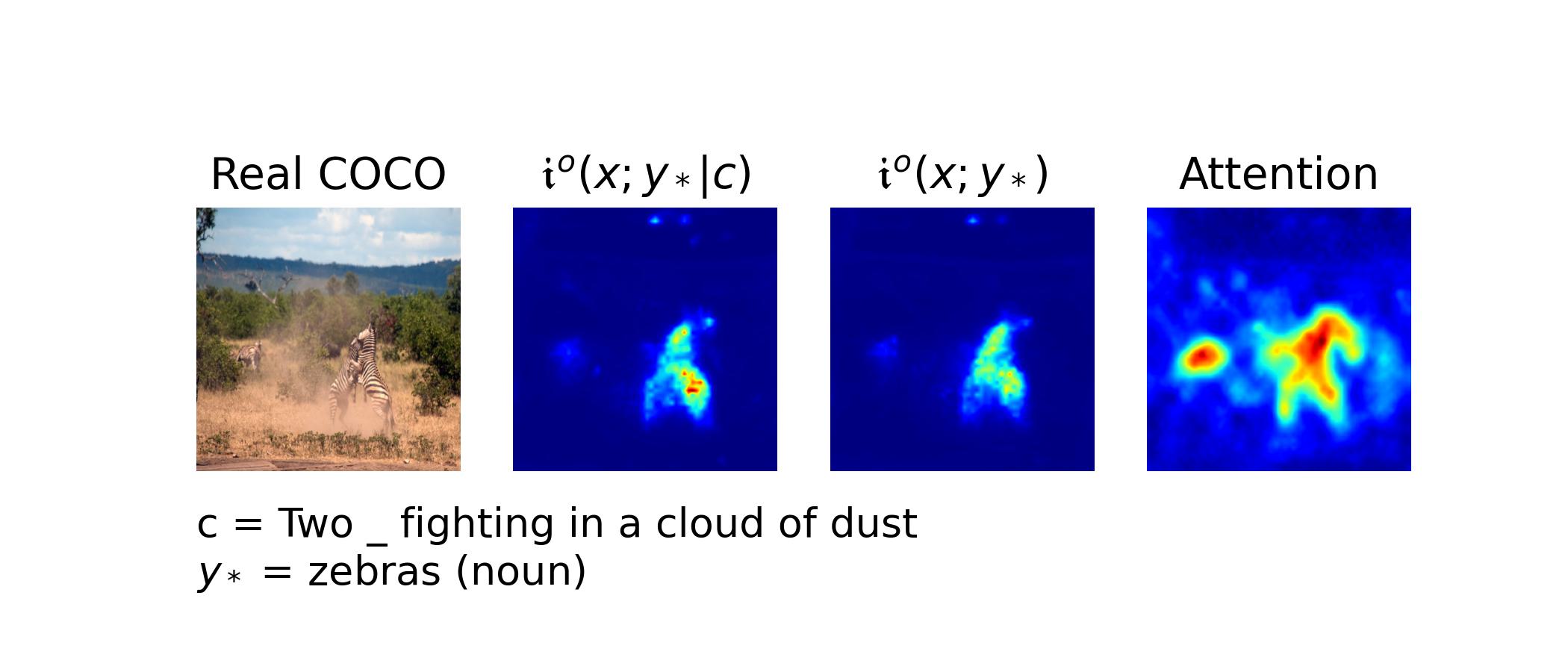} 
    \includegraphics[width=0.49\textwidth,trim={2cm 9mm 1.7cm 15mm},clip]{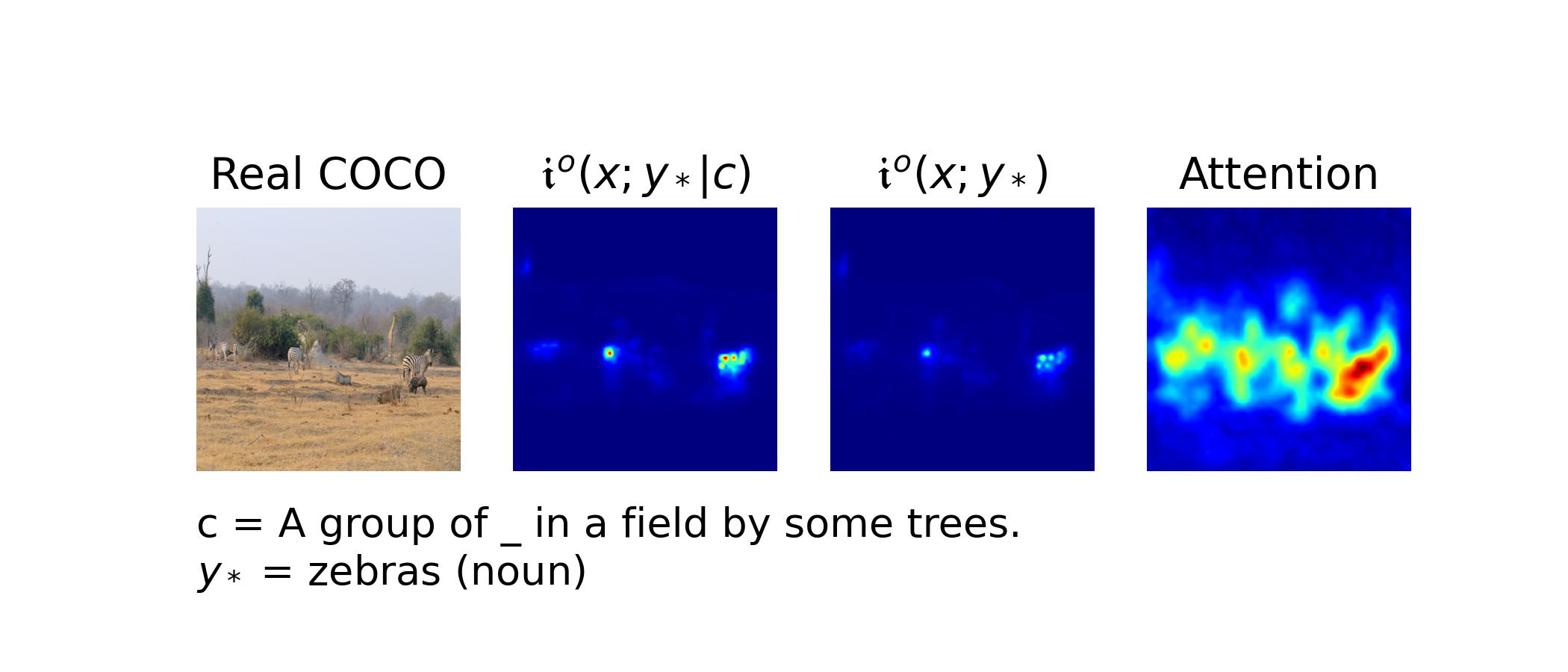} 
    \caption{Examples of localizing noun words in images.}
    \label{fig:2d_mi_cmi_attn_noun_3}
\end{figure}

We put more pixel-level MI and CMI visualization examples from COCO100-IT, see Fig. \ref{fig:2d_mi_cmi_attn_noun_1} \& \ref{fig:2d_mi_cmi_attn_noun_2} \& \ref{fig:2d_mi_cmi_attn_noun_3}.

\subsection{Examples of word location for other seven entities} \label{app:2d_mi_cmi_attn_7}
\begin{figure}[t]
    \centering
    \includegraphics[width=0.49\textwidth,trim={2cm 6mm 1.7cm 15mm},clip]{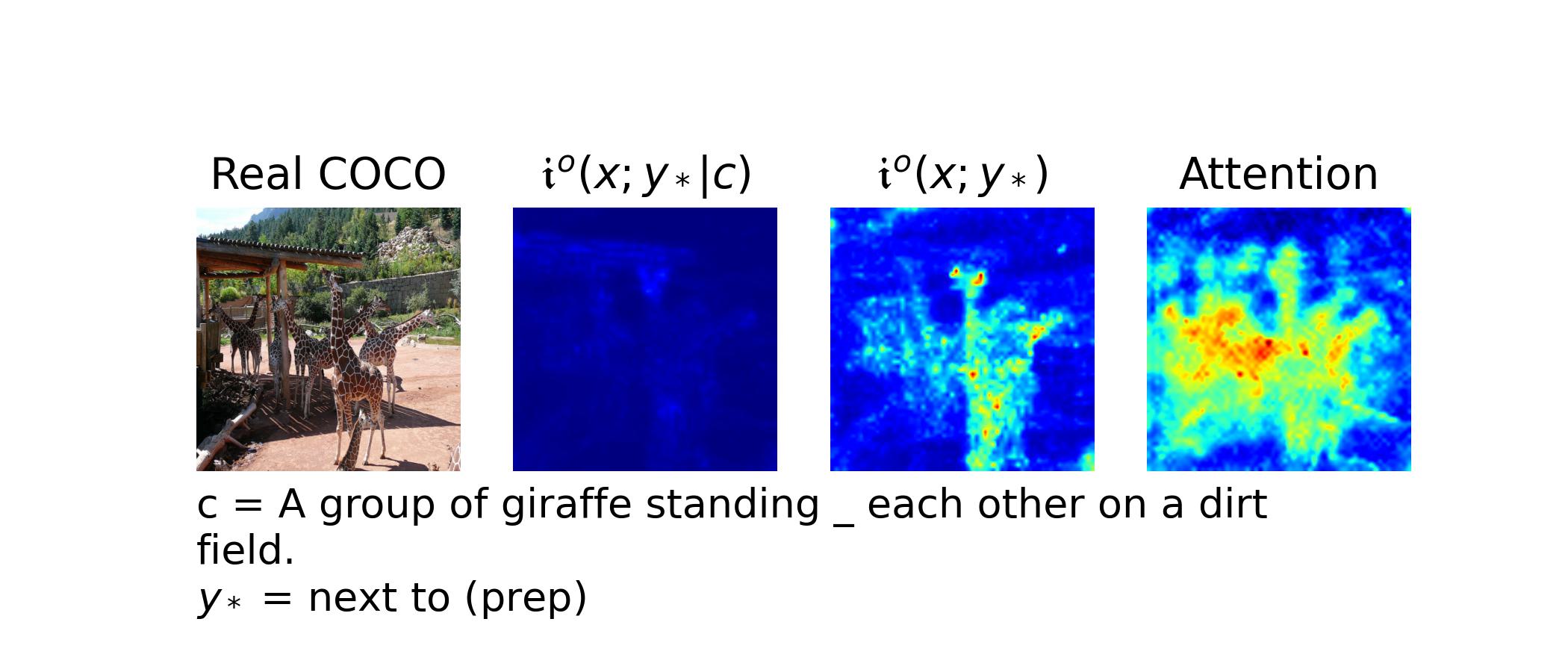} 
    \includegraphics[width=0.49\textwidth,trim={2cm 6mm 1.7cm 15mm},clip]{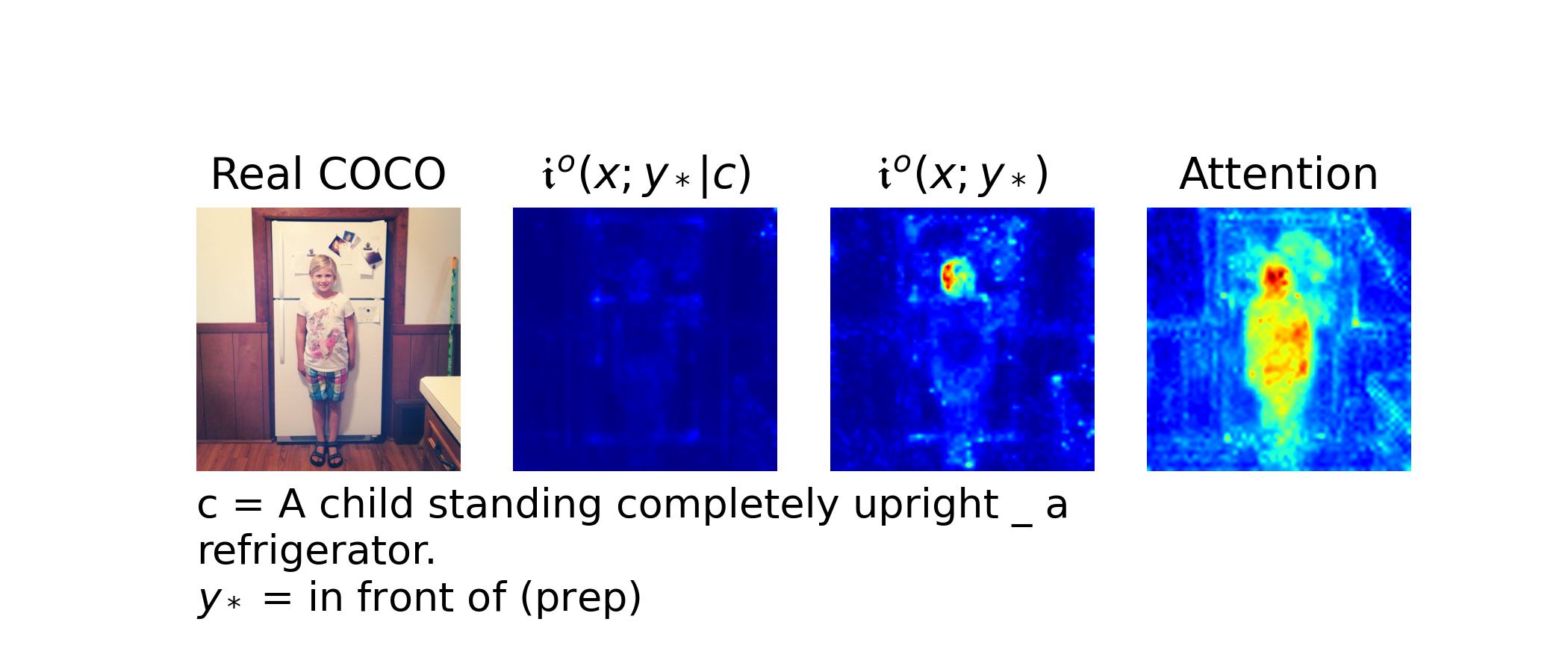} 
    \includegraphics[width=0.49\textwidth,trim={2cm 9mm 1.7cm 15mm},clip]{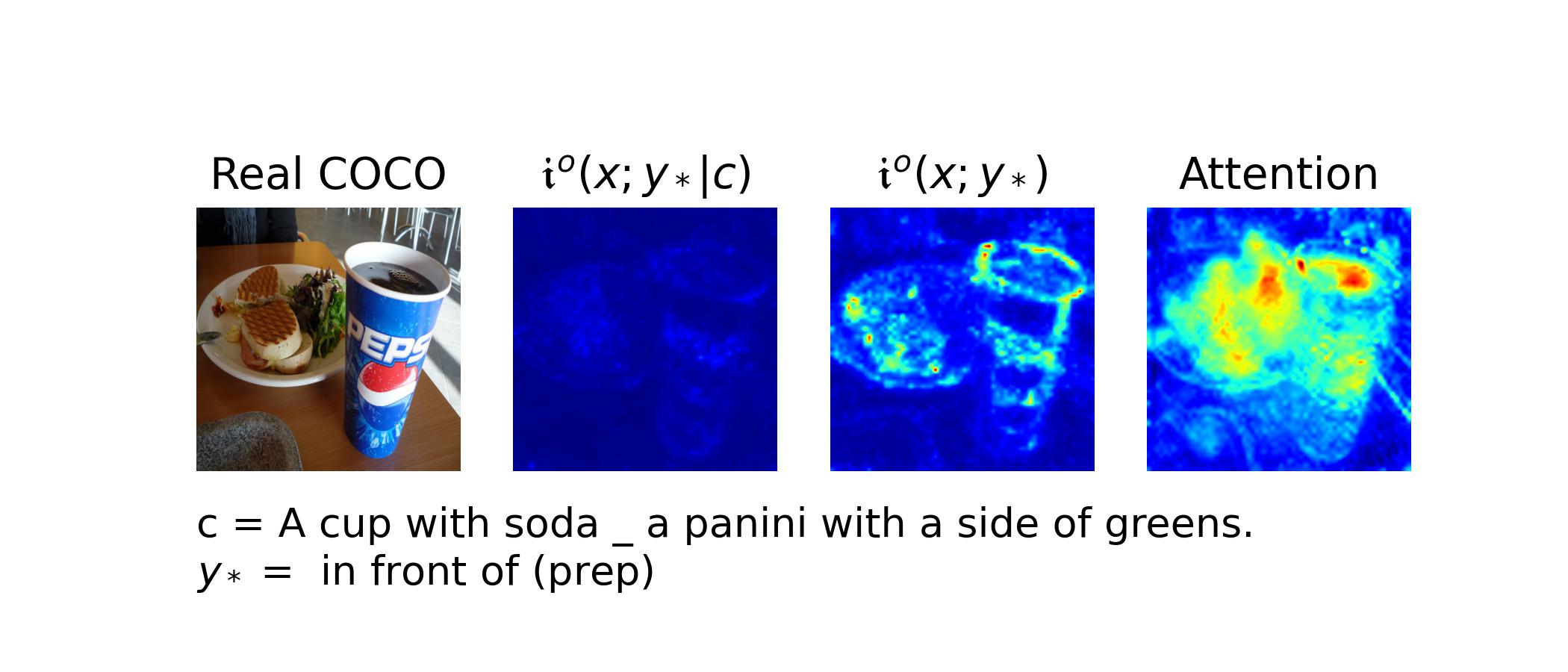} 
    \includegraphics[width=0.49\textwidth,trim={2cm 9mm 1.7cm 15mm},clip]{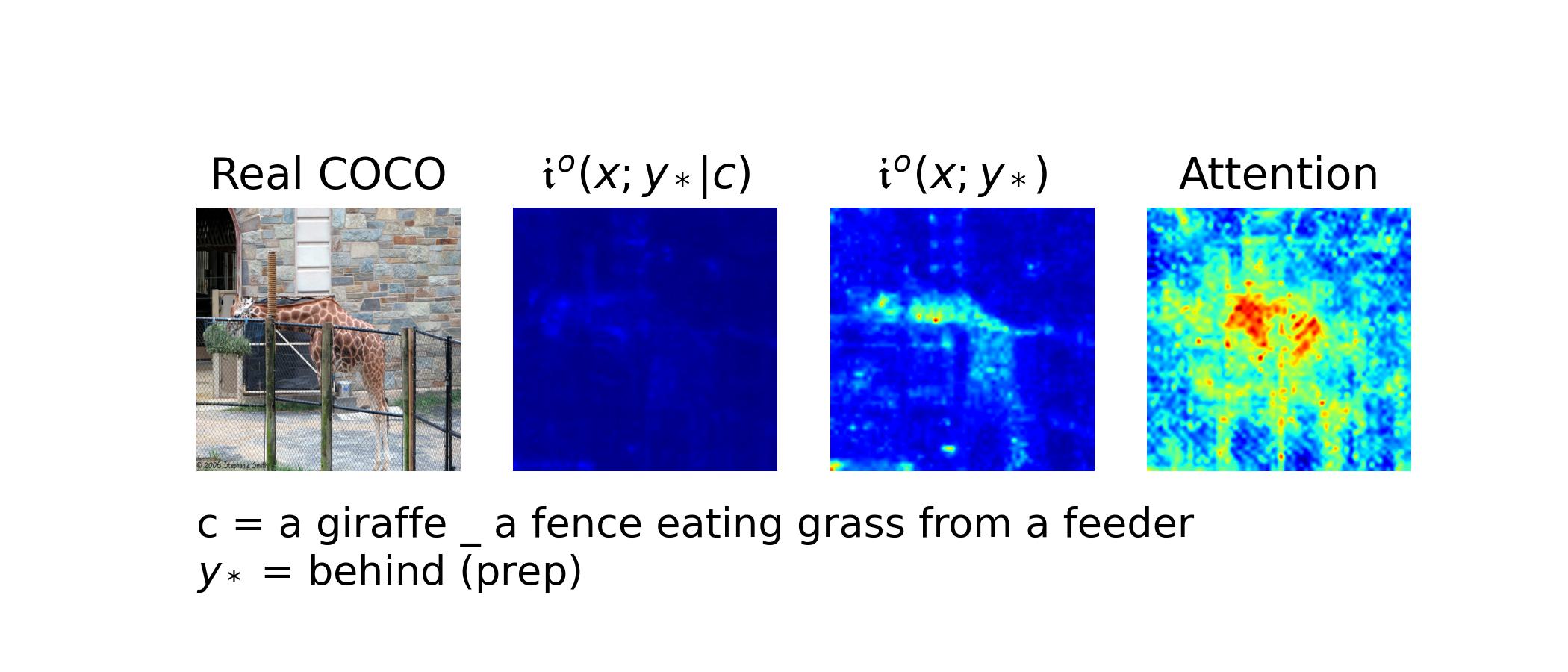} 
    \includegraphics[width=0.49\textwidth,trim={2cm 6mm 1.7cm 15mm},clip]{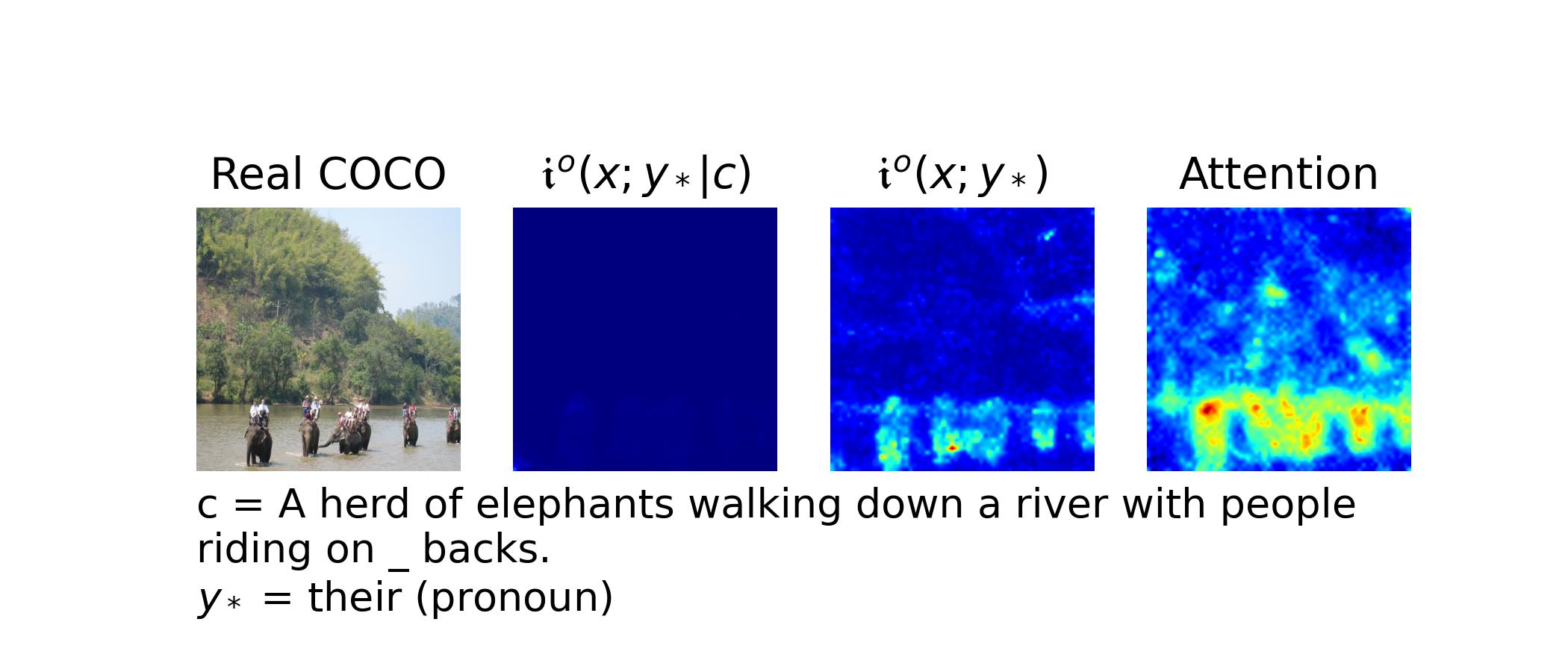} 
    \includegraphics[width=0.49\textwidth,trim={2cm 6mm 1.7cm 15mm},clip]{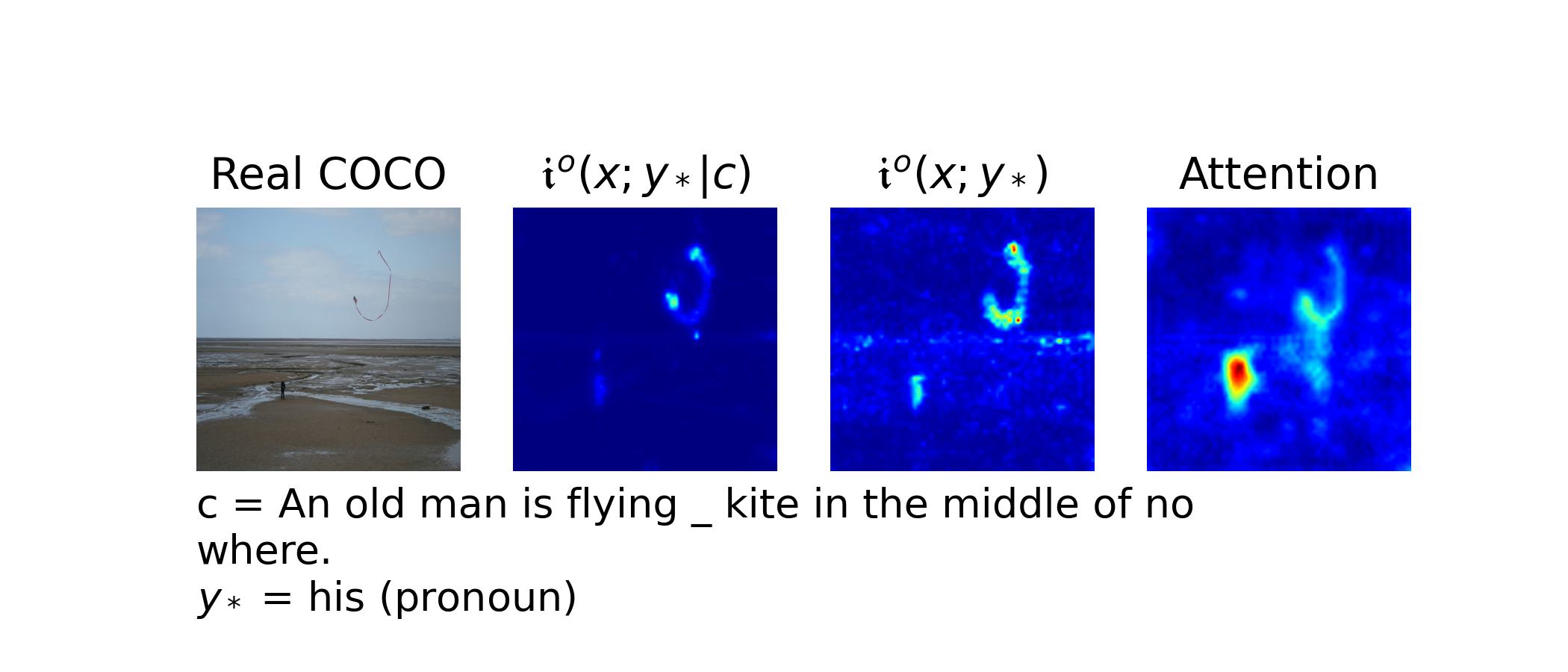} 
    \includegraphics[width=0.49\textwidth,trim={2cm 9mm 1.7cm 15mm},clip]{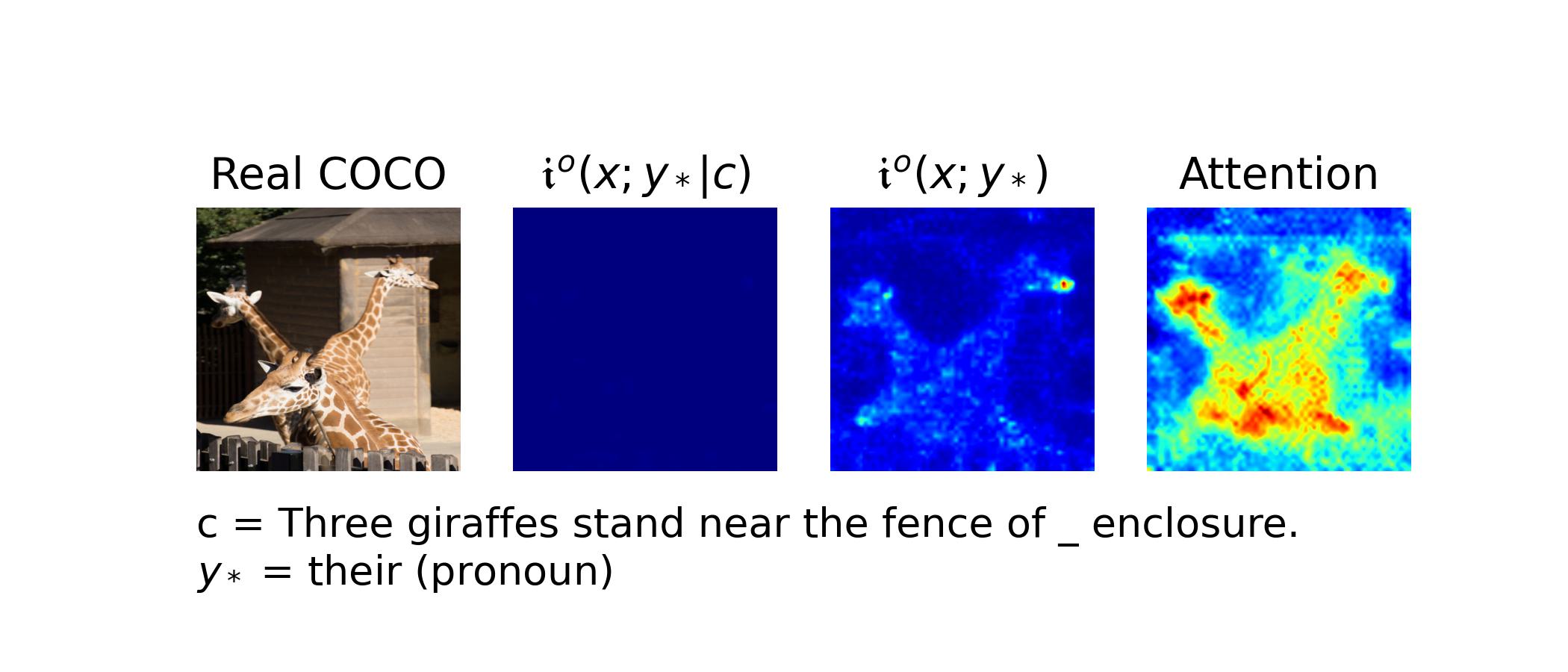} 
    \includegraphics[width=0.49\textwidth,trim={2cm 9mm 1.7cm 15mm},clip]{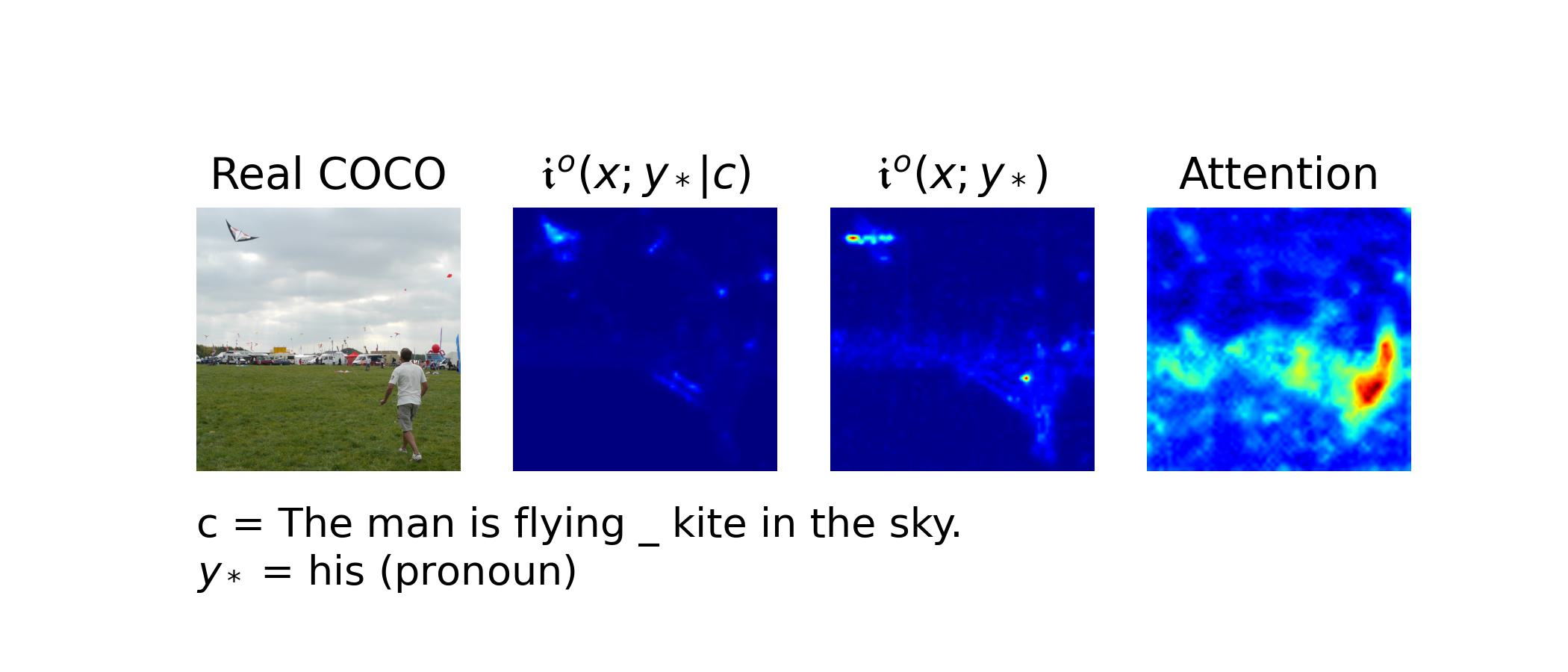} 
    \includegraphics[width=0.49\textwidth,trim={2cm 6mm 1.7cm 15mm},clip]{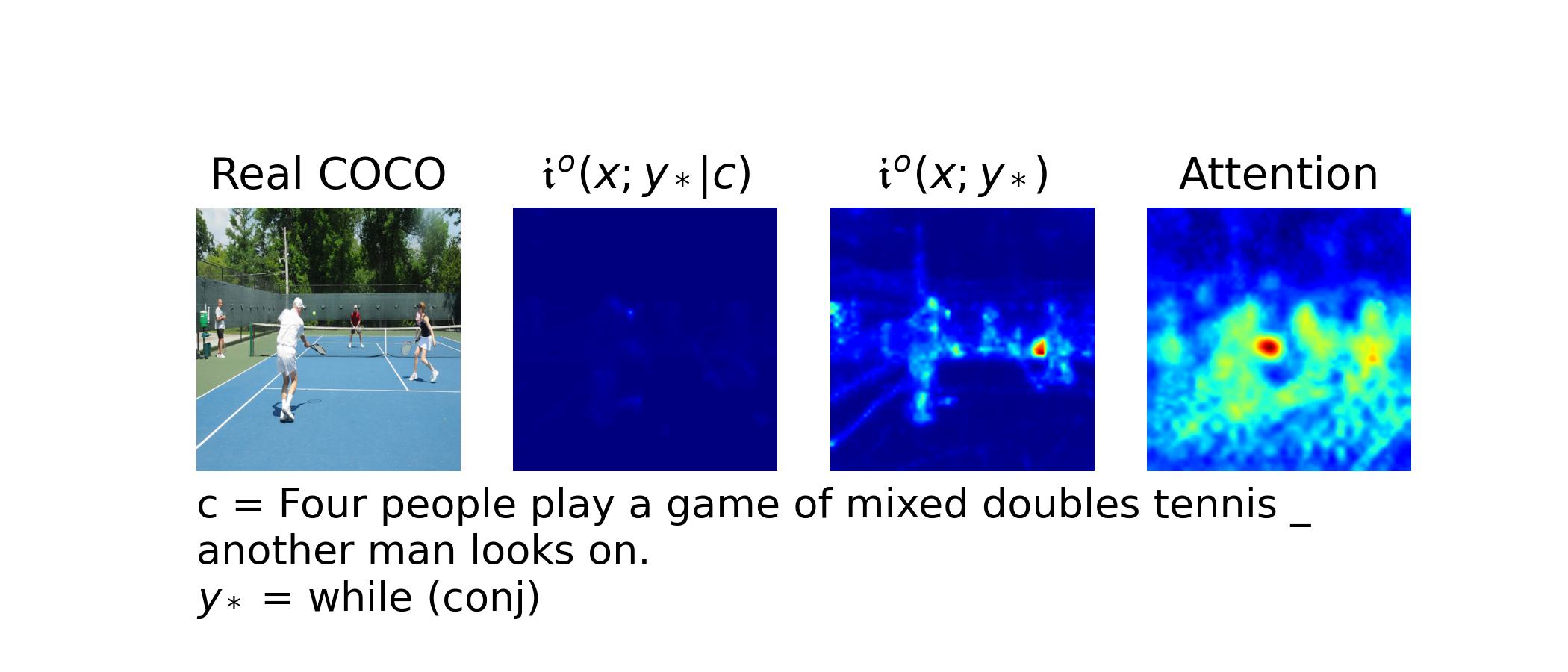} 
    \includegraphics[width=0.49\textwidth,trim={2cm 6mm 1.7cm 15mm},clip]{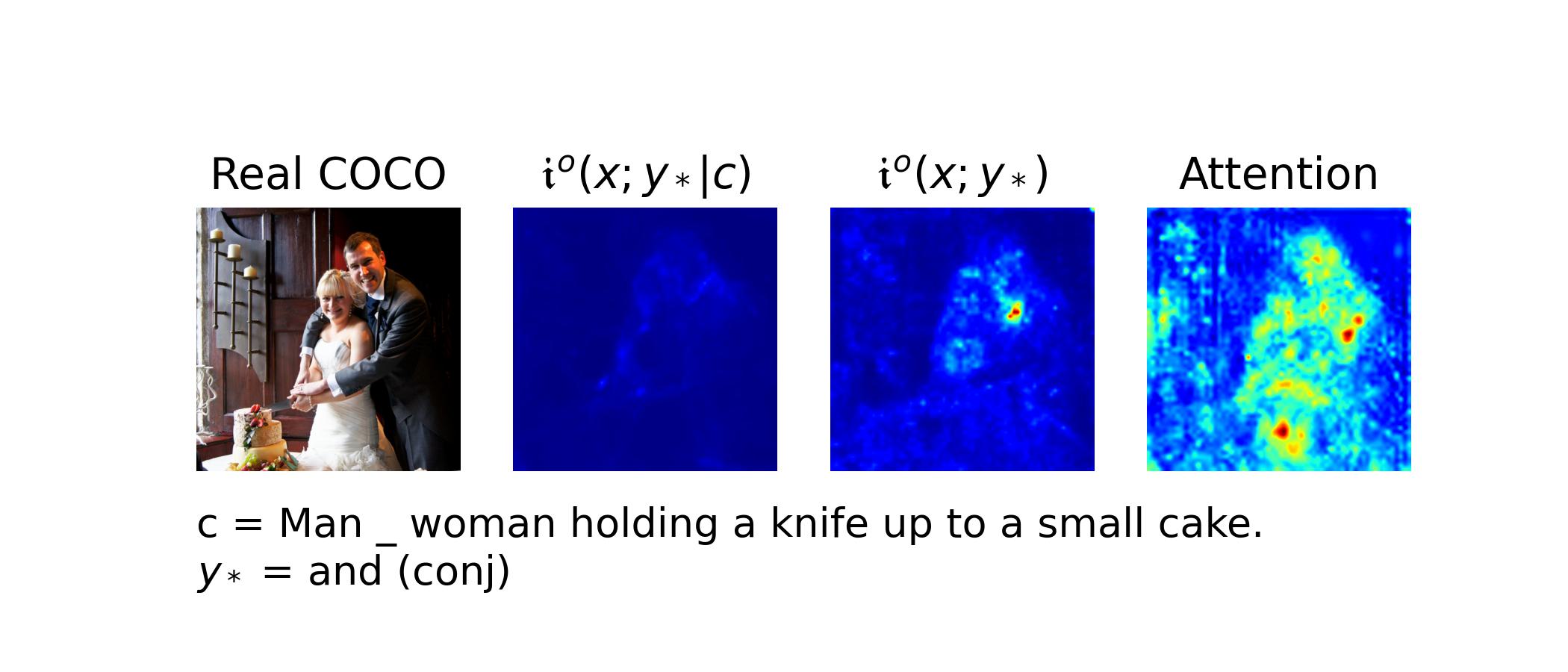} 
    \includegraphics[width=0.49\textwidth,trim={2cm 6mm 1.7cm 15mm},clip]{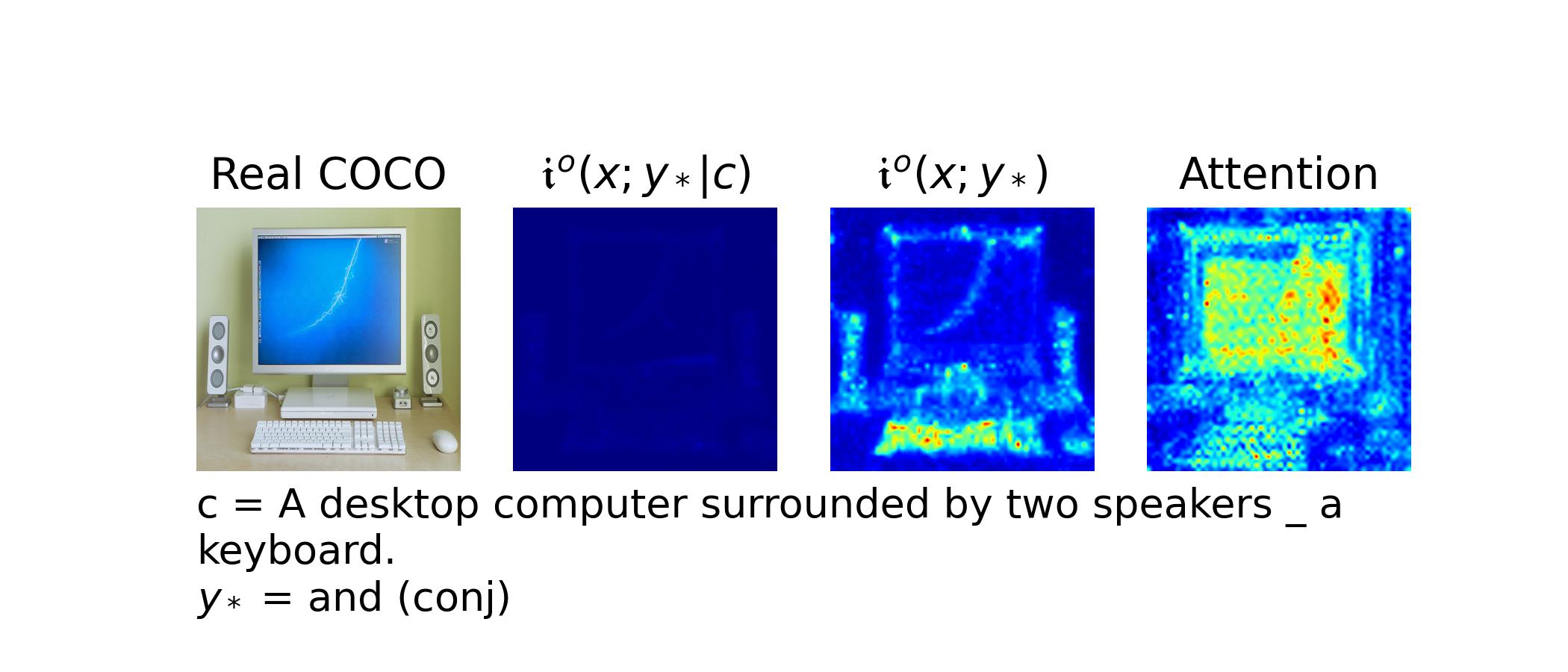} 
    \includegraphics[width=0.49\textwidth,trim={2cm 6mm 1.7cm 15mm},clip]{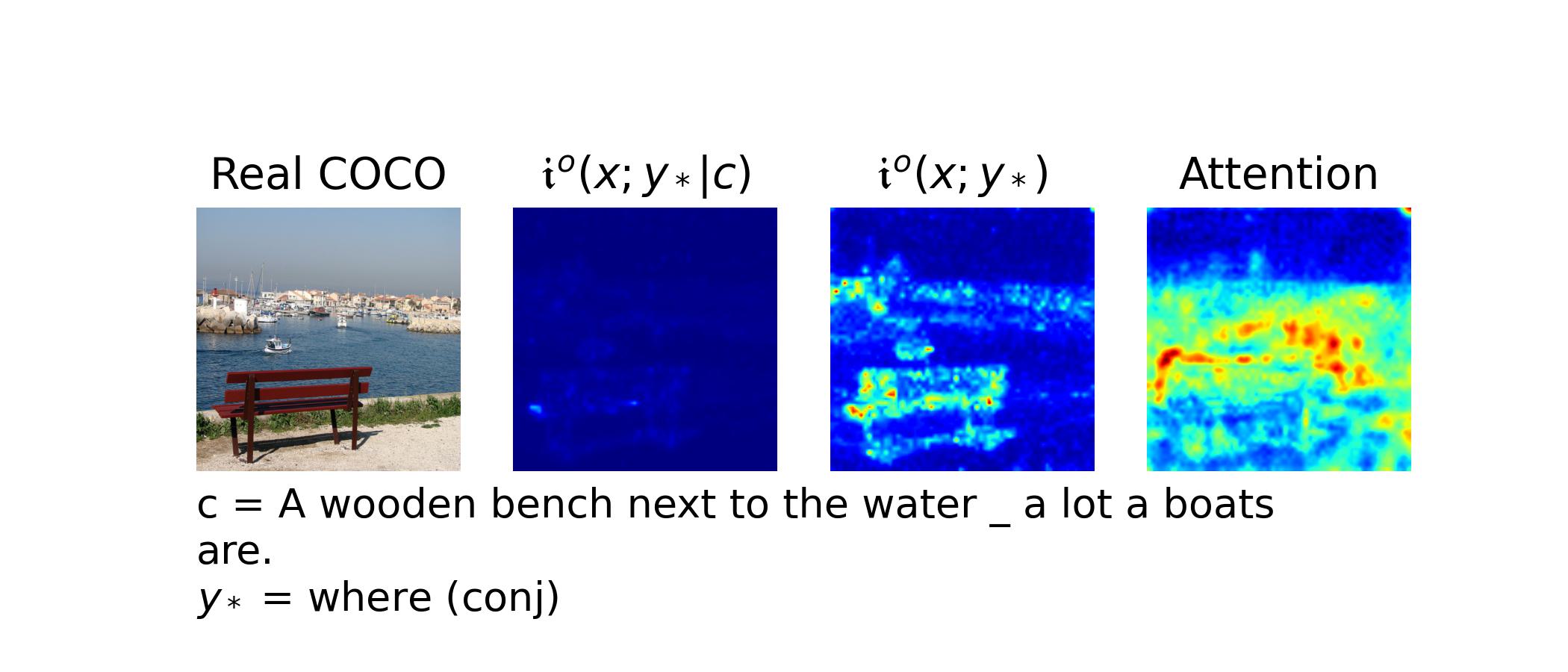} 
    \caption{Examples of localizing abstract words in images.}
    \label{fig:2d_mi_cmi_attn_7_2}
\end{figure}

\begin{figure}[t]
    \centering
    \includegraphics[width=0.49\textwidth,trim={2cm 9mm 1.7cm 15mm},clip]{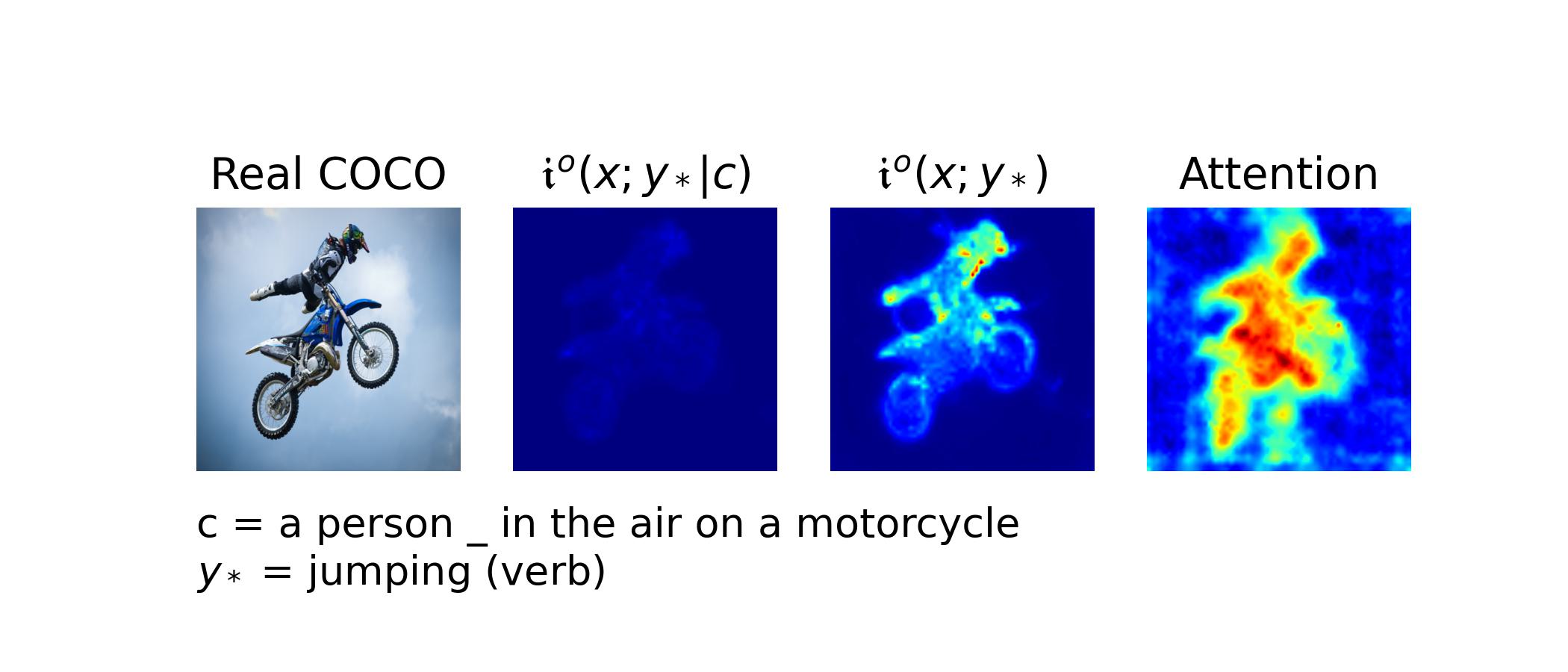} 
    \includegraphics[width=0.49\textwidth,trim={2cm 9mm 1.7cm 15mm},clip]{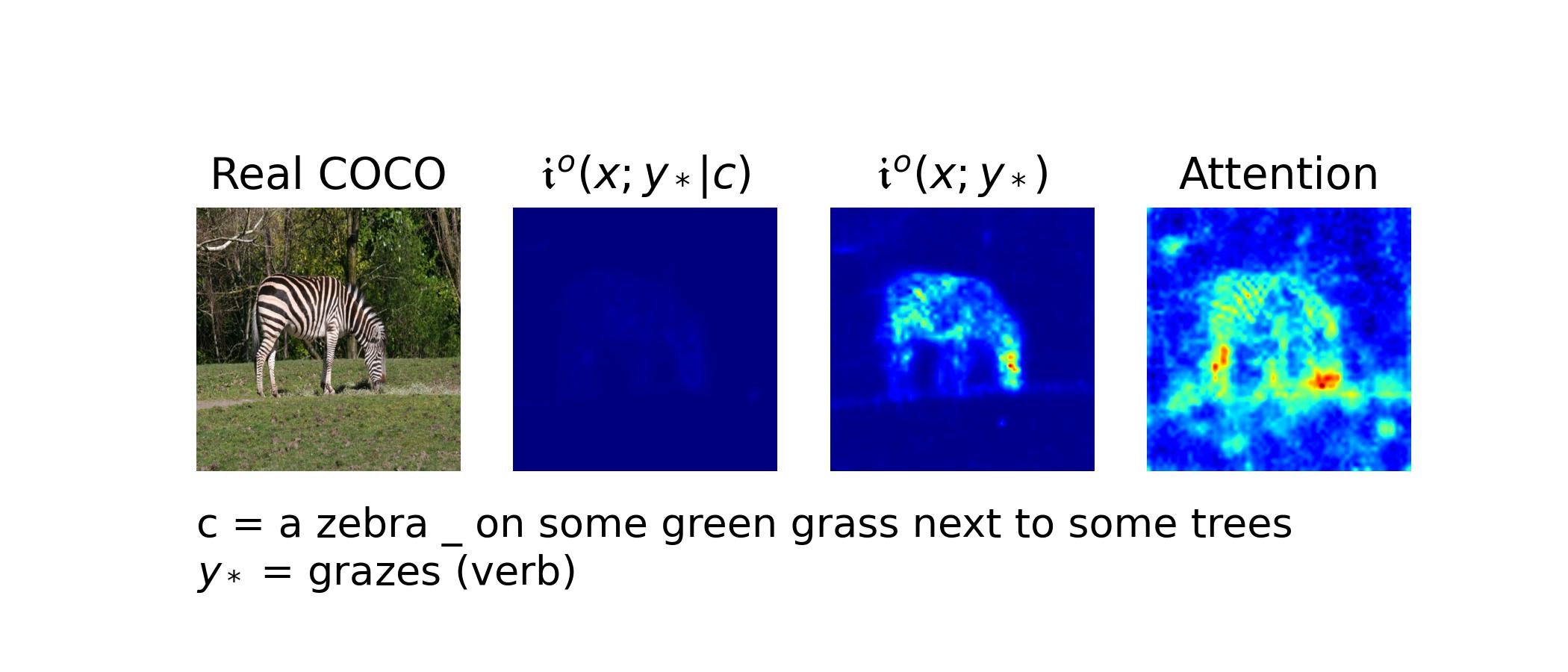} 
    \includegraphics[width=0.49\textwidth,trim={2cm 9mm 1.7cm 15mm},clip]{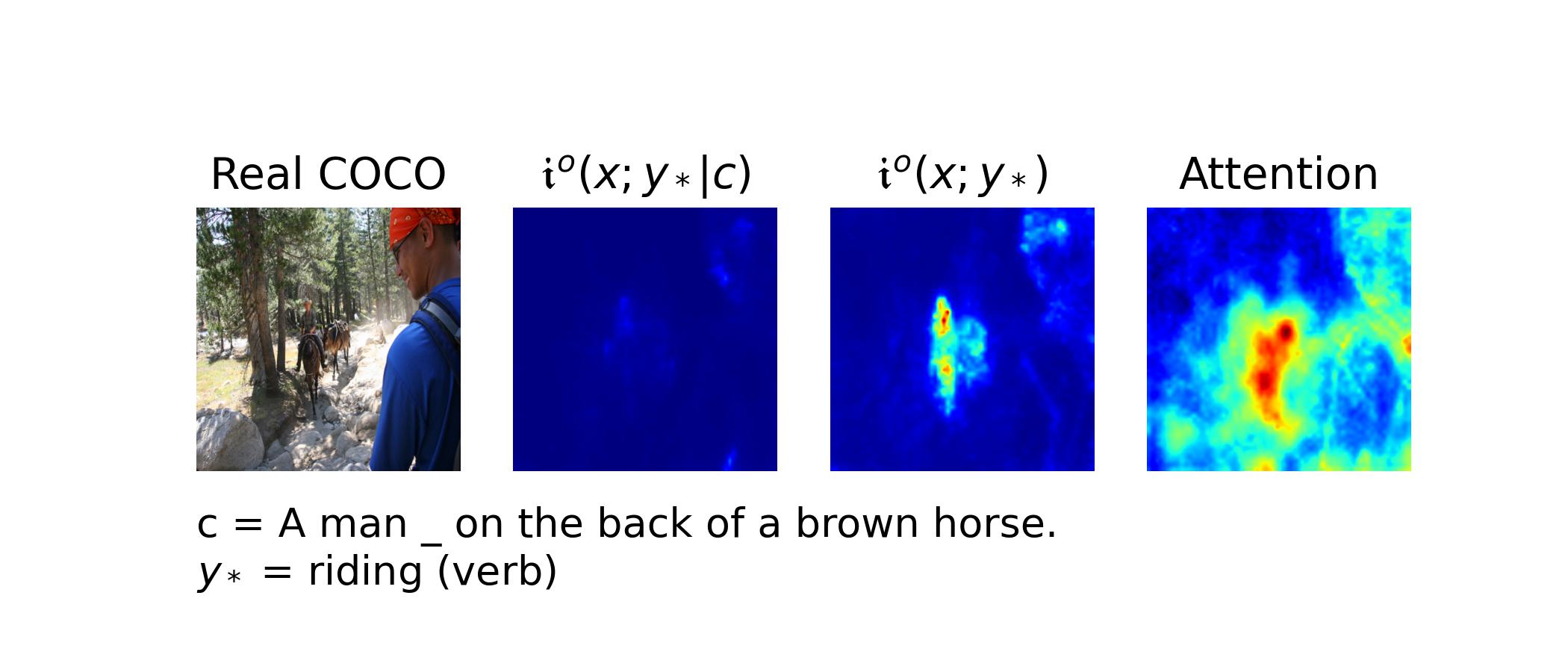} 
    \includegraphics[width=0.49\textwidth,trim={2cm 9mm 1.7cm 15mm},clip]{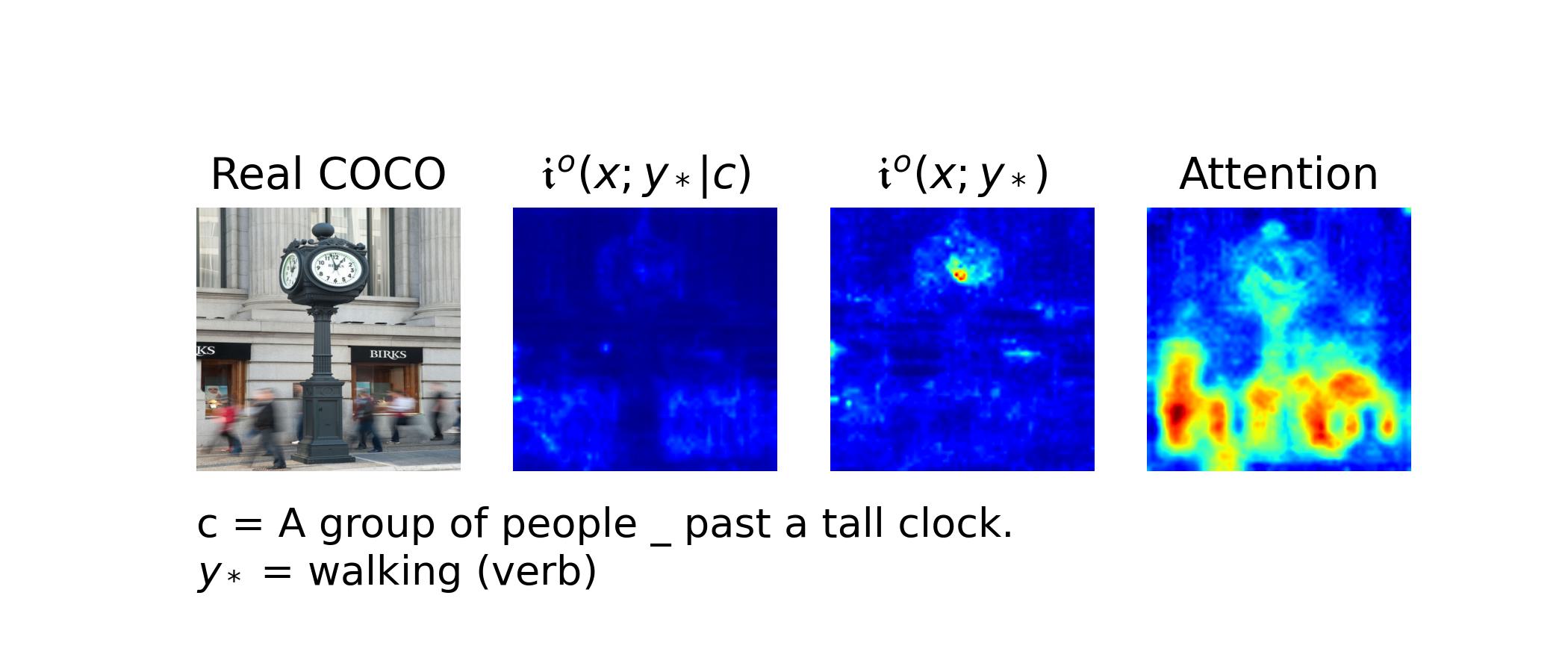} 
    \includegraphics[width=0.49\textwidth,trim={2cm 6mm 1.7cm 15mm},clip]{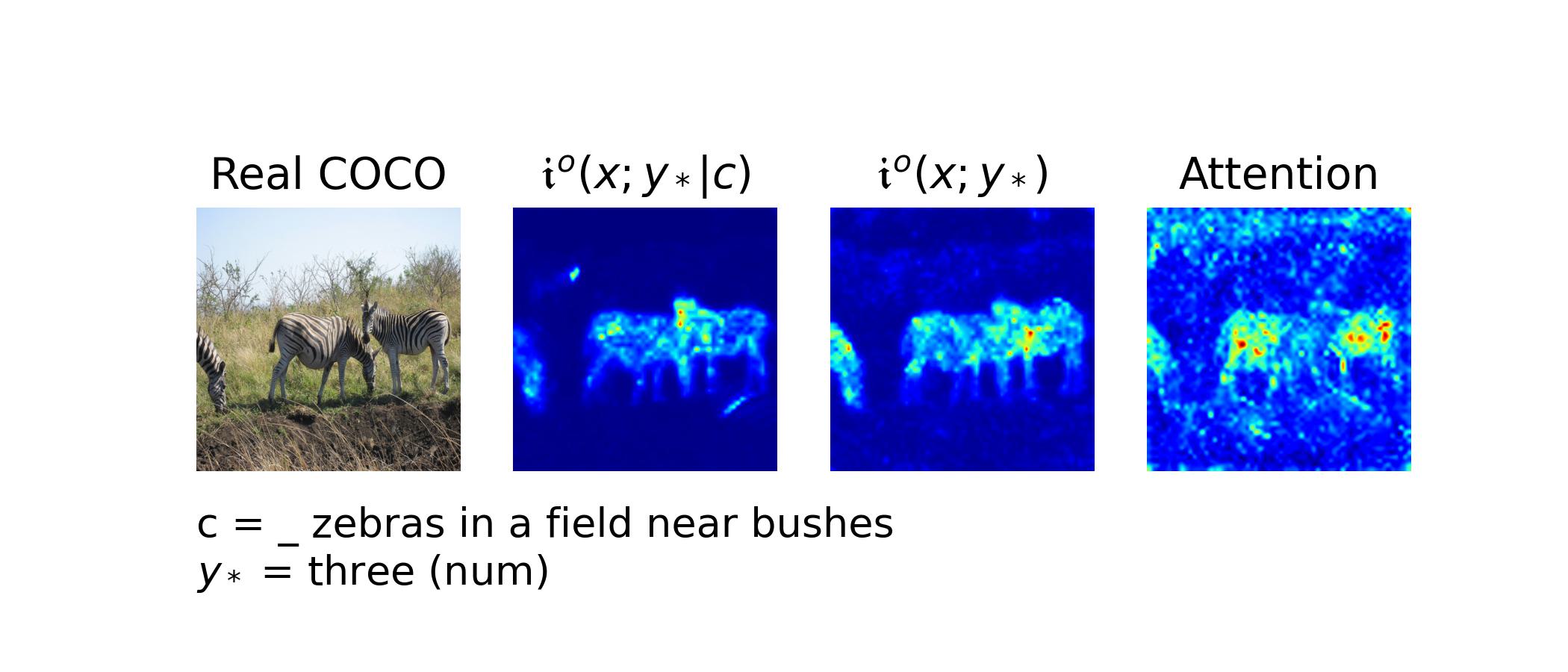} 
    \includegraphics[width=0.49\textwidth,trim={2cm 6mm 1.7cm 15mm},clip]{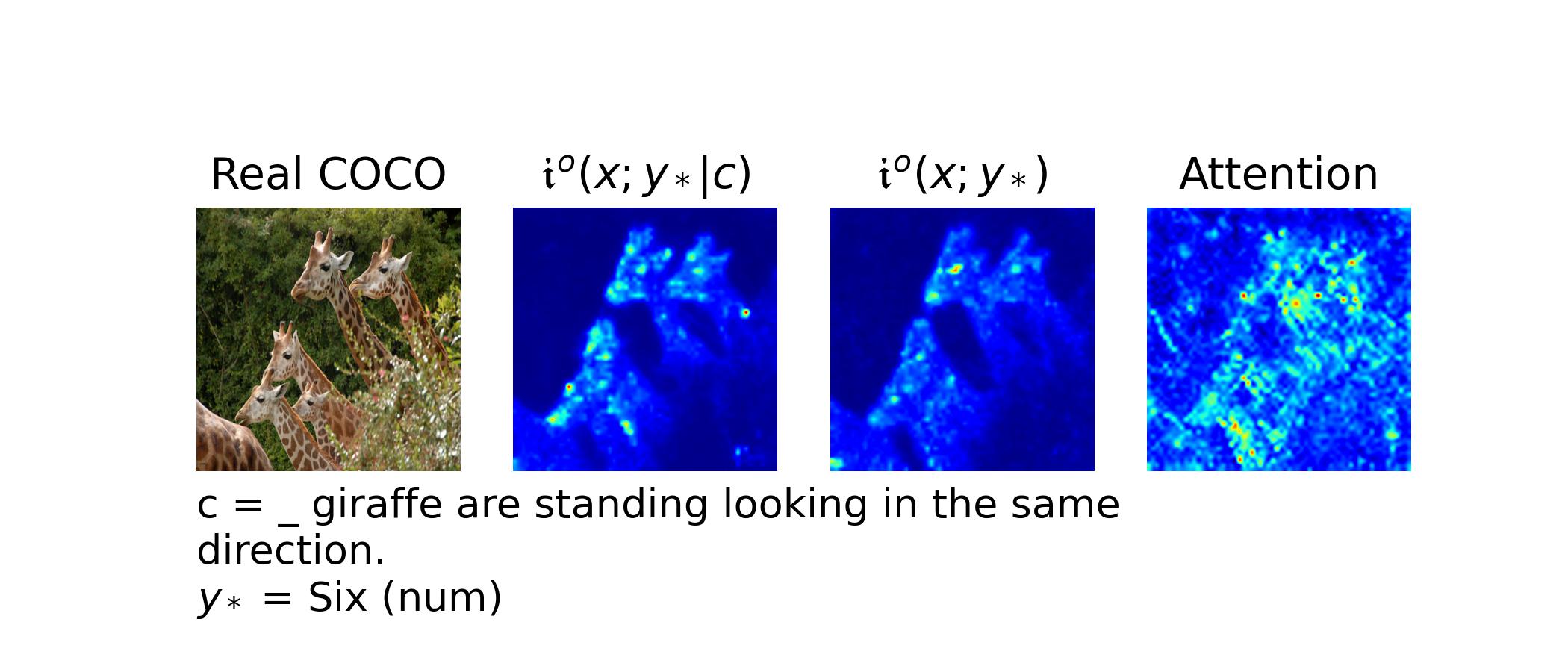} 
    \includegraphics[width=0.49\textwidth,trim={2cm 9mm 1.7cm 15mm},clip]{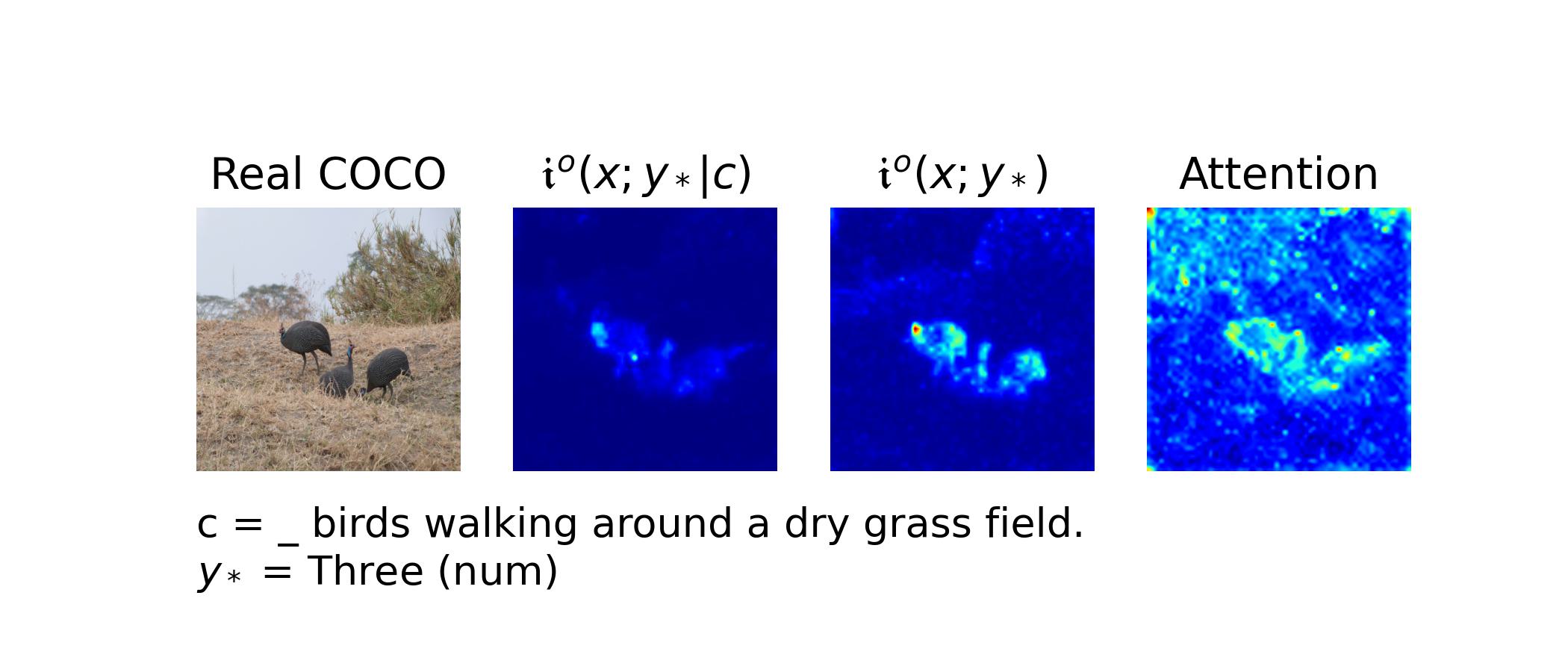} 
    \includegraphics[width=0.49\textwidth,trim={2cm 9mm 1.7cm 15mm},clip]{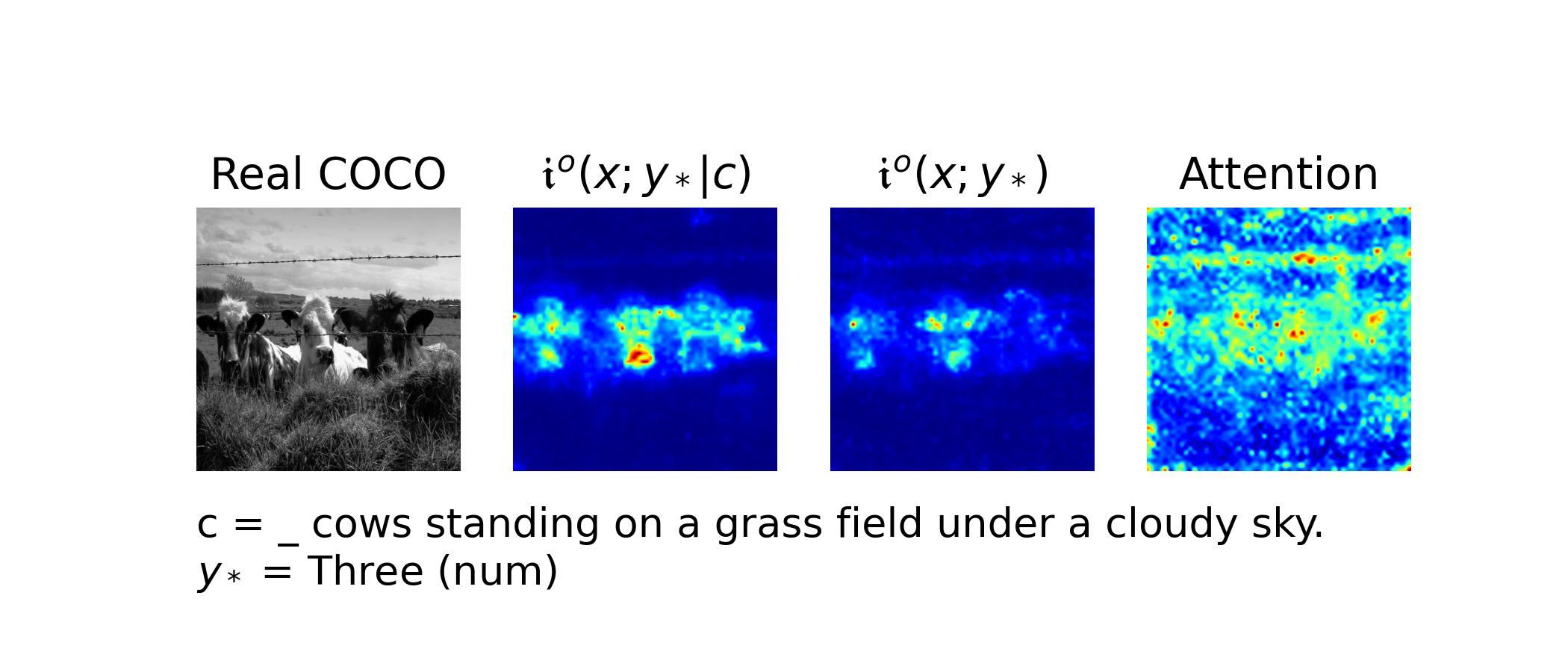} 
    \includegraphics[width=0.49\textwidth,trim={2cm 9mm 1.7cm 15mm},clip]{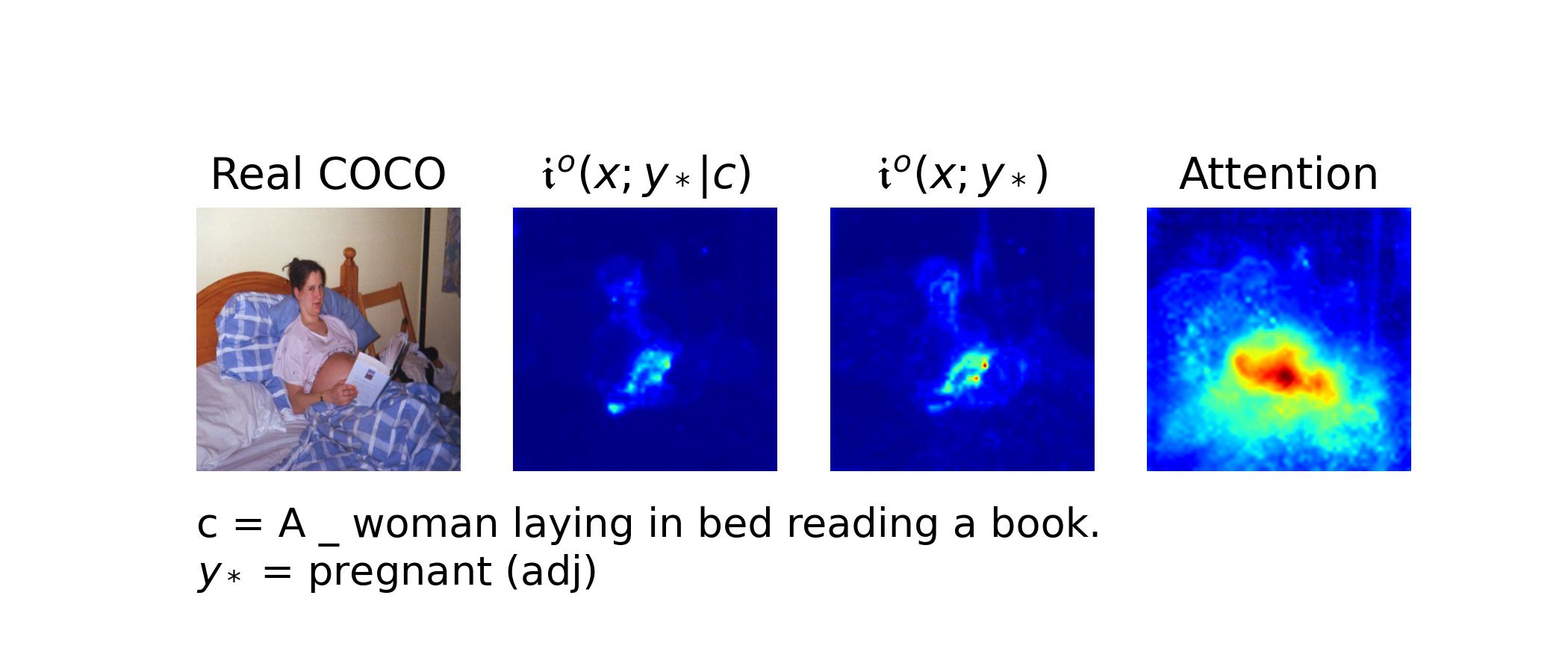} 
    \includegraphics[width=0.49\textwidth,trim={2cm 9mm 1.7cm 15mm},clip]{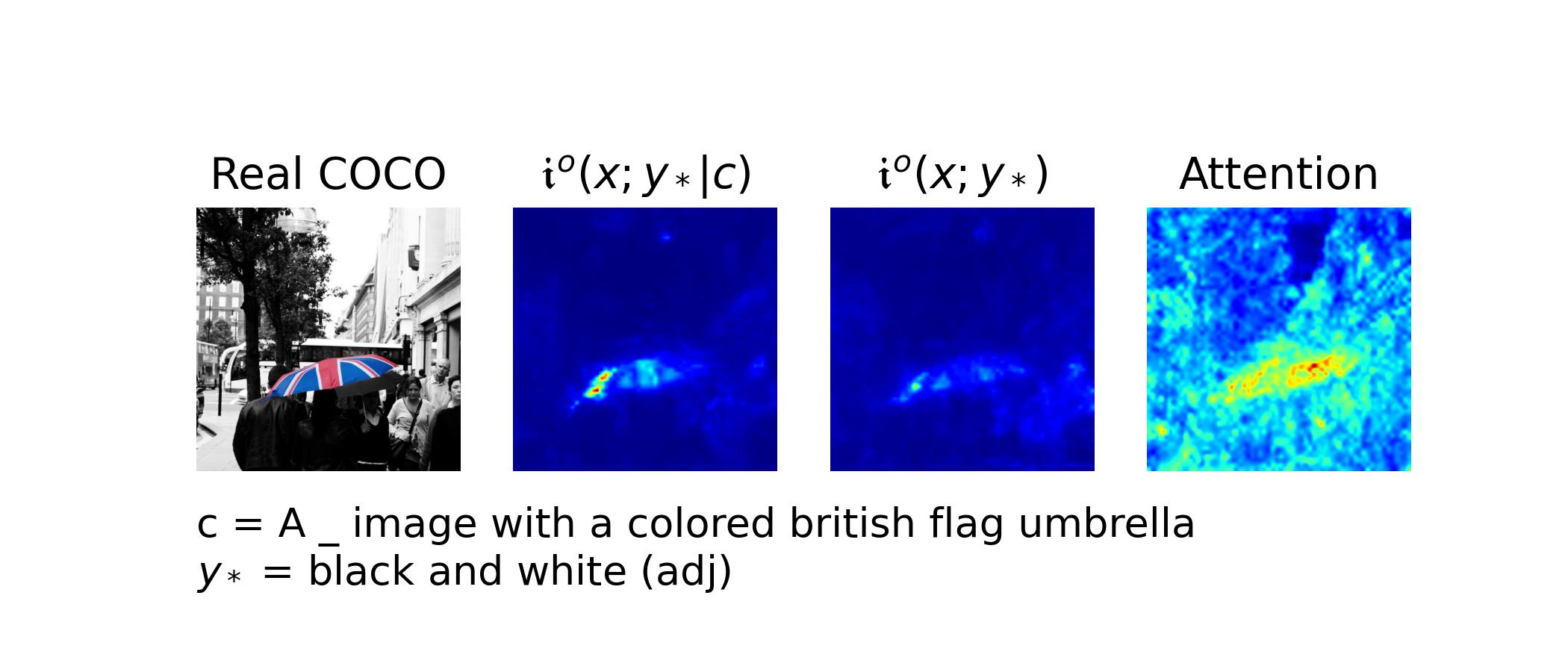} 
    \includegraphics[width=0.49\textwidth,trim={2cm 9mm 1.7cm 15mm},clip]{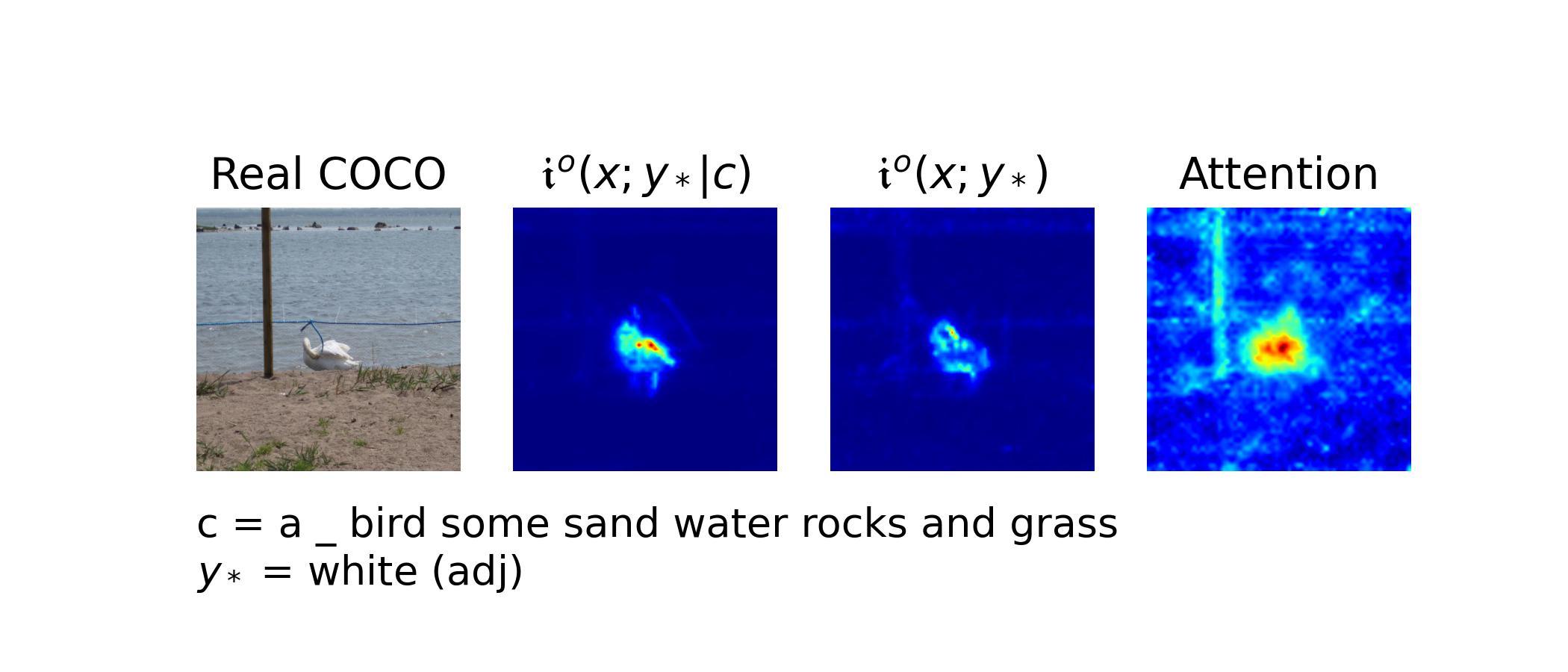} 
    \includegraphics[width=0.49\textwidth,trim={2cm 9mm 1.7cm 15mm},clip]{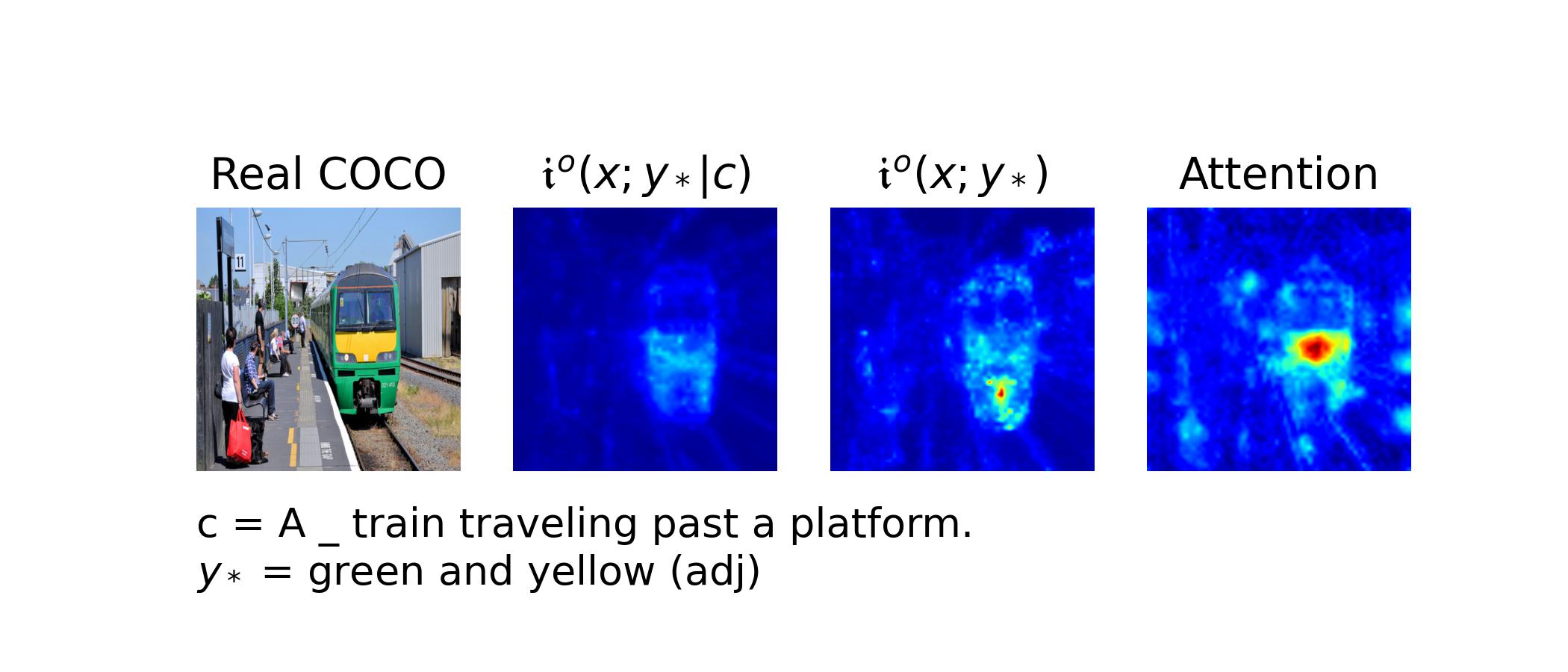} 
    \includegraphics[width=0.49\textwidth,trim={2cm 6mm 1.7cm 15mm},clip]{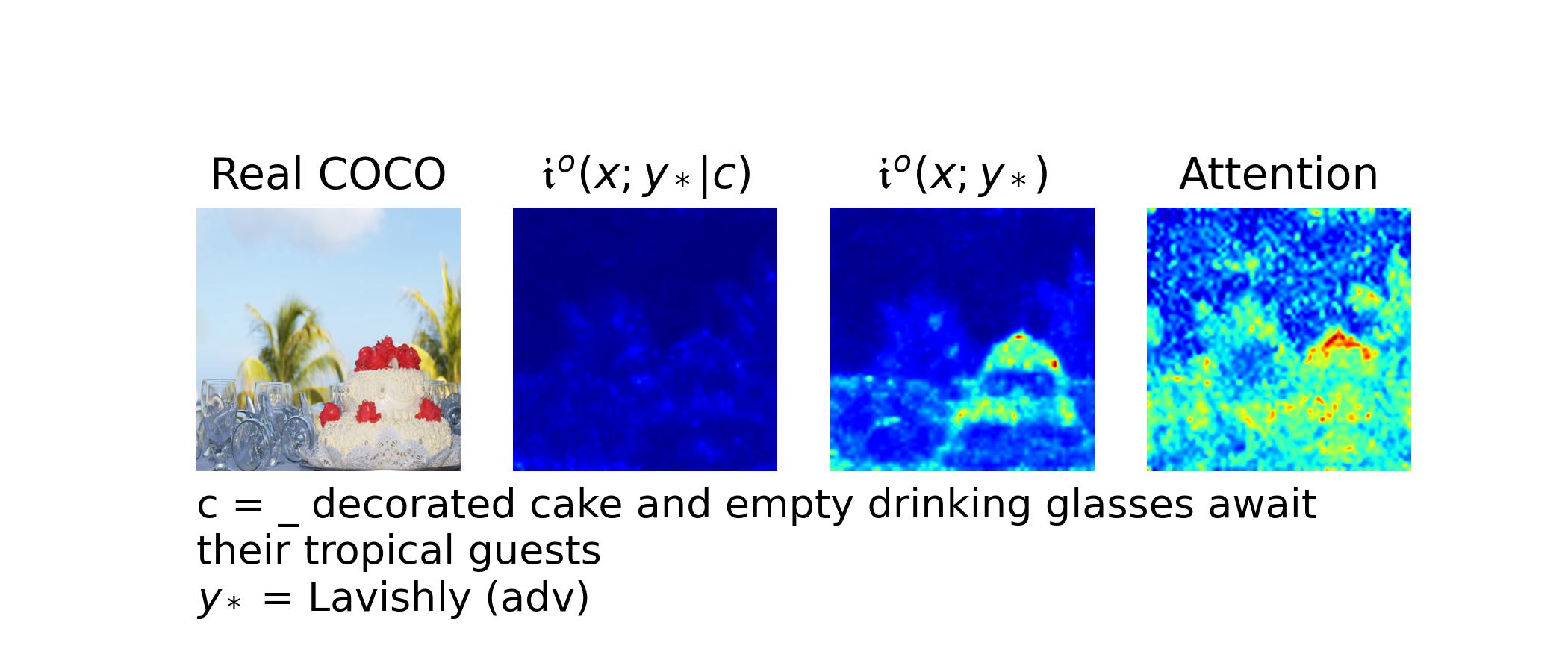} 
    \includegraphics[width=0.49\textwidth,trim={2cm 6mm 1.7cm 15mm},clip]{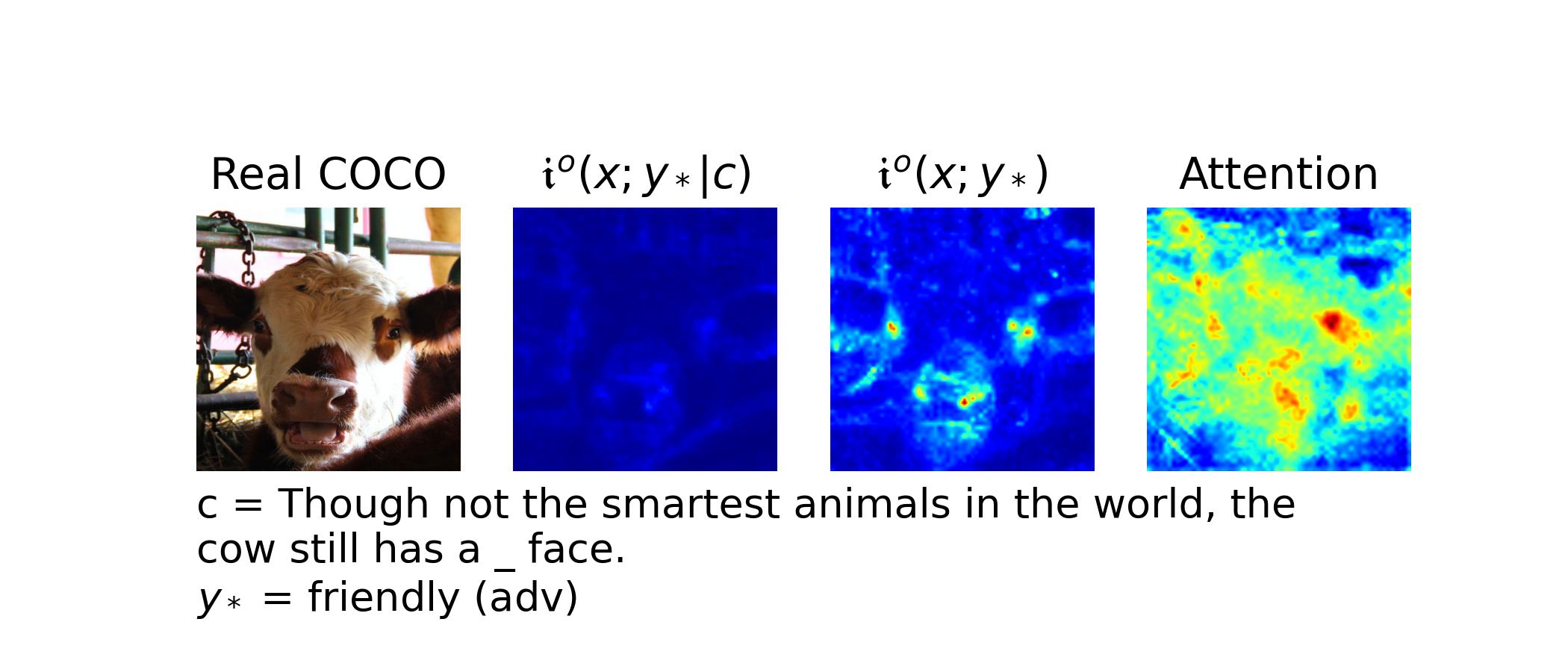} 
    \includegraphics[width=0.49\textwidth,trim={2cm 9mm 1.7cm 15mm},clip]{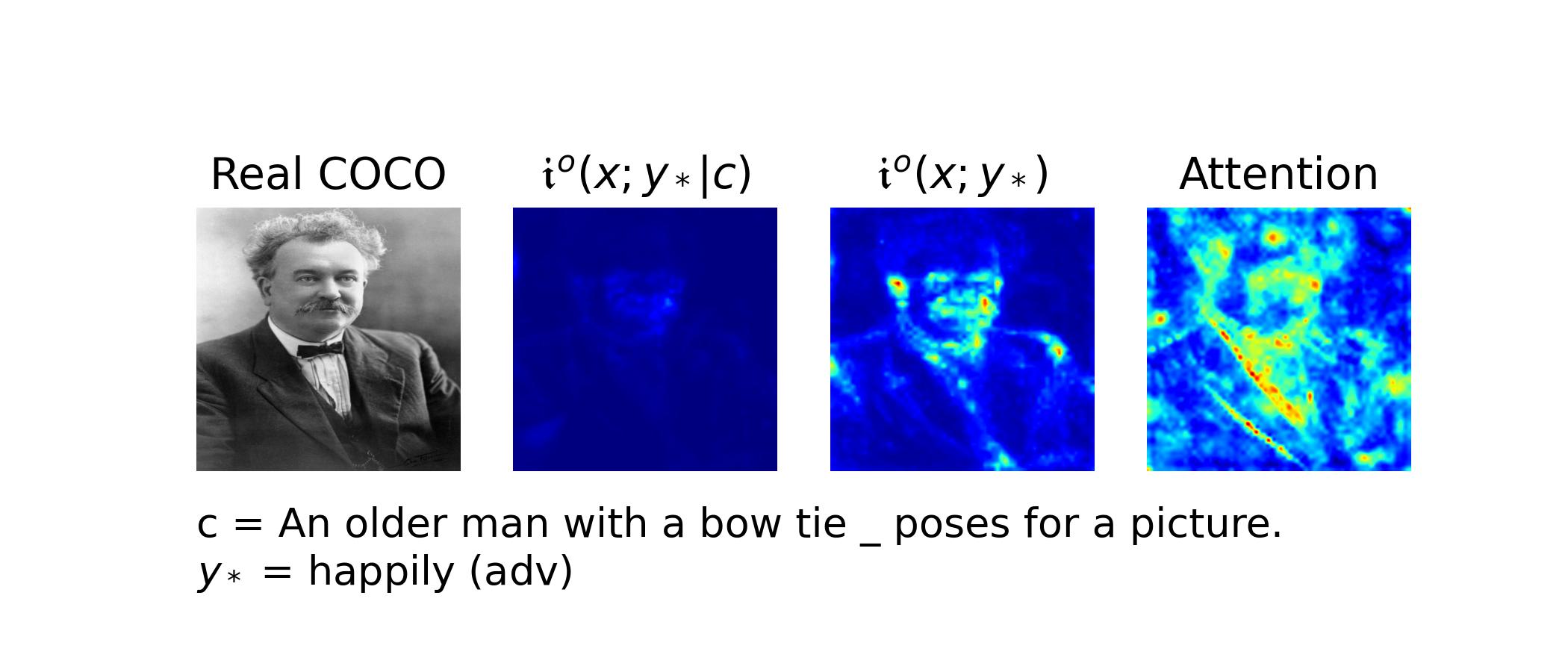} 
    \includegraphics[width=0.49\textwidth,trim={2cm 9mm 1.7cm 15mm},clip]{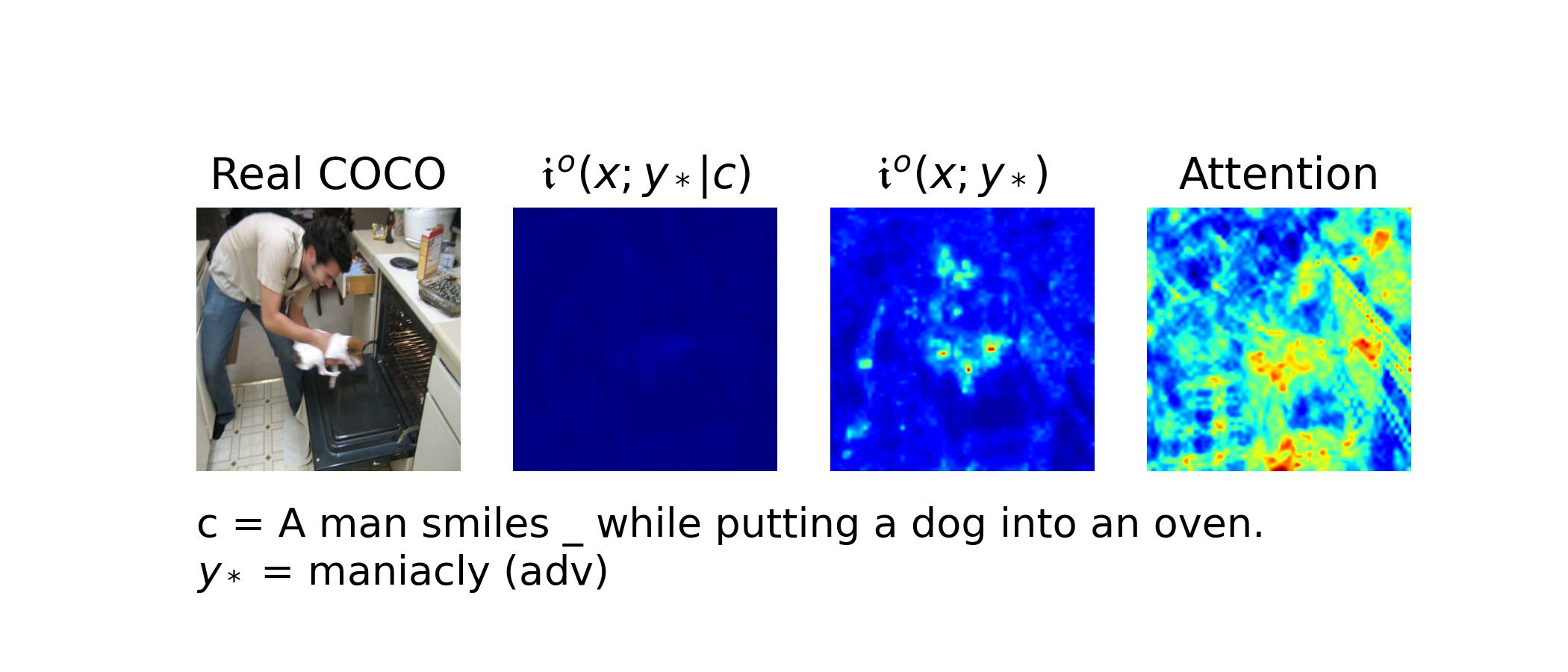} 
    \caption{Examples of localizing abstract words in images.}
    \label{fig:2d_mi_cmi_attn_7_1}
\end{figure}

We put more word location visualization examples for seven entities from COCO-WL, see Fig. \ref{fig:2d_mi_cmi_attn_7_1} \& \ref{fig:2d_mi_cmi_attn_7_2}.

\subsection{Intervention experiments additional results}\label{app:intervention}
\begin{figure}[h]
    \centering
    \includegraphics[width=0.99\textwidth,trim={5cm 8mm 4cm 1mm},clip]{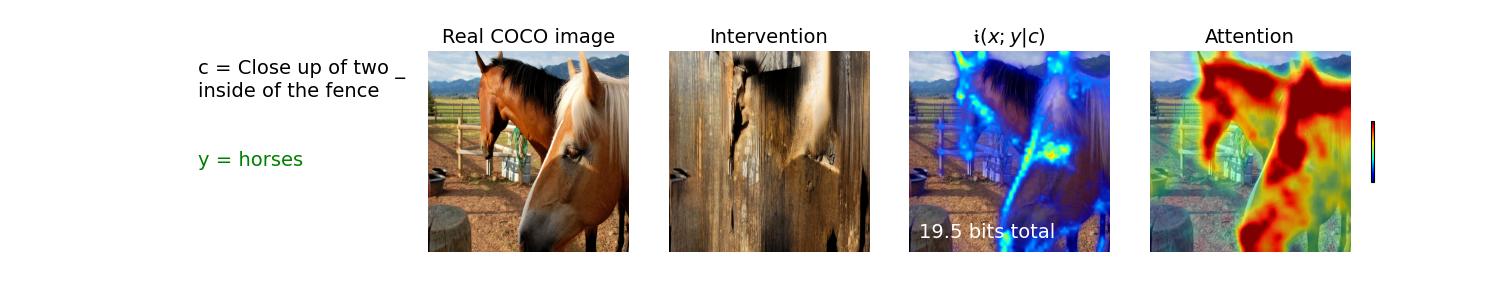} 
    (a) Low correlation between image change and heatmaps \\
    
    \includegraphics[width=0.99\textwidth,trim={5cm 8mm 4cm 13mm},clip]{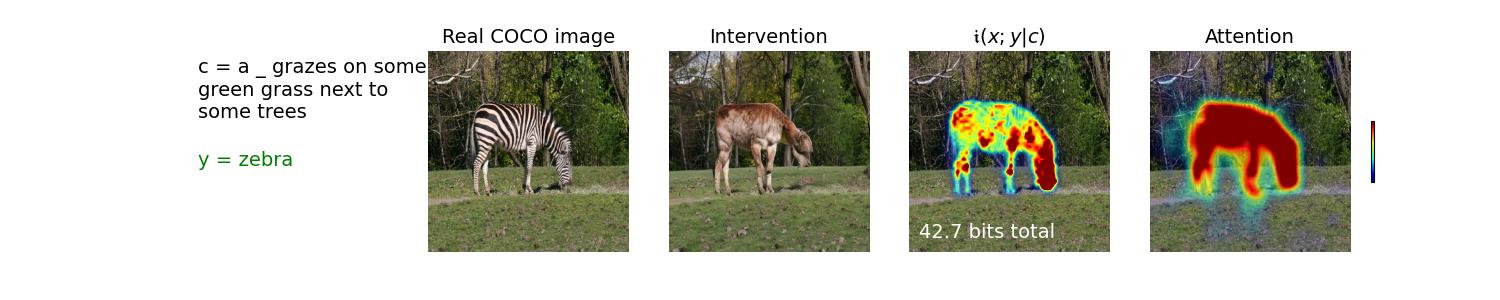} 
    (b) High correlation between image change and heatmaps
    \caption{(a) An example where the correlation between pixel-level changes and CMI or attention are low (0.25 and -0.21 respectively). (b) An example where the pixel-level correlation is high (0.73 and 0.77 respectively).}
    \label{fig:cor-examples}
\end{figure}
We observe that correlations between heatmaps (from conditional mutual information or from attention) often correlate strongly with changes in the image after intervention. However, this is not always the case. We show a negative and positive example in Fig.~\ref{fig:cor-examples}. We see in some images that one word change globally changes the image, leading to poor correlation. 
This result, however, does not contradict our original hypothesis, which is that small CMI implies that omitting a word will have no effect. We generally observe this to be true. However, when the CMI is large, the effect may or may not be correctly localized. The reason that the effect is not always correctly localized is that generation is an iterative procedure: a small change in the first step can lead to global changes in the image. 

\begin{figure}[p]
    \includegraphics[width=0.99\textwidth,trim={5cm 8mm 4cm 1mm},clip]{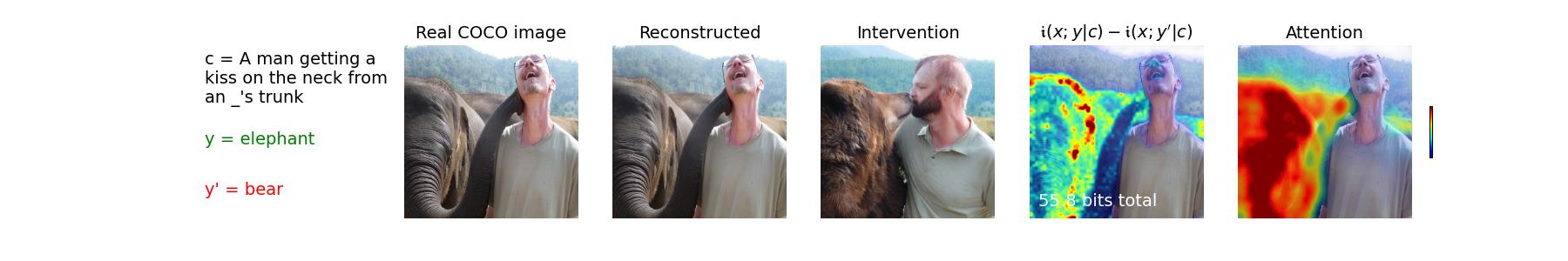} 
    \includegraphics[width=0.99\textwidth,trim={5cm 8mm 4cm 13mm},clip]{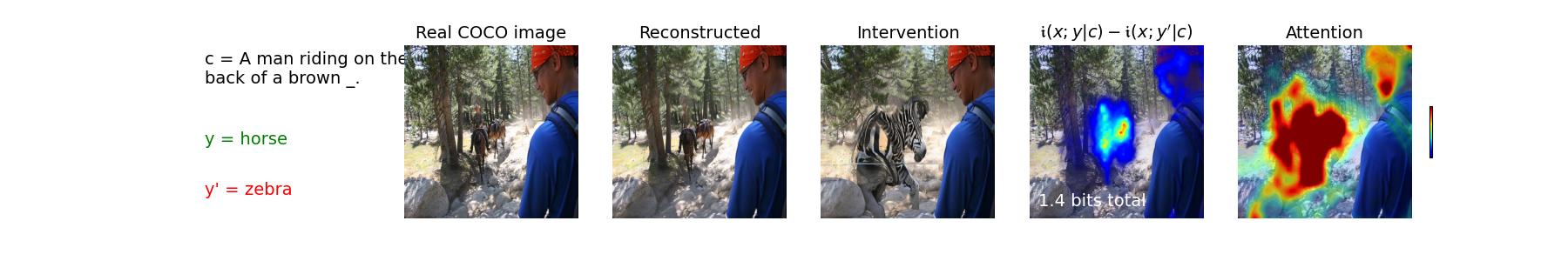} 
    \includegraphics[width=0.99\textwidth,trim={5cm 8mm 4cm 13mm},clip]{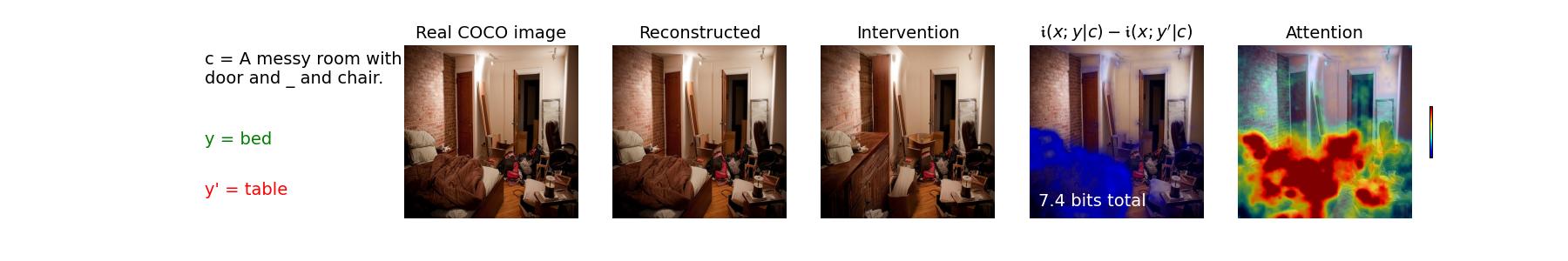} 
    \includegraphics[width=0.99\textwidth,trim={5cm 8mm 4cm 13mm},clip]{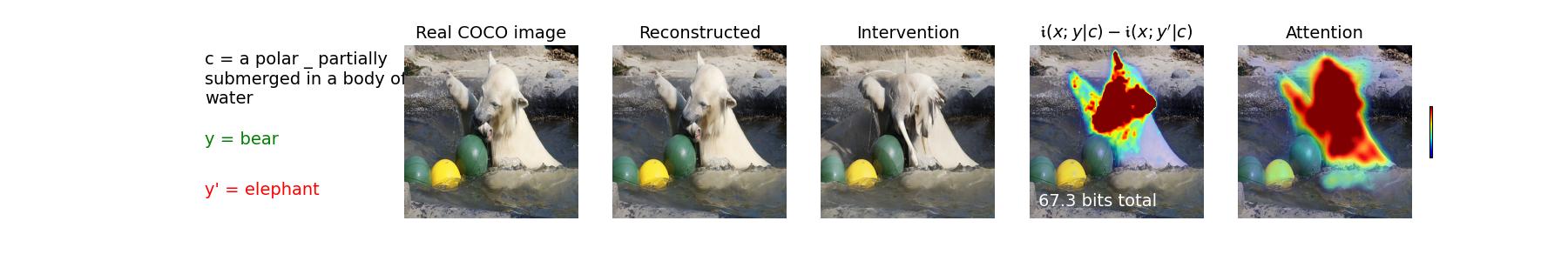} 
    \includegraphics[width=0.99\textwidth,trim={5cm 8mm 4cm 13mm},clip]{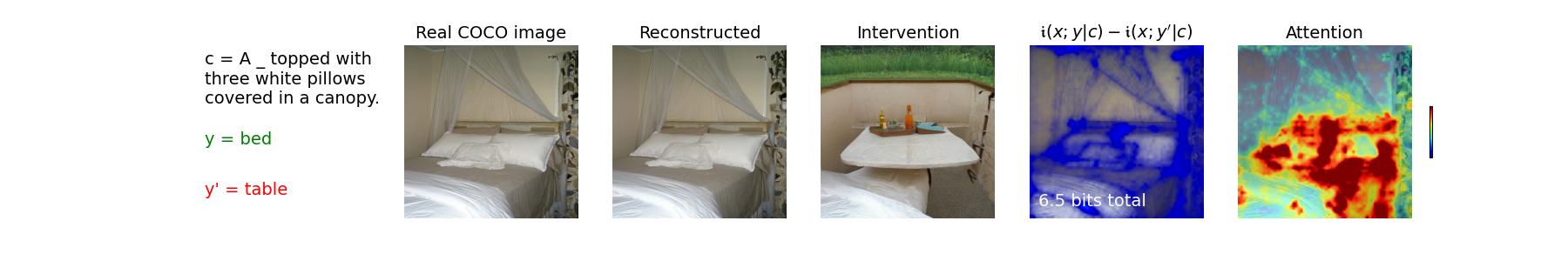} 
    \includegraphics[width=0.99\textwidth,trim={5cm 8mm 4cm 13mm},clip]{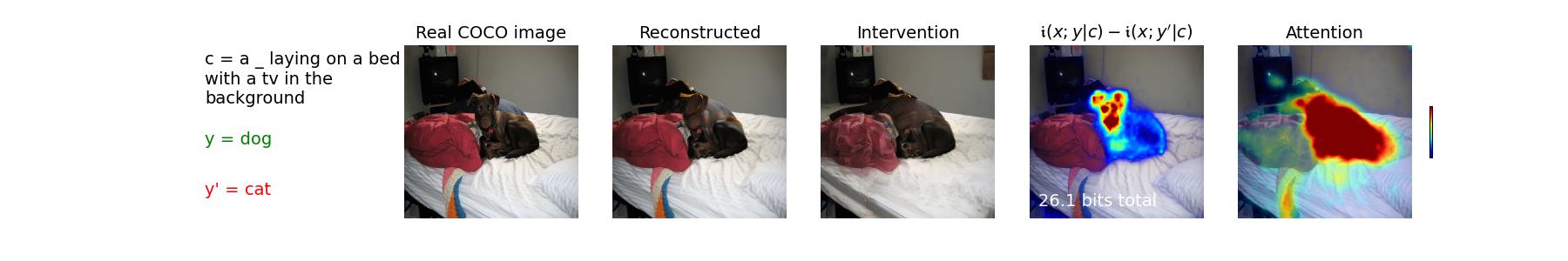} 
    \includegraphics[width=0.99\textwidth,trim={5cm 8mm 4cm 13mm},clip]{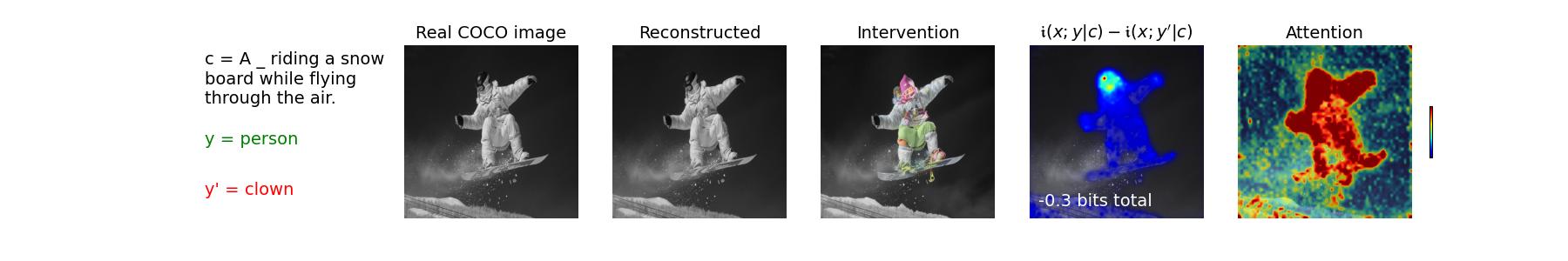} 
    \includegraphics[width=0.99\textwidth,trim={5cm 8mm 4cm 13mm},clip]{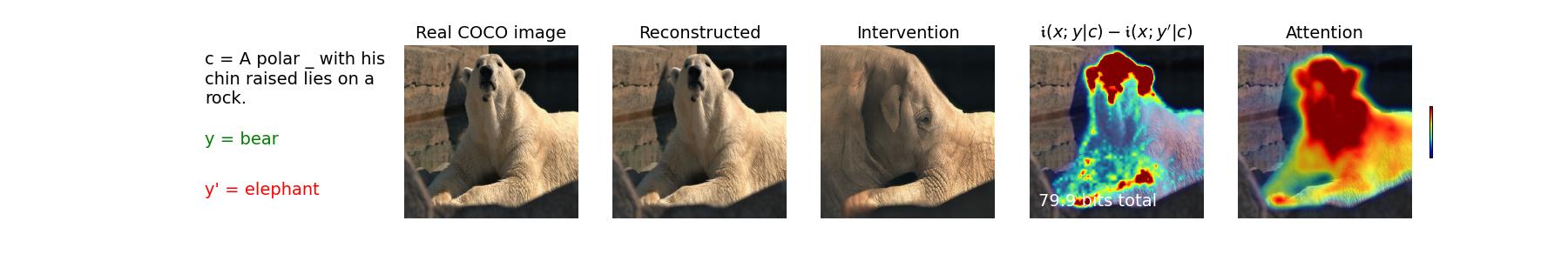} 
        \includegraphics[width=0.99\textwidth,trim={5cm 8mm 4cm 13mm},clip]{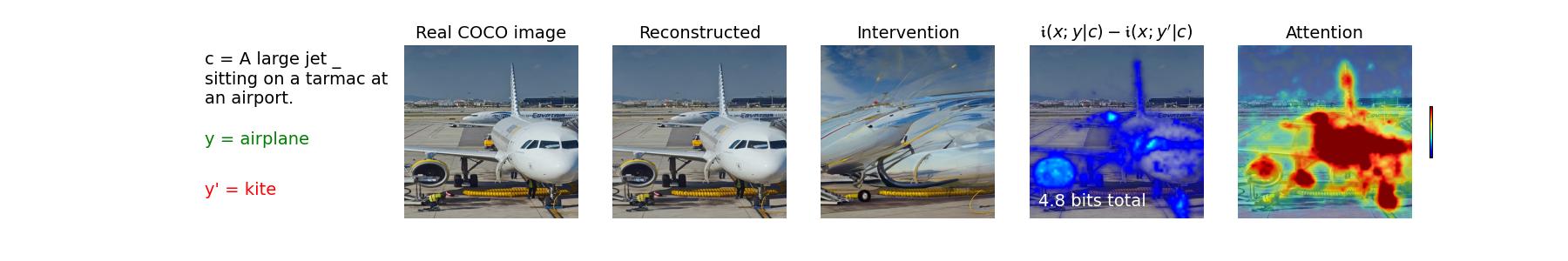} 
    \caption{Examples of word swap interventions}
    \label{fig:swap1}
\end{figure}

\begin{figure}[p]
    \centering
    \includegraphics[width=0.99\textwidth,trim={5cm 8mm 4cm 1mm},clip]{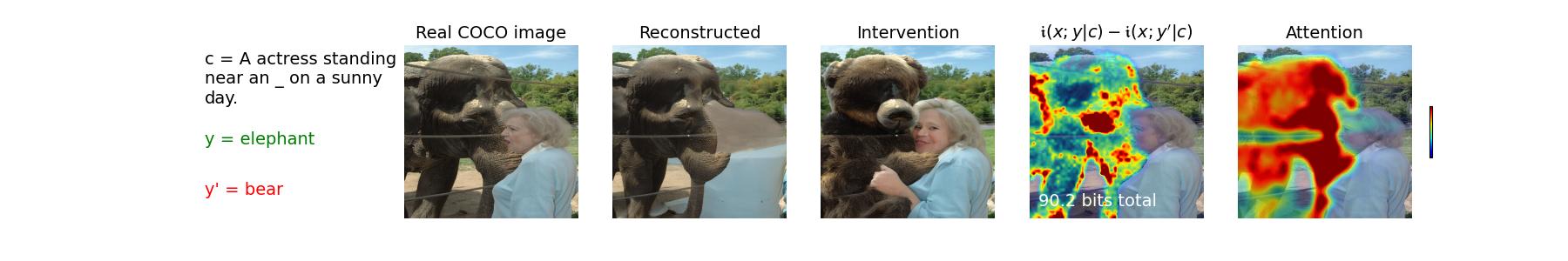} 
    \includegraphics[width=0.99\textwidth,trim={5cm 8mm 4cm 13mm},clip]{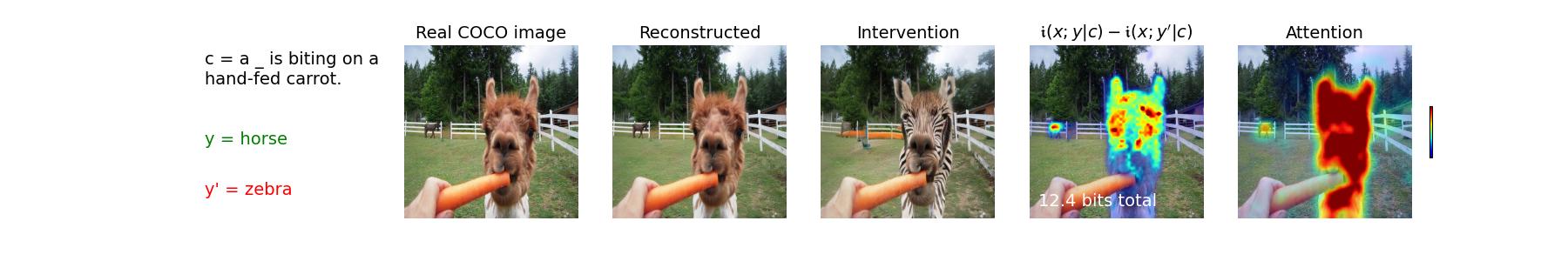} 
    \includegraphics[width=0.99\textwidth,trim={5cm 8mm 4cm 13mm},clip]{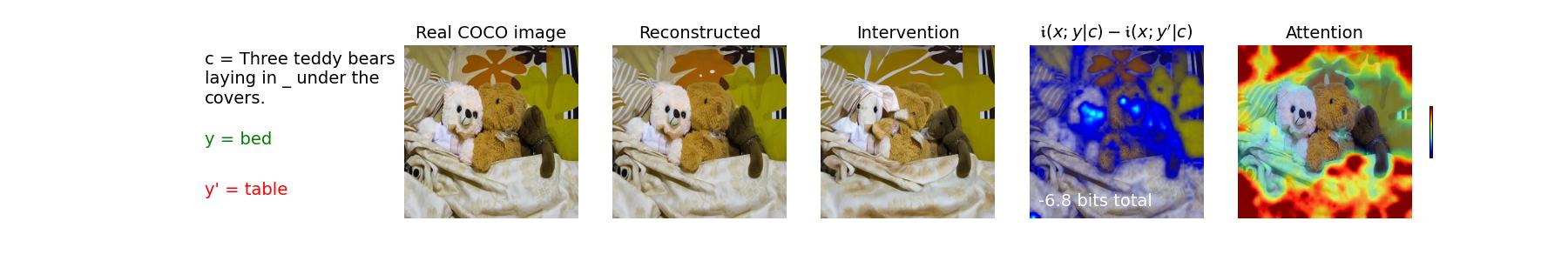} 
    \includegraphics[width=0.99\textwidth,trim={5cm 8mm 4cm 13mm},clip]{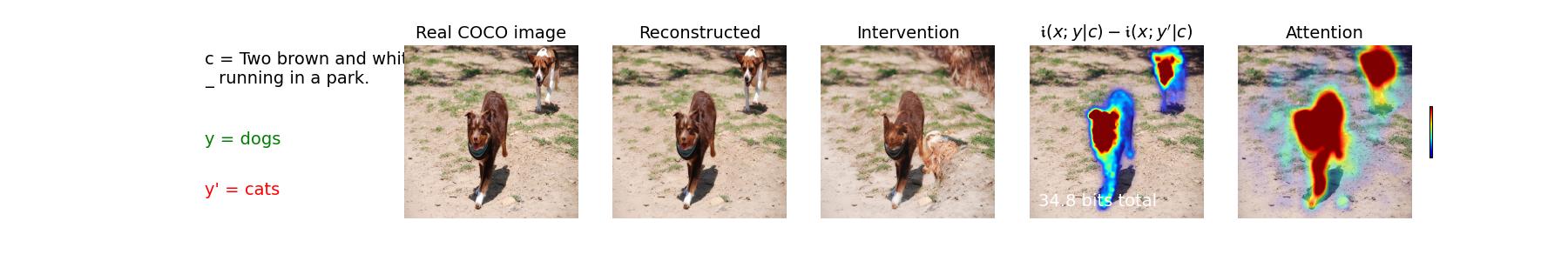} 
    \includegraphics[width=0.99\textwidth,trim={5cm 8mm 4cm 13mm},clip]{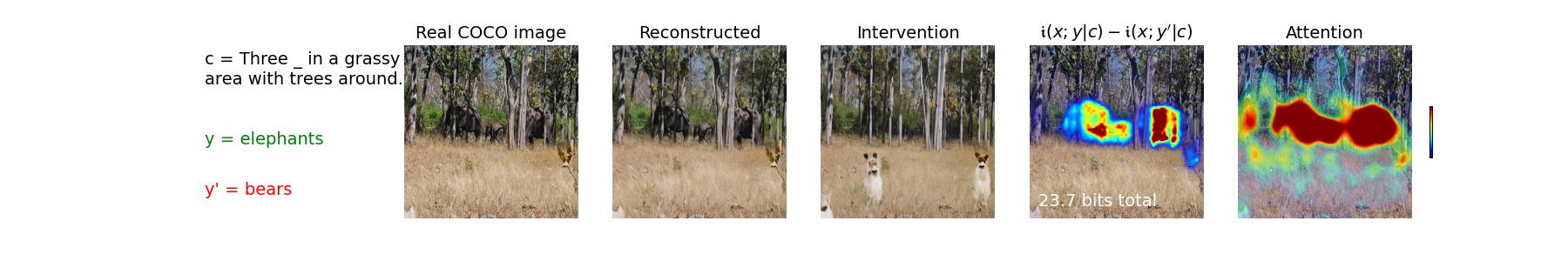} 
    \includegraphics[width=0.99\textwidth,trim={5cm 8mm 4cm 13mm},clip]{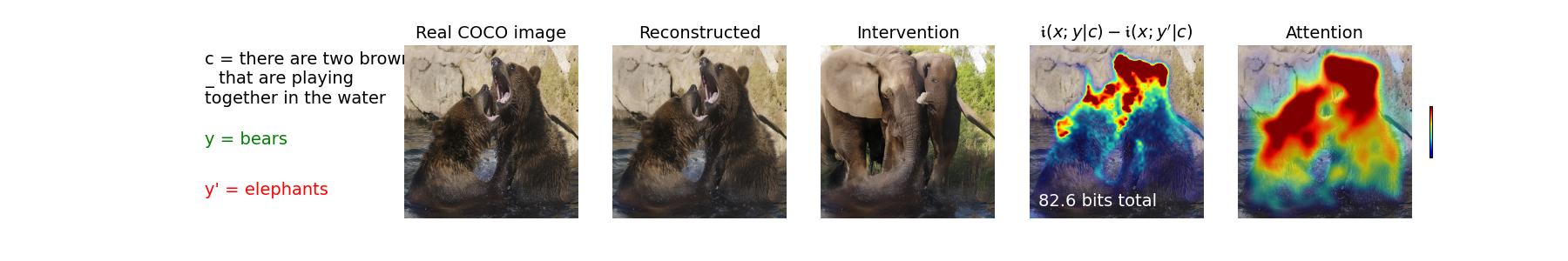} 
    \includegraphics[width=0.99\textwidth,trim={5cm 8mm 4cm 13mm},clip]{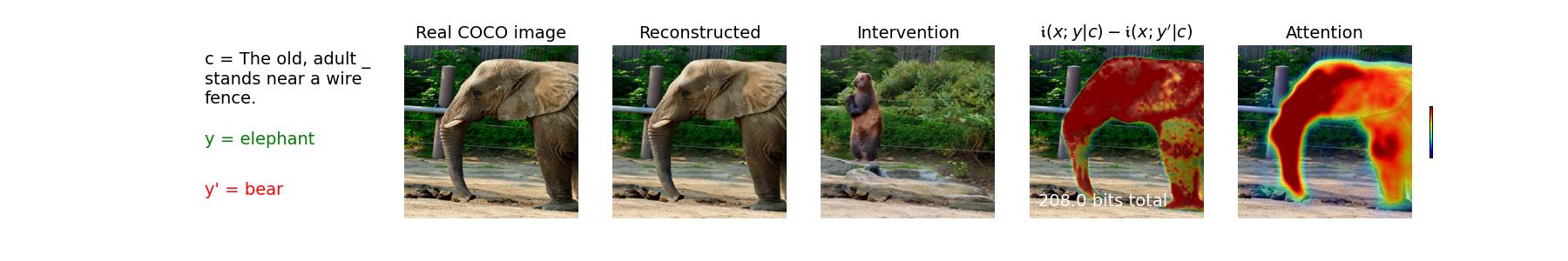} 
    \includegraphics[width=0.99\textwidth,trim={5cm 8mm 4cm 13mm},clip]{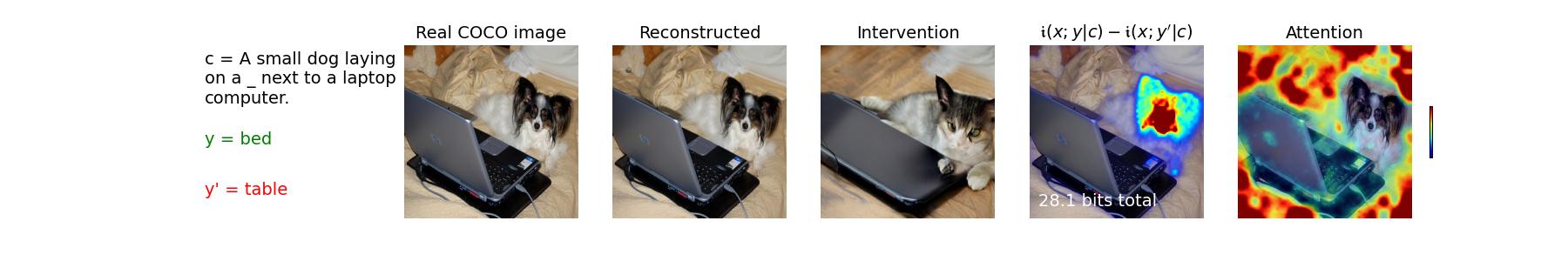} 
    \includegraphics[width=0.99\textwidth,trim={5cm 8mm 4cm 13mm},clip]{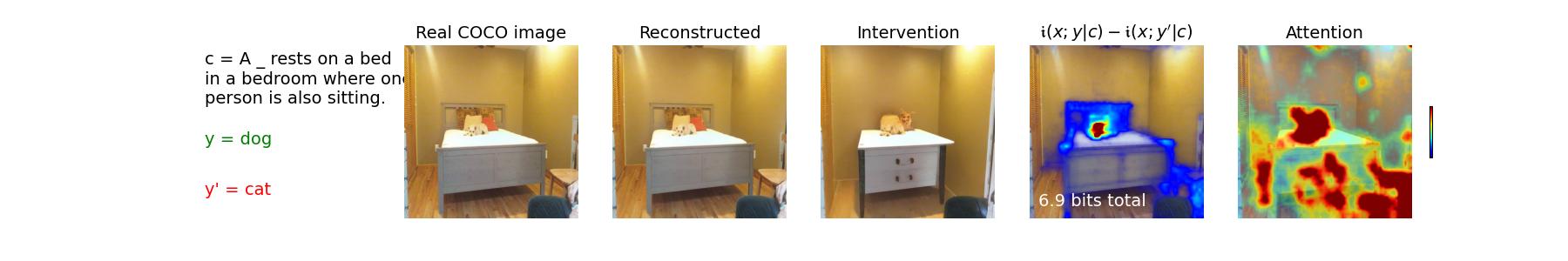} 
    \caption{Examples of word swap interventions}
    \label{fig:swap2}
\end{figure}

Fig.~\ref{fig:swap1} and ~\ref{fig:swap2} visualize additional examples where we swap a word in a caption with a categorically similar word. For the COCO-IT dataset described in \S\ref{app:data}, we explored the following word swaps: dog $\leftrightarrow$ cat, zebra $\leftrightarrow$ horse, bed $\leftrightarrow$ table, bear $\leftrightarrow$ elephant, airplane $\leftrightarrow$ kite, person $\leftrightarrow$ clown, plus plural versions. 
In these plots, and also Fig.~\ref{fig:abstract} and \ref{fig:cor-examples}, the pixel values represent $\ii^o$ and are shown with a colormap where the maximum value corresponds to $0.15$ bits/pixel. However, the ``total information'' shown in white text uses the unbiased estimate $\ii^s$, and hence can sometimes be negative.  
Attention color maps are normalized as was done by \citet{tang2022daam}.  

\section{Experimental Settings}\label{sec:settings}
We provide code for reproducing our experiments at \url{https://github.com/kxh001/Info-Decomp}. 
\subsection{Relation Testing with Pointwise Information}
\label{sec:aro-appendix}
We refer readers to Table \ref{tab:aro-hyperparameter} for additional implementation details for evaluating the ARO benchmark. All datasets are prepared following the official implementation of \citep{vlmbagofwords} available at \url{https://github.com/mertyg/vision-language-models-are-bows.git}. All experiments are run on NVIDIA RTX A6000 GPUs. 
\begin{table}[h]
\caption{Additional Experiment Details for the ARO Benchmark}
\label{tab:aro-hyperparameter}
\begin{center}
\begin{tabular}{lcccc}
\hline
\multicolumn{1}{c}{}  & \bf VG-A & \bf VG-R & \bf COCO & \bf Flickr30k
\\ \hline 
Perturbation size & 1 & 1 & 4 & 4\\
Dataset size & 28,748 & 23,937 & 25,010 & 5,000 \\
Inference batch size & 10 & 10 & 5 & 5\\
SNR sample size & 100 & 100 & 100 & 100\\
\hline
\end{tabular}
\end{center}
\end{table}

We evaluate the OpenCLIP checkpoint \texttt{laion/CLIP-ViT-H-14-laion2B-s32B-b79K}. This checkpoint consists of a 330M BERT-style encoder trained on the \href{https://huggingface.co/datasets/laion/laion2B-en}{LAION-2B Dataset}. Its text encoder is consistent with the one deployed by Stable Diffusion version 2.1 to ensure fair comparison. We use a batch size of 80 for all OpenCLIP evaluations.

Table \ref{tab:aro-extra} reports a more detailed set of experiments conducted on the ARO benchmark. First, we empirically study the effect of SNR importance sampling distributions on the consistency of our estimator, by reporting performances on uniform and logistic distributions. Sampling from the latter marginally but consistently out-performs its uniform counterpart, which supports the theoretical findings of \citep{kong22}. We report disagreements across relations in Table \ref{appendix:table-relation-results}, and observe disagreements to be consistently low between two distributions.

In the third-row of Table \ref{tab:aro-extra}, we report the performance of OpenCLIP wherein the last layer of its text encoder is removed. This setup is consistent with Stable Diffusion's usage of text encoder, but we observe the results to be similar for \textit{VG-Relation} and \textit{VG-Attribution} (1 perturbation), and significantly worse for \textit{COCO-Order} and \textit{Flickr30k-Order} (4 perturbations). All other entries are identical to Table \ref{tab:aro} for reference. 

In Table \ref{appendix:table-relation-results}, we report fine-grained performance of Stable Diffusion and OpenCLIP systems across each relation type. In column 3, we report normalized prediction disagreement between uniform and logistic sampling, and observe the predictions to be generally consistent. 

Finally, we assess the consistency of our estimator across random seeds. For each dataset, we select the first 1000 samples, evaluate our estimator across 3 random seeds, and provide OpenCLIP baseline on the same subsets as for reference. Numerical results are provided in Table \ref{tab:aro-stddev}. The information estimates are relatively consistent, and establish statistically significant performance gap compared to OpenCLIP. 

\begin{table}[h]
\caption{Additional Accuracy (\%) of Stable Diffusion and OpenCLIP.}
\label{tab:aro-extra}
\begin{center}
\small
\begin{tabular}{lcccc}
\hline
\multicolumn{1}{c}{\bf Method}  & \bf VG-A & \bf VG-R & \bf COCO & \bf Flickr30k
\\ \hline 
Baseline (Random Guess) & 50.0 & 50.0 & 20.0 & 20.0\\
OpenCLIP \citep{ilharco_gabriel_2021_5143773} & 64.6 & 51.4 & 32.8 & 40.5\\
OpenCLIP (all-but-last) & 65.6 & 50.9 & 22.4 & 28.8\\
Info. (Ours, Uniform) & 71.2 & 68.5 & 39.3 & 48.7\\
Info. (Ours, Logistic) & \textbf{72.0} & \textbf{69.1} & \textbf{40.1} & \textbf{49.3}\\
\hline
\end{tabular}
\end{center}
\end{table}

\begin{table}[h]
\caption{Mean Estimator Accuracy and Std. Dev. across Random Seeds}
\label{tab:aro-stddev}
\begin{center}
\begin{tabular}{lcccc}
\hline
 {}  & \bf VG-A & \bf VG-R & \bf COCO & \bf Flickr30k \\ \cline{2-5}
&\multicolumn{4}{c}{Accuracy$\pm$Standard Error (\%)} \\ \hline
Info. (Unif.) & 71.7 $\pm$ 0.59 & 72.3 $\pm$ 2.87 & 36.3 $\pm$ 0.87 & 50.5 $\pm$ 0.63\\
Info. (Log.) & \textbf{72.7} $\pm$ 1.32 &\textbf{73.4} $\pm$ 1.13 & \textbf{36.8} $\pm$ 0.50 & \textbf{51.2} $\pm$ 0.73\\
\hline
\end{tabular}
\end{center}
\end{table}

\subsection{Localizing Word Information in Images} \label{app:locword}
In our word localization experiments, we utilized a pre-trained Stable Diffusion v2-1-base model card available at \href{https://huggingface.co/stabilityai/stable-diffusion-2-1-base}{Huggingface}. Input images were resized to 512 $\times$ 512 and then normalized to the [0, 1] pixel value range to ensure compatibility with the pre-trained model. 

The DAAM \citep{tang2022daam} is essentially an extension integrated into Stable Diffusion models, designed to generate attention-based heatmaps concurrently with the image generation process. To leverage DAAM, it is imperative to pair it with a diffusion scheduler. In our experiments, we draw inspiration from \citep{concept_discovery} and employ a DDIM \citep{song2022denoising} scheduler as a baseline. While our ITD model \citep{kong22} is capable of independently generating MI and CMI heatmaps using the principles outlined in \S\ref{sec:methods}, we also had the option to enhance attention heatmaps by integrating DAAM with the information-theoretic diffusion process. Hence, we established three sets of comparative experiments: DAAM+DDIM (Attention), ITD (CMI), and DAAM+ITD (Attention+Info.) respectively. We opt not to use classifier-free guidance since it primarily aids in image generation and introduces additional undesired content onto images in the denoising process. We utilize the ``alphas\_cumprod'' in the scheduler of the Stable Diffusion model to compute the $\logsnr$ range spans from -5 to 7. For specific $t$-to-$\alpha$ transformation calculations, you could refer to Appendix B.2 in \citep{kong22}. Thus, the parameters of the corresponding logistic distribution are [loc, scale, clip] = [1, 2, 3]. Unfortunately, DAAM only supports a batch size of 1. For DAAM+DDIM and DAAM+ITD, we utilize the following input: ($\vx$, $\vy$, $y*$). In the case of ITD, the input configuration is ($\vx$, $\vy$, $\vc$). This distinction arises from the fact that DAAM generates heatmaps based on the cross-attention map, enabling the direct calculation of the score for an individual object word on each pixel. On the other hand, ITD relies on conditional mutual information, $\ii^{o}_j(\vx;y_*|\vc)$.
\begin{table}[h]
\centering
\caption{The hyper-parameters used in DDIM and ITD schedulers.}
\label{tab:hparam}
\begin{tabular}{cccc}
\hline
 \textbf{random seed} & \textbf{batch size (DDIM/ITD)} & \textbf{logistic distribution (ITD)}            &         \textbf{guidance (DDIM)}          \\ \hline
42          & 1/10                                     & {[}loc, scale, clip{]} = {[}1, 2, 3{]} &        1          \\ \hline
\end{tabular} 
\end{table}

All experiments were conducted using Nvidia RTX 6000 GPU cards. The hyper-parameters used in these experiments are summarized in Table \ref{tab:hparam}, with variations in the number of diffusion steps set at 50, 100, and 200. Once the heatmap of the image is computed, we initially rescale it to the [0, 1] range and subsequently apply a uniform hard threshold on them for segmentation. After experimenting with hard thresholds vary in [0, 1], we identify the optimal threshold that yields the highest mIoU value, then record the mIoU in Table \ref{tab:segment}. Unless explicitly stated, all visualization for MI, CMI, and attention heatmaps are based on 100 diffusion steps.

\begin{table}[h]
\caption{Unsupervised Object Segmentation mIoU (\%) Standard Error Analysis on COCO-IT}
\label{tab:segment_std}
\begin{center}
\begin{tabular}{ccccc}
\hline
\bf Method & \bf 1 step & \bf 50 steps & \bf 100 steps & \bf 200 steps
\\ \hline 
Whole Image Mask & 14.94 $\pm$ 0.0022 & 14.94 $\pm$ 0.0022 & 14.94 $\pm$ 0.0022 & 14.94 $\pm$ 0.0022\\
Attention & 37.89 $\pm$ 0.0030 & 34.52 $\pm$ 0.0030 & 34.90 $\pm$ 0.0030 & 35.35 $\pm$ 0.0030\\
CMI      & 21.73 $\pm$ 0.0023 & 32.31 $\pm$ 0.0026 & 33.24 $\pm$ 0.0026 & 33.63 $\pm$ 0.0026\\
Attention+Info.         & \textbf{37.96} $\pm$ 0.0030 & \textbf{42.46} $\pm$ 0.0032 & \textbf{42.71} $\pm$ 0.0032 & \textbf{42.84} $\pm$ 0.0032\\
\hline
\end{tabular}
\end{center}
\end{table}

\begin{figure}[h]
	\centering
	\includegraphics[width=0.65\textwidth]{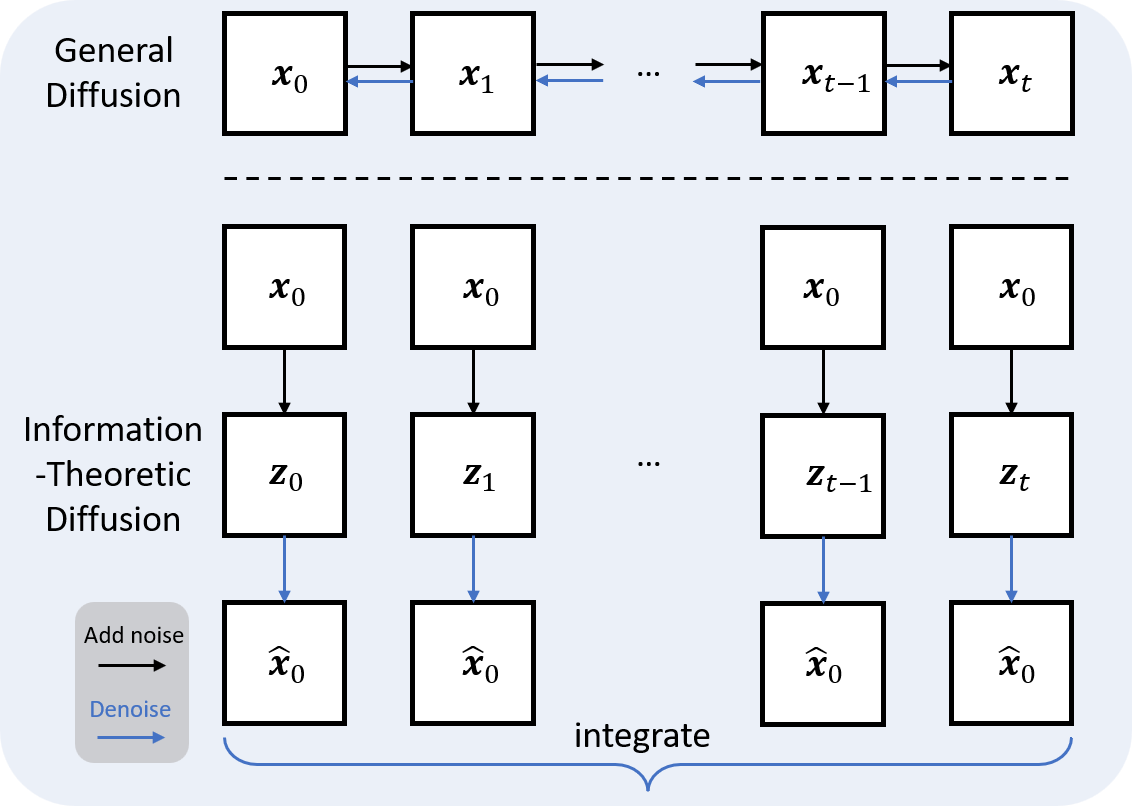}
	\caption{The diagram of two different diffusion processes.}
	\label{fig:parallel}
\end{figure}

We calculated the standard error of the IoU values for object segmentation experiments conducted on COCO-IT, and the results are documented in Table \ref{tab:segment_std}. This indicates that the number of diffusion steps does not significantly affect the variation in IoU values. Notice that Table \ref{tab:segment_std} includes an additional column for the 1-step experiment results. The 1-step DAAM-DDIM diffusion process can be regarded as denoising images with imperceptible noise, which is surprisingly effective compared to the multi-step results.
However, computing MI and CMI only at a single step, or $\logsnr$, is not directly comparable. The steps in that case are interpreted as elements in a sum approximating an integral, and we don't expect a one step sum to be a good estimate. Additionally, as per the analysis in \S\ref{app:MMSE}, peaks are required for an accurate match between relevant pixels and object words, which cannot be predicted in advance. Nonetheless, the results still demonstrate that the information-theoretic diffusion process enhances attention with respect to object segmentation. {Additionally, it's noteworthy to mention that the generation process for MI or CMI from ITD differs from the generation of DAAM. DAAM requires continuous noise addition and denoising iterations to compute, while ITD first samples a series of $\logsnr$, and then each $\logsnr$ can undergo independent noise addition and denoising computations. Finally, MI or CMI is calculated by one integration, which facilitates parallel computing, see Fig. \ref{fig:parallel}.}

\section{COCO-IT Dataset Preparation} \label{app:data}
While the MSCOCO \citep{lin2015microsoft} dataset boasts ample image-text pairs, not every object present in the images is mentioned in the captions, even if these objects have been labeled and annotated. In our experiments, we would like to test (1) the mutual information between a complete prompt and the corresponding image, and (2) the conditional mutual information between the object word and the image. Therefore, we filtered the original COCO 2017 validation dataset using the following steps:
\begin{enumerate}[label=(\alph*)]
    \item Traverse all the objects in an image.
    \item Match each object word to the caption containing that object.
    \item Generate one data point with four contains: [image, caption, context, object].
    \item If one object doesn't appear in the captions, then omit that data point.
\end{enumerate}
After applying this filter, we acquired a dataset, COCO-IT, comprising 6,927 validation image-text data points and 79 categories. To facilitate more effective visualization, we further randomly selected 10 categories from it, choosing 10 image-text pairs for each to create a smaller dataset, COCO100-IT. Additionally, we constructed a dataset, COCO-WL, for word localization by selecting 10 cases for seven different entities (verb, num., adj., adv., prep., pron., conj.).

\begin{table}[t]
\centering
\caption{Fine-grained results in Visual Genome Relation dataset.}
\label{appendix:table-relation-results}
\small
\begin{tabular}{lrrrrrrr}
\toprule
{} &      Info. (Unif.) ($\uparrow$) &      Info. (Log.) ($\uparrow$) &   Disagreement ($\downarrow$) &   OpenCLIP ($\uparrow$) &  \# Samples \\
\midrule
\textbf{Accuracy (\%)} & 68.5 & 69.1 & 6.7 & 51.4 & {} \\ \midrule
Spatial Relationships \\ \midrule
above           &  49.8 &  53.2 & 5.6 & 55.0 &  269 \\
at              &  70.7 &  72.0 &  9.3 & 66.7 &  75 \\
behind          &  39.9 &  39.9 &  4.5 & 54.4 &  574 \\
below           &  49.3 &  46.4 &  7.7 & 49.8 &  209 \\
{\color{magenta} beneath}         &  {\color{magenta} 50.0} &  {\color{magenta} 50.0} &  0.0 & {\color{magenta} 90.0} &  10 \\
in              &  76.7 &  79.9 &  5.5 & 51.6 &  708\\
in front of     &  70.2 &  68.7 &  7.3 & 63.1 &  588 \\
inside          &  69.0 &  74.1 &  8.6 & 56.9 &  58 \\
on              &  75.1 &  75.6 &  6.4 & 51.0 &  1684 \\
on top of       &  62.7 &  63.7 &  9.0 & 46.3 &  201 \\
to the left of  &  51.1 &  51.2 &  7.8 & 50.5 &  7741\\
to the right of &  48.6 &  49.3 &  7.9 & 49.8 &  7741\\
under           &  47.0 &  46.2 &  3.8 & 43.9 &  132 \\ \midrule
Verbs \\ \midrule
carrying          &  58.3 &  66.7 &  8.3 & 33.3 &  12 \\
covered by        &  36.1 &  33.3 &  8.3 & 55.6 &  36 \\
{\color{magenta} covered in}        &  {\color{magenta} 14.3} &  {\color{magenta} 14.3} &  14.3 & {\color{magenta} 50.0} &  14 \\
{\color{magenta} covered with}      &  {\color{magenta} 18.8} &  {\color{magenta} 18.8} &  0.0 & {\color{magenta} 43.8} &  16 \\
covering          &  63.6 &  72.7 &  15.2 & 54.5 &  33 \\
cutting           &  91.7 &  91.7 &  0.0 & 66.7 &  12 \\
eating            &  85.7 &  85.7 &  0.0 & 57.1 &  21\\
{\color{magenta} feeding}           &  {\color{magenta} 40.0} &  {\color{magenta} 50.0} &  10.0 & {\color{magenta} 100.0} &  10\\
{\color{darkpastelgreen} grazing on}        &  {\color{darkpastelgreen} 60.0} &  {\color{darkpastelgreen} 60.0} &  0.0 & {\color{darkpastelgreen} 30.0} &  10\\
hanging on        &  57.1 &  71.4 &  14.3 & 78.6 &  14\\
{\color{darkpastelgreen} holding}           &  {\color{darkpastelgreen} 90.1} &  {\color{darkpastelgreen} 87.3} &  5.6 & {\color{darkpastelgreen} 52.1} &  142\\
leaning on        &  66.7 &  66.7 &  0.0 & 66.7 &  12\\
{\color{darkpastelgreen} looking at}        &  {\color{darkpastelgreen} 80.6} &  {\color{darkpastelgreen} 83.9} &  3.2 & {\color{darkpastelgreen} 48.4} &  31\\
{\color{darkpastelgreen} lying in}          &  {\color{darkpastelgreen} 100.0} &  {\color{darkpastelgreen} 100.0} & 0.0 & {\color{darkpastelgreen} 33.3} &  15\\
{\color{darkpastelgreen} lying on}          &  {\color{darkpastelgreen} 81.7} &  {\color{darkpastelgreen} 86.7} &  5.0 & {\color{darkpastelgreen} 40.0} &  60\\
parked on         &  76.2 &  71.4 &  4.8 & 61.9 &  21\\
reflected in      &  71.4 &  64.3 & 7.1 & 61.9 &  14\\
{\color{darkpastelgreen} resting on}        &  {\color{darkpastelgreen} 69.2} &  {\color{darkpastelgreen} 84.6} &  15.4 & {\color{darkpastelgreen} 15.4} &  13\\
{\color{darkpastelgreen} riding}            &  {\color{darkpastelgreen} 80.4} &  {\color{darkpastelgreen} 76.5} &  7.8 & {\color{darkpastelgreen} 37.3} &  51\\
sitting at        &  65.4 &  69.2 & 3.8 & 38.5 &  26\\
sitting in        &  82.6 &  82.6 & 8.7 & 65.2 &  23\\
{\color{darkpastelgreen} sitting on}        &  {\color{darkpastelgreen} 80.6} &  {\color{darkpastelgreen} 78.9} &  8.6 & {\color{darkpastelgreen} 49.7} &  175\\
sitting on top of &  80.0 &  50.0 &  30.0 & 60.0 &  10\\
{\color{darkpastelgreen} standing by}       &  {\color{darkpastelgreen} 91.7} &  {\color{darkpastelgreen} 91.7} &  16.7 & {\color{darkpastelgreen} 50.0} &  12\\
{\color{darkpastelgreen} standing in}       &  {\color{darkpastelgreen} 89.8} &  {\color{darkpastelgreen} 91.5} &  8.5 & {\color{darkpastelgreen} 49.2} &  59\\
standing on       &  78.8 &  82.7 &  3.8 & 55.8 &  52\\
surrounded by     &  64.3 &  57.1 & 7.1 & 42.9 &  14\\
{\color{darkpastelgreen} using}             &  {\color{darkpastelgreen} 100.0} &  {\color{darkpastelgreen} 100.0} & 0.0 & {\color{darkpastelgreen} 21.1} &  19\\
walking in        &  90.0 &  90.0 & 0.0 & 70.0 &  10\\
{\color{darkpastelgreen} walking on}        &  {\color{darkpastelgreen} 94.7} &  {\color{darkpastelgreen} 94.7} & 0.0 & {\color{darkpastelgreen} 36.8} &  19\\
{\color{darkpastelgreen} watching}          &  {\color{darkpastelgreen} 72.7} &  {\color{darkpastelgreen} 77.3} & 4.5 & {\color{darkpastelgreen} 31.8} &  22\\
{\color{darkpastelgreen} wearing}           &  {\color{darkpastelgreen} 82.7} &  {\color{darkpastelgreen} 84.1} & 6.4 & {\color{darkpastelgreen} 44.9} &  949\\

\bottomrule
\end{tabular}
\end{table}

\begin{figure}[p]
    \centering
    \includegraphics[width=0.99\textwidth,trim={5cm 8mm 4cm 1mm},clip]{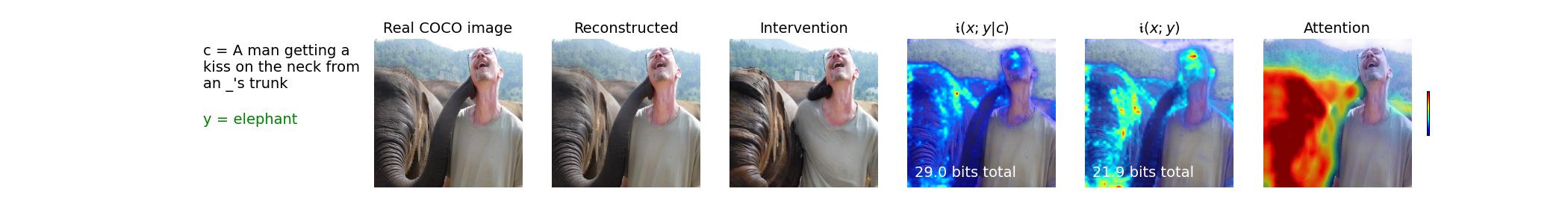} 
    \includegraphics[width=0.99\textwidth,trim={5cm 8mm 4cm 13mm},clip]{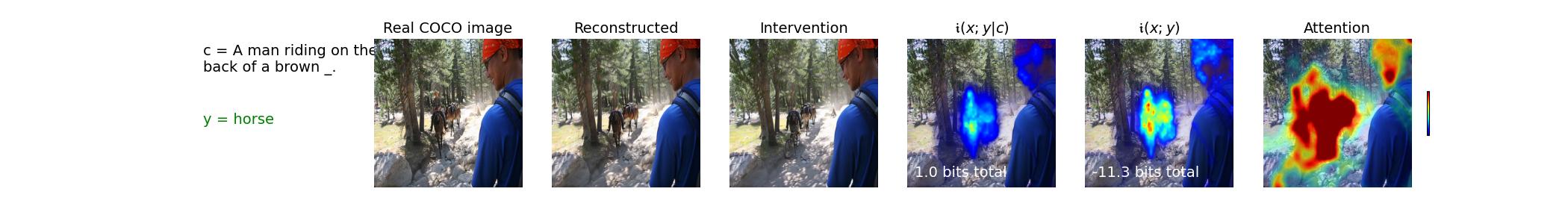} 
    \includegraphics[width=0.99\textwidth,trim={5cm 8mm 4cm 13mm},clip]{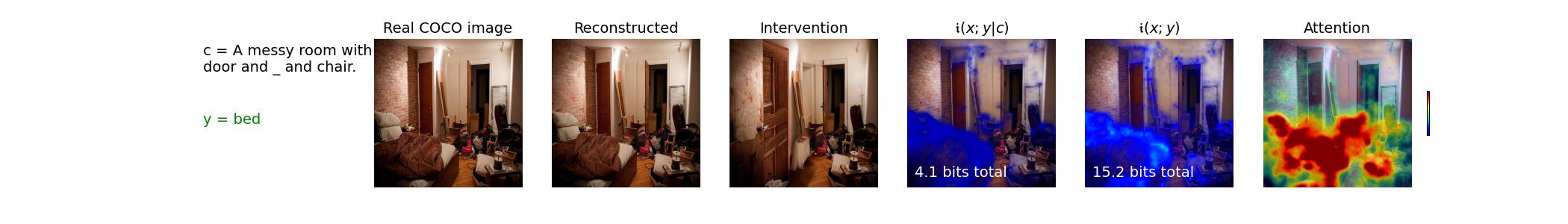} 
    \includegraphics[width=0.99\textwidth,trim={5cm 8mm 4cm 13mm},clip]{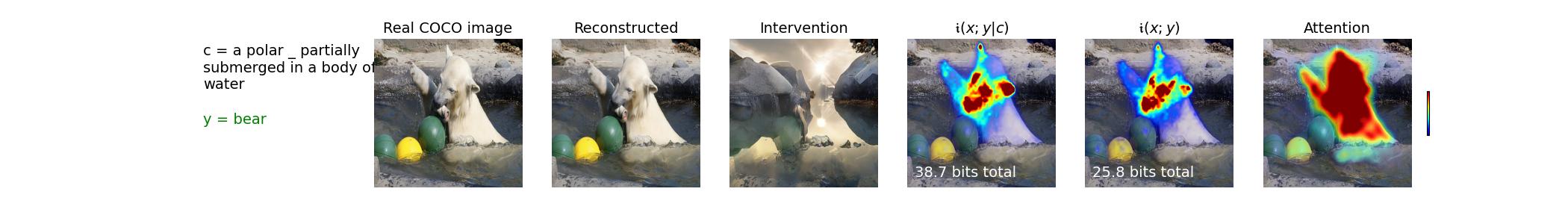} 
    \includegraphics[width=0.99\textwidth,trim={5cm 8mm 4cm 13mm},clip]{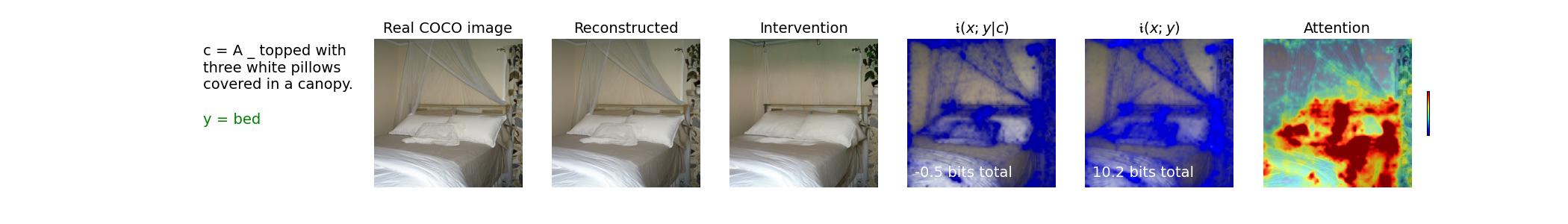} 
    \includegraphics[width=0.99\textwidth,trim={5cm 8mm 4cm 13mm},clip]{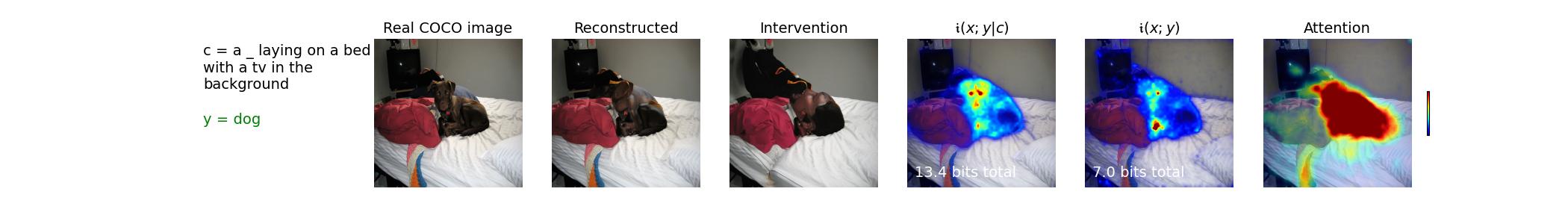} 
    \includegraphics[width=0.99\textwidth,trim={5cm 8mm 4cm 13mm},clip]{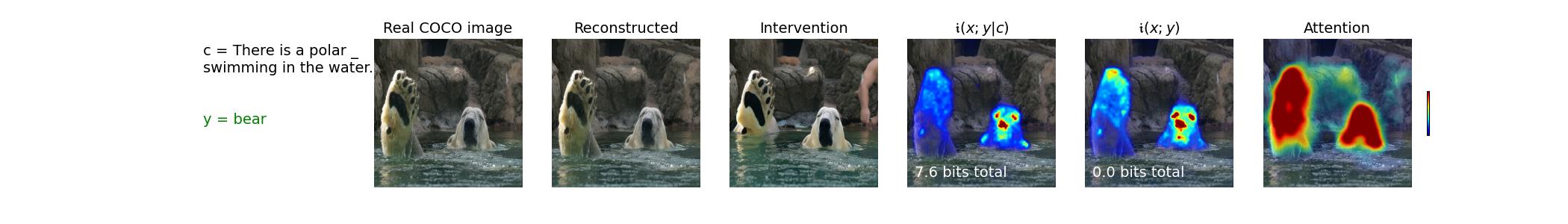} 
    \includegraphics[width=0.99\textwidth,trim={5cm 8mm 4cm 13mm},clip]{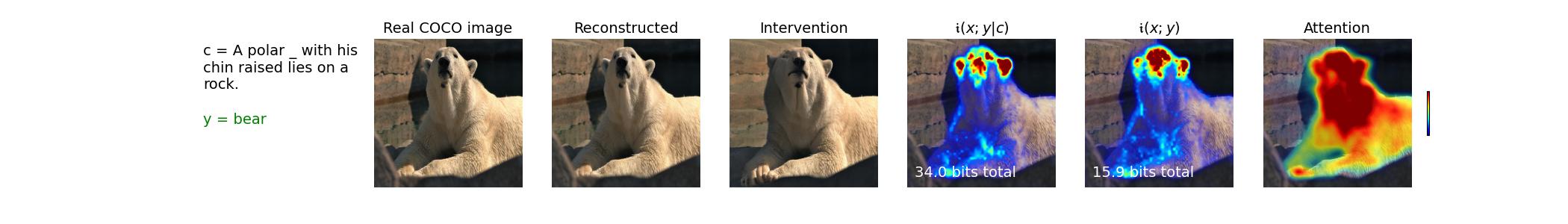} 
    \includegraphics[width=0.99\textwidth,trim={5cm 8mm 4cm 13mm},clip]{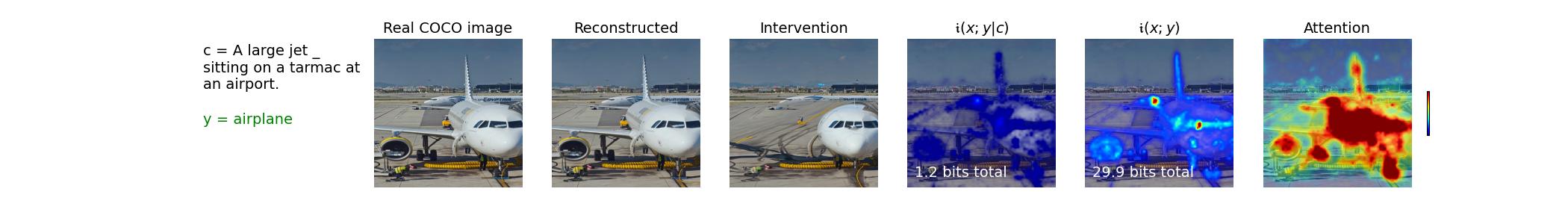} 
    \includegraphics[width=0.99\textwidth,trim={5cm 8mm 4cm 13mm},clip]{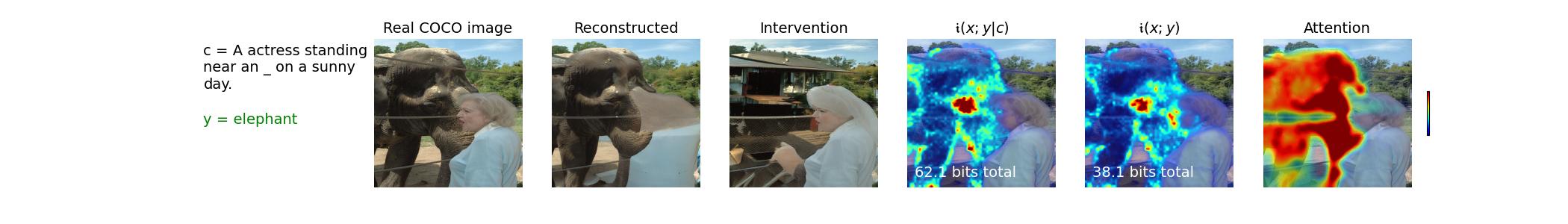} 
    \includegraphics[width=0.99\textwidth,trim={5cm 8mm 4cm 13mm},clip]{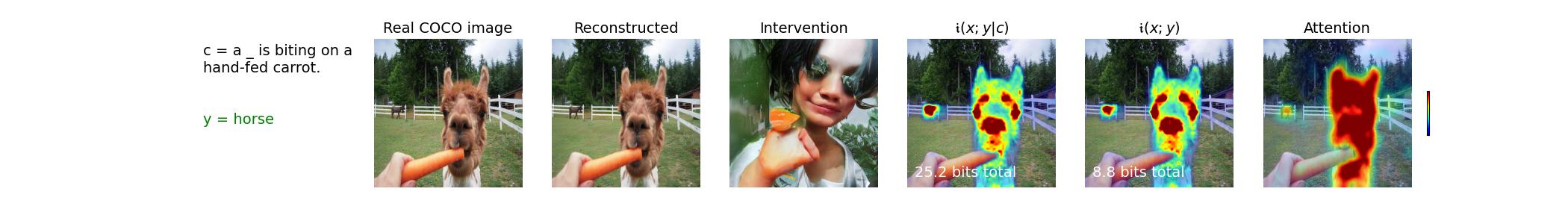} 
    \caption{Examples of word omission interventions}
    \label{fig:mod1}
\end{figure}

\begin{figure}[p]
    \centering
    \includegraphics[width=0.99\textwidth,trim={5cm 8mm 4cm 1mm},clip]{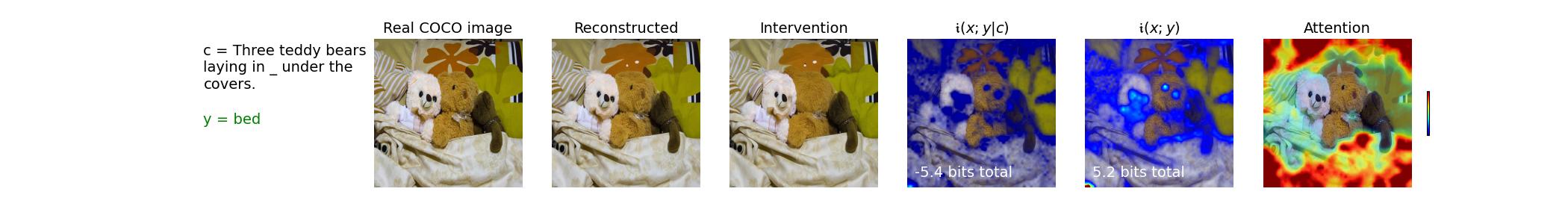} 
    \includegraphics[width=0.99\textwidth,trim={5cm 8mm 4cm 13mm},clip]{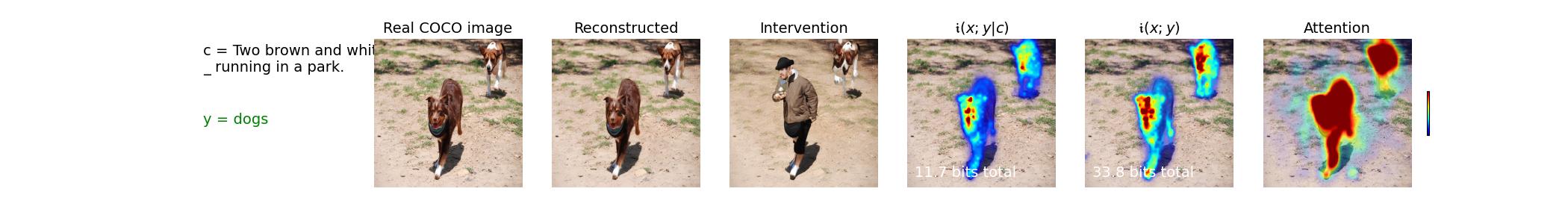} 
    \includegraphics[width=0.99\textwidth,trim={5cm 8mm 4cm 13mm},clip]{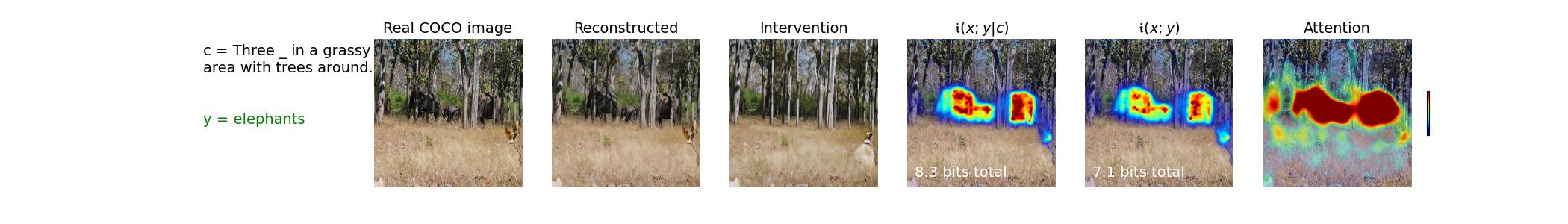} 
    \includegraphics[width=0.99\textwidth,trim={5cm 8mm 4cm 13mm},clip]{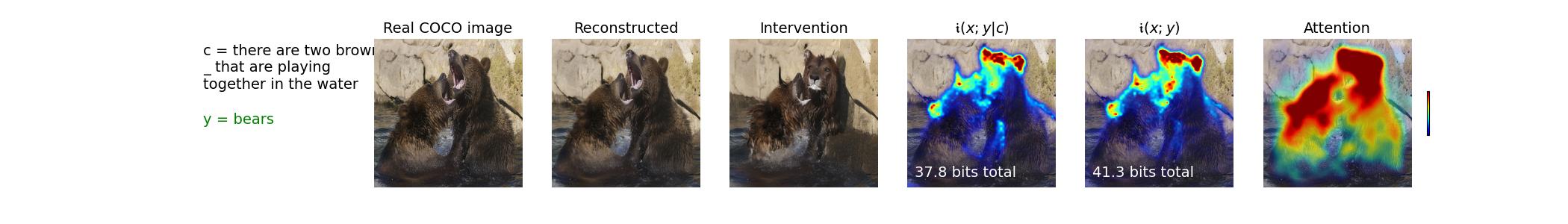} 
    \includegraphics[width=0.99\textwidth,trim={5cm 8mm 4cm 13mm},clip]{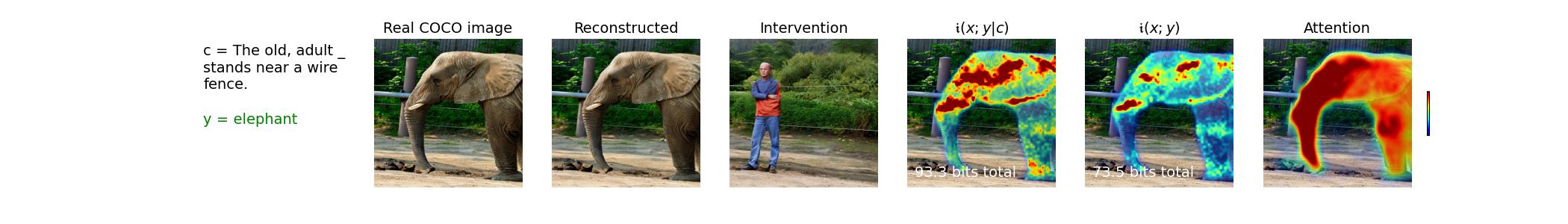} 
    \includegraphics[width=0.99\textwidth,trim={5cm 8mm 4cm 13mm},clip]{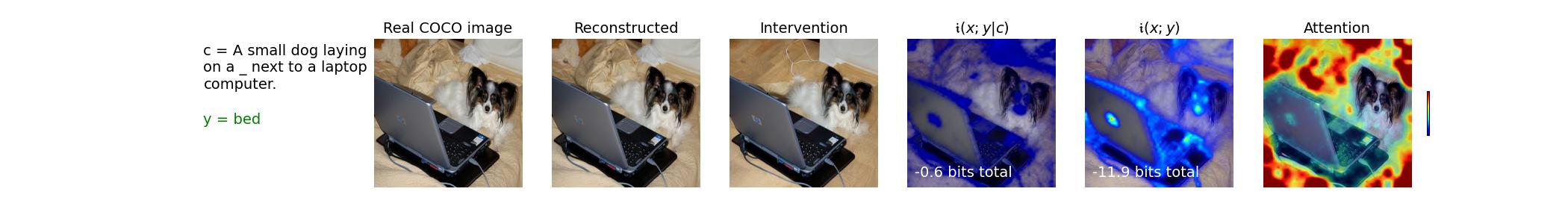} 
    \includegraphics[width=0.99\textwidth,trim={5cm 8mm 4cm 13mm},clip]{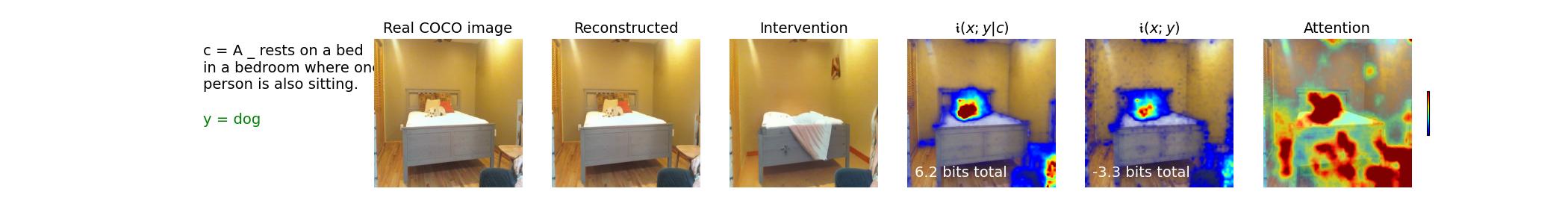} 
    \includegraphics[width=0.99\textwidth,trim={5cm 8mm 4cm 13mm},clip]{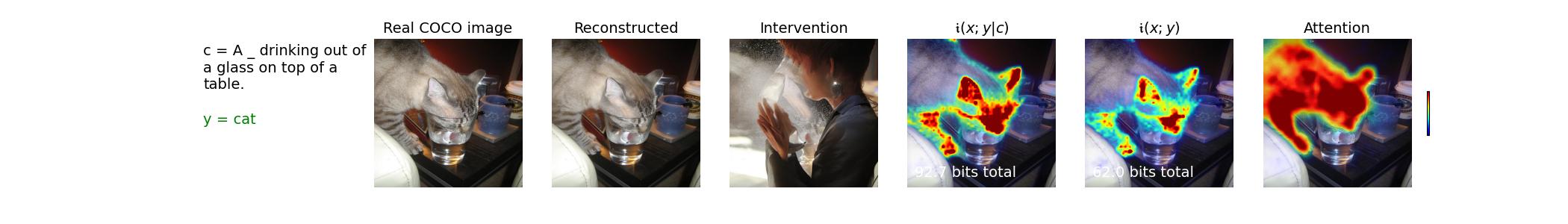} 
    \includegraphics[width=0.99\textwidth,trim={5cm 8mm 4cm 13mm},clip]{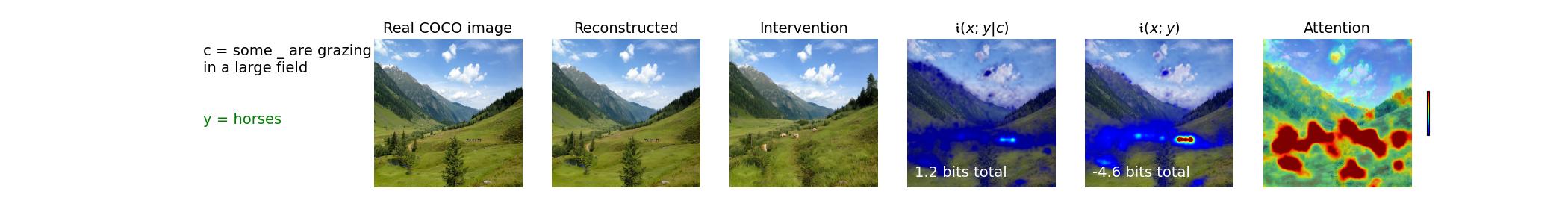} 
    \includegraphics[width=0.99\textwidth,trim={5cm 8mm 4cm 13mm},clip]{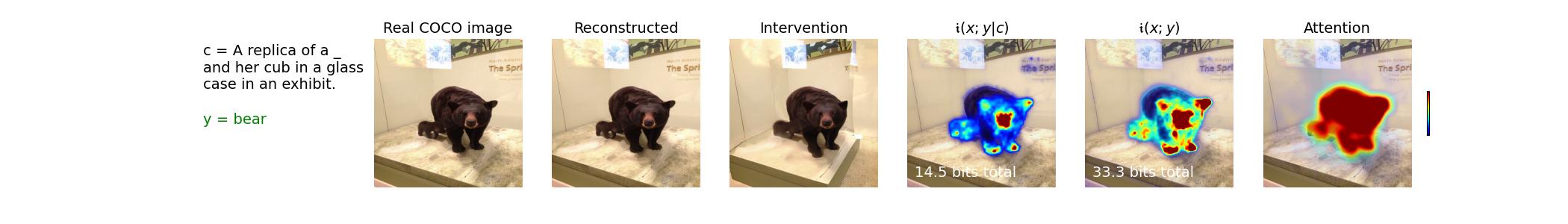} 
    \includegraphics[width=0.99\textwidth,trim={5cm 8mm 4cm 13mm},clip]{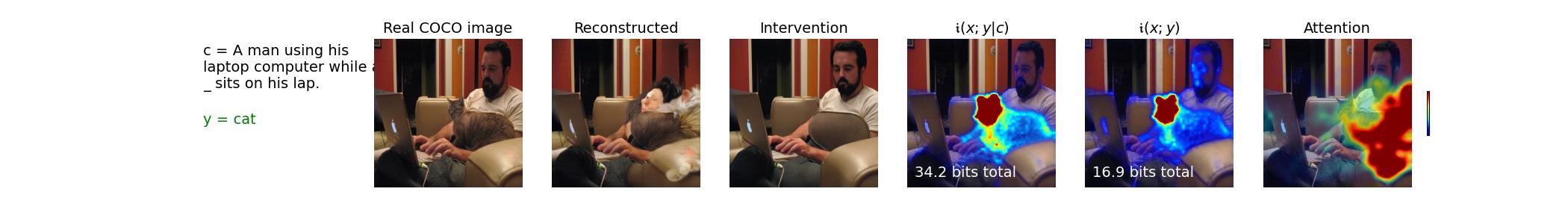} 
    \includegraphics[width=0.99\textwidth,trim={5cm 8mm 4cm 13mm},clip]{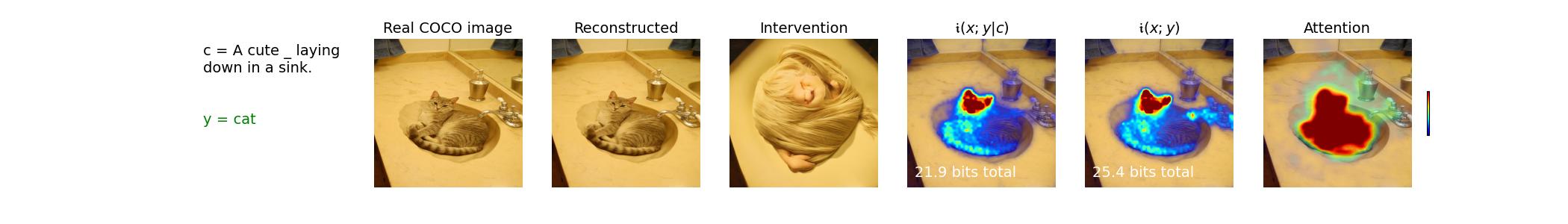} 
    \caption{Examples of word omission interventions}
    \label{fig:mod2}
\end{figure}

\end{document}

%% file: math_commands.tex
\usepackage{amsmath,amsfonts,bm,nicefrac}

\def\eqref#1{Eq.~\ref{#1}}

\def\1{\bm{1}}

\def\vc{{\bm{c}}}

\def\vf{{\bm{f}}}

\def\vx{{\bm{x}}}
\def\vy{{\bm{y}}}

\DeclareMathAlphabet{\mathsfit}{\encodingdefault}{\sfdefault}{m}{sl}
\SetMathAlphabet{\mathsfit}{bold}{\encodingdefault}{\sfdefault}{bx}{n}

\def\gP{{\mathcal{P}}}

\def\gX{{\mathcal{X}}}
\def\gY{{\mathcal{Y}}}

\def\sR{{\mathbb{R}}}

\newcommand{\E}{\mathbb{E}}

\newcommand{\R}{\mathbb{R}}

\DeclareMathOperator*{\argmax}{arg\,max}

\newcommand{\snr}{\gamma}

\newcommand{\logsnr}{\alpha}
\newcommand{\half}{\nicefrac{1}{2}}

\def\eps{{\bm \epsilon}}

\newcommand{\be}{\begin{eqnarray} \begin{aligned}}
\newcommand{\ee}{\end{aligned} \end{eqnarray} }
\newcommand{\benn}{\begin{eqnarray*} \begin{aligned}}
\newcommand{\eenn}{\end{aligned} \end{eqnarray*} }
\newcommand{\xhat}{\hat{\bm x}}

\newcommand{\epshat}{\bm \hat \eps_\logsnr}
\newcommand{\epsbar}{\bm \bar \eps}

\newcommand{\ds}{\frac{d}{d \snr}}

\newcommand{\vz}{\bm x_\logsnr}
\newcommand{\ii}{\mathfrak{i}}  

\DeclareMathOperator{\mmse}{mmse}

\newcommand{\norm}[1]{{\| #1 \|^2 }}